\documentclass[11pt]{article}

\usepackage[final]{acl}

\usepackage{microtype}
\usepackage{subcaption}
\usepackage{booktabs} % for professional tables

\usepackage{times}
\usepackage{latexsym}

\usepackage[T1]{fontenc}
\usepackage[utf8]{inputenc}

\usepackage{microtype}

\usepackage{inconsolata}

\usepackage{graphicx}

\usepackage{caption}
\usepackage{makecell}
\usepackage{multirow}
\usepackage{newfloat}
\usepackage{float}
\usepackage{afterpage}
\usepackage{tabularx}
\usepackage{array}
\usepackage{pifont}
\usepackage{amsmath}
\usepackage{amssymb}
\usepackage{mathtools}
\usepackage{amsthm}
\newcommand{\DatasetName}{LLMTrace}

\title{LLMTrace: A Corpus for Classification and Fine-Grained Localization of AI-Written Text}

\author{
  \textbf{Irina Tolstykh},
  \textbf{Aleksandra Tsybina},
  \textbf{Sergey Yakubson},
  \textbf{Maksim Kuprashevich}
\\
\\
  SALUTEDEV LLC, Tashkent, Uzbekistan
\\
  \small{
    \textbf{Correspondence:} \href{mailto:irinakr4snova@gmail.com}{irinakr4snova@gmail.com}
  }
}

\begin{document}
\maketitle
\begin{abstract}
The widespread use of human-like text from Large Language Models (LLMs) necessitates the development of robust detection systems. However, progress is limited by a critical lack of suitable training data; existing datasets are often generated with outdated models, are predominantly in English, and fail to address the increasingly common scenario of mixed human-AI authorship. Crucially, while some datasets address mixed authorship, none provide the character-level annotations required for the precise localization of AI-generated segments within a text. To address these gaps, we introduce \texttt{\DatasetName{}}, a new large-scale, bilingual (English and Russian) corpus for AI-generated text detection. Constructed using a diverse range of modern proprietary and open-source LLMs, our dataset is designed to support two key tasks: traditional full-text binary classification (human vs. AI) and AI-generated interval detection, facilitated by character-level annotations. We believe \texttt{\DatasetName{}} will serve as a vital resource for training and evaluating the next generation of more nuanced and practical AI detection models. 
\end{abstract}

\section{Introduction}
\label{sec:introduction}

The rapid advancement and widespread adoption of Large Language Models (LLMs) have enabled the generation of human-like text at an unprecedented scale. This text is often so convincing that human evaluators struggle to distinguish it from human writing, with performance frequently hovering around chance levels \citep{jakesch2023human, milivcka2025humans}. This capability, while offering enormous benefits, presents a significant dual-use challenge: the same models that assist in creative writing can also generate misinformation, compromise academic integrity, and automate malicious communication. The unreliability of human detection underscores the critical need for robust, automated systems to maintain a trustworthy digital ecosystem.

\begin{figure*}[t]
    \centering
    \begin{subfigure}{0.53\textwidth}
        \centering
        \includegraphics[width=\linewidth]{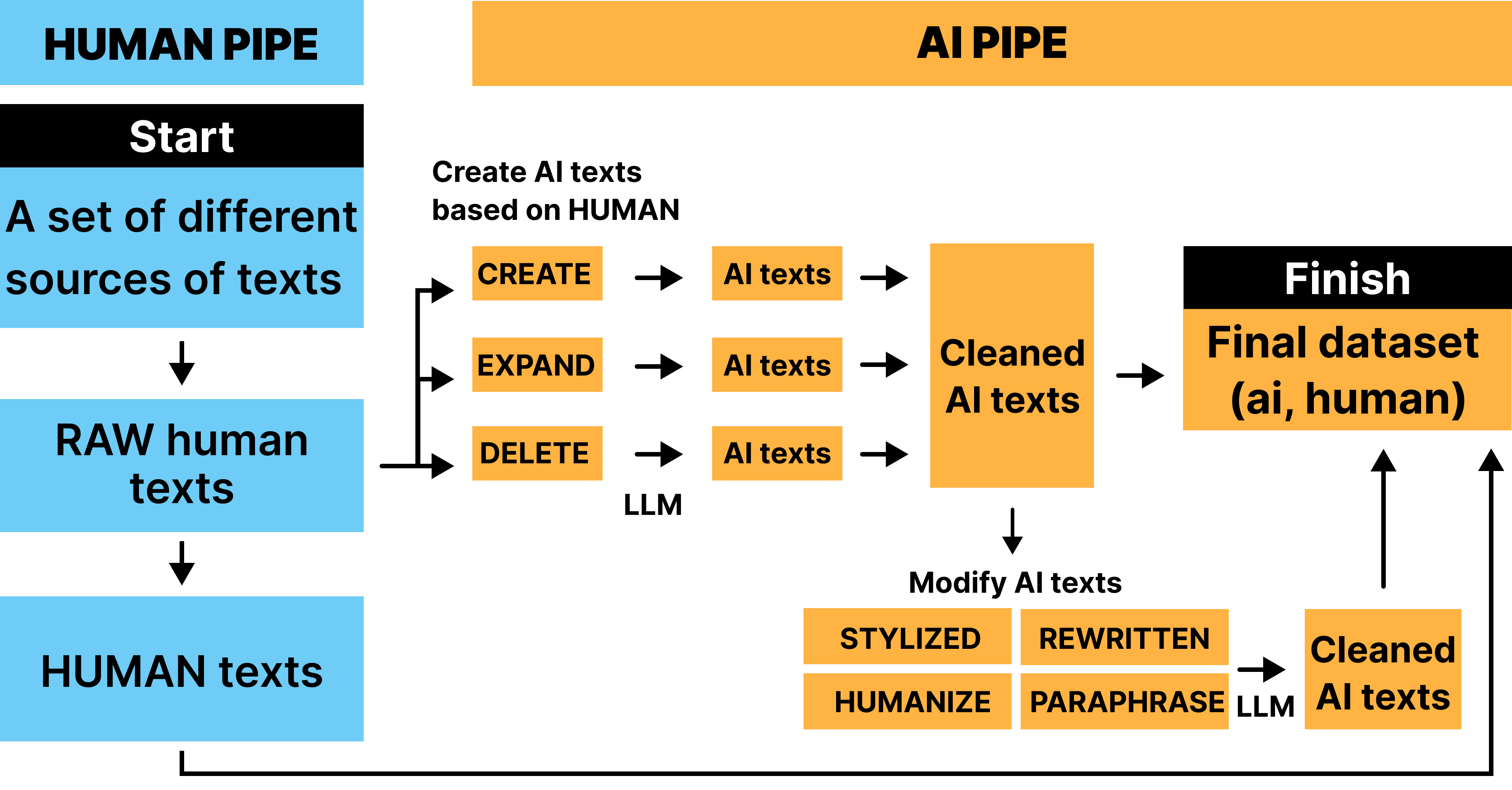}
        \caption{Classification dataset}
        \label{fig:human_ai_pipeline}
    \end{subfigure}
    \begin{subfigure}{0.43\textwidth}
        \centering
        \includegraphics[width=\linewidth]{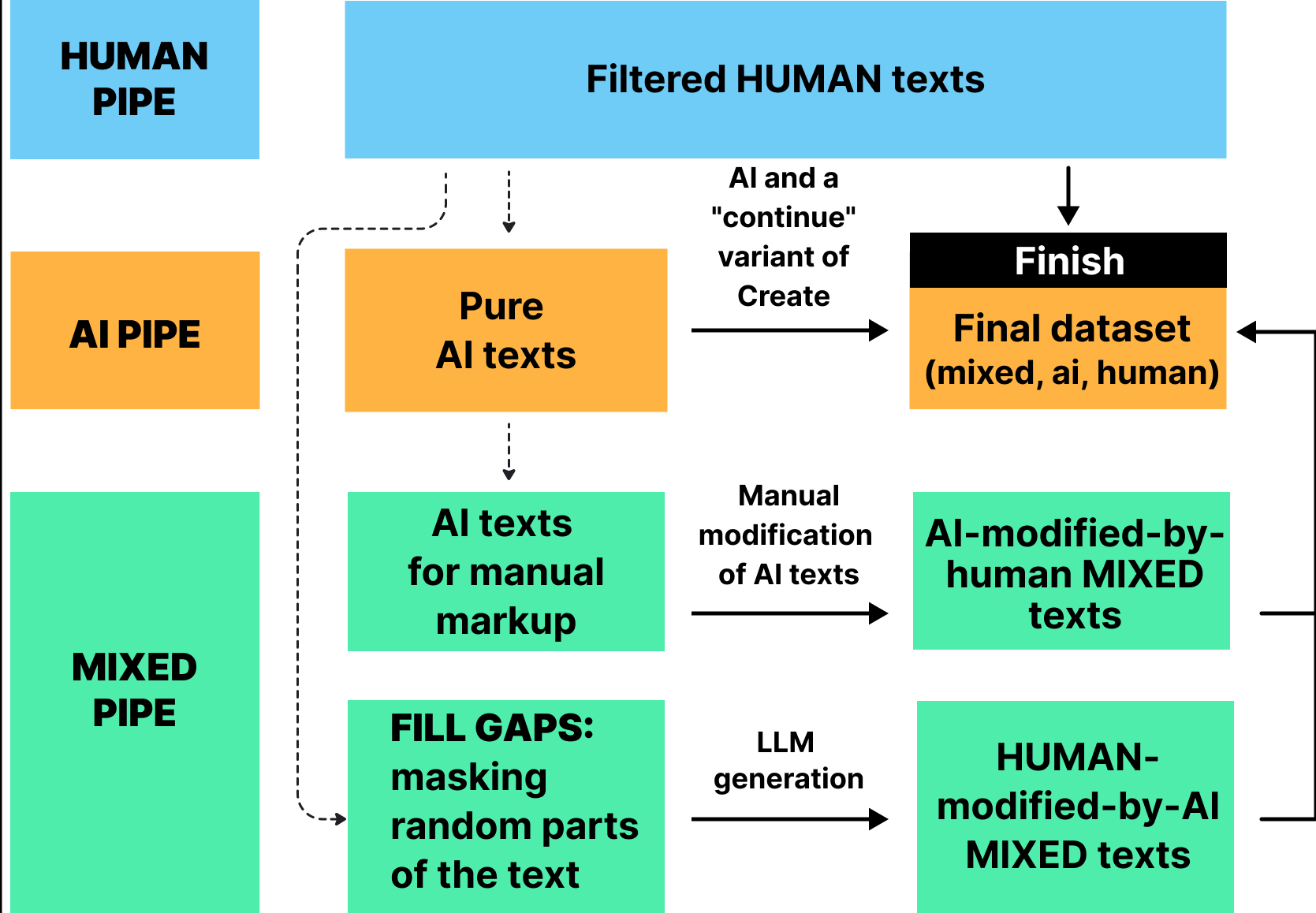}
        \caption{Detection dataset}
        \label{fig:human_ai_mixed_pipeline}
    \end{subfigure}

    \caption{Data curation pipelines: (a) classification builds parallel human/AI corpora; (b) detection aggregates filtered human, pure-AI, and mixed-authorship samples. Dashed arrows mark source texts reused for generation or modification (e.g., ``Fill Gaps'' inserts AI spans into human texts; manual markup edits AI texts).}
    \label{fig:two_gen_pipelines}
\end{figure*}

However, the effectiveness of any detection model is fundamentally limited by its training data. A review of the current landscape (Section~\ref{sec:related_works}) reveals several critical gaps: existing datasets often rely on outdated LLMs, are predominantly English-centric with a scarcity of resources for languages like Russian, and focus on simple classification, overlooking the increasingly common scenario of mixed human-AI authorship.

Perhaps the most significant gap is the absence of large-scale resources for the task of \textit{localizing} AI-generated segments within these mixed texts. While a few datasets support boundary detection between authorship (Table \ref{tab:detection_benchmarks}), they are often limited to single transition points or lack the precise, character-level annotations required to train granular, high-resolution detectors.

To address these limitations, we introduce \texttt{\DatasetName{}}, a new large-scale corpus designed to push the boundaries of AI text detection. We instantiate \DatasetName{} for English (the most studied target) and Russian (a high-impact under-served language with its own generator families, e.g.\ GigaChat, YaGPT). The pipeline itself is language-agnostic and extends to new languages given a human corpus and adapted prompts, positioning \DatasetName{} as an extensible framework rather than a fixed snapshot.

Our work makes two primary contributions. First, we provide a comprehensive classification dataset built from a diverse suite of 38 LLMs across nine domains, utilizing a wide variety of prompt types to create sophisticated and realistic examples. Second, and most crucially, we introduce the large-scale detection dataset, constructed using 31 generators, with precise, character-level annotations for mixed-authorship texts. These mixed examples are created through a robust pipeline involving automated gap-filling in human texts, AI-based continuation of human texts, and meticulous manual editing of AI texts by humans, ensuring a high degree of complexity. By making this resource publicly available, we aim to facilitate a new wave of research into more nuanced, practical, and robust AI detection models.
The dataset, models, and training code will be publicly available at \textbf{https://sweetdream779.github.io/LLMTrace-info/}.

\begin{table*}[t]
    \centering
    \caption{Statistics of the detection datasets. Our \DatasetName{} dataset includes human, AI, and mixed texts, with the size of the mixed subset noted in parentheses.}
    \label{tab:detection_benchmarks}
    \small
    \begin{tabular}{l c c c c c c}
        \hline
        \textbf{Dataset} & \textbf{\makecell[c]{Dataset Size}} & \textbf{\# Domains} & \textbf{\# Lang} & \textbf{\# Gen} & \textbf{\# Intervals} & \textbf{\makecell[c]{Human collab}}\\
        \hline
        M4GT-Bench & 31.9k & 2 & 1 & 5 & 1 & \ding{55} \\
        RoFT & 27.6k & 4 & 1 & 5 & 1 & \ding{55} \\
        RoFT-chatgpt & 6.9k & 4 & 1 & 1 & 1 & \ding{55} \\
        TriBERT & 17.1k & 1 & 1 & 1 & 1-3 & \ding{55} \\
        CoAuthor & 1.4k & 2 & 1 & 1 & arbitrary & \ding{51} \\
        \hline
        \textbf{\DatasetName{} (ours)} & 79,342 \textbf{(27.7k mixed)} & 9 & 2 & 31 & arbitrary & \ding{51} \\
        \hline
    \end{tabular}
\end{table*}

\section{Related Works}
\label{sec:related_works}

Robust evaluation of AI text detectors requires diverse benchmarks for two key tasks: binary classification (human vs. AI) and the localization of AI-generated content within mixed-authorship texts. Existing detection resources are summarized in Table~\ref{tab:detection_benchmarks}; a parallel comparison against 17 classification benchmarks is provided in Appendix Table~\ref{tab:classification_benchmarks}.

\subsection{Classification Benchmarks}

Foundational benchmarks like MAGE \citep{li2023mage}, MGTBench \citep{he2024mgtbench}, BUST \citep{cornelius2024bust}, RAID \citep{dugan2024raid}, DetectRL \citep{wu2024detectrl}, and HC3 \citep{guo2023close} have laid important groundwork but are often limited by narrow prompt diversity, a small suite of generator models, or restricted domain coverage. Multilingual resources such as MULTITuDE \citep{macko2023multitude}, M4GT-Bench \citep{wang2024m4gt}, and MultiSocial \citep{macko2024multisocial} have expanded language coverage but are frequently confined to specific domains like news or social media.

The RuATD dataset \citep{shamardina2022findings}, currently the only large-scale benchmark for AI text detection in Russian covering 14 generators, spans diverse domains but remains limited to a single language and lacks full domain coverage.

Domain-specific corpora, while valuable, target narrow areas like Wikipedia edits (WETBench \citep{quaremba2025wetbench}), scholarly reviews (Peer Review Detection \citep{yu2025your}), or student essays (GEDE \citep{gehring2025assessing}). Similarly, robustness benchmarks like ESPERANTO \citep{ayoobi2024esperanto} and SHIELD \citep{ayoobi2025beyond} explore adversarial attacks or "humanification" but remain limited in scope. Finally, while datasets like Beemo \citep{artemova2024beemo} and MixSet \citep{zhang2024llm} consider mixed authorship, they do so only in a classification setting and at a smaller scale.

We also note the existence of active defense methods such as watermarking \citep{kirchenbauer2023watermark}; however, our work focuses on passive detection in non-watermarked settings, which remains the prevalent scenario for open-ended text generation.

\subsection{Detection Benchmarks}

Existing mixed-authorship benchmarks often face structural or domain limitations. M4GT-Bench \cite{wang2024m4gt} and RoFT \cite{dugan2023real} (along with RoFT-chatgpt \cite{kushnareva2023ai}) model only single-transition continuations, failing to capture complex arbitrary boundaries between human and AI segments. While TriBERT \cite{zeng2024towards} supports up to three such transitions, it is confined to student essays. Similarly, CoAuthor \cite{lee2022coauthor} captures collaborative writing, but its frequent minor edits introduce noise that complicates systematic evaluation.

\section{\DatasetName{}: Design Principles and Curation Pipeline}
\label{sec:dataset}

\begin{figure*}[!th]
    \centering
    \includegraphics[width=0.99\textwidth,keepaspectratio]{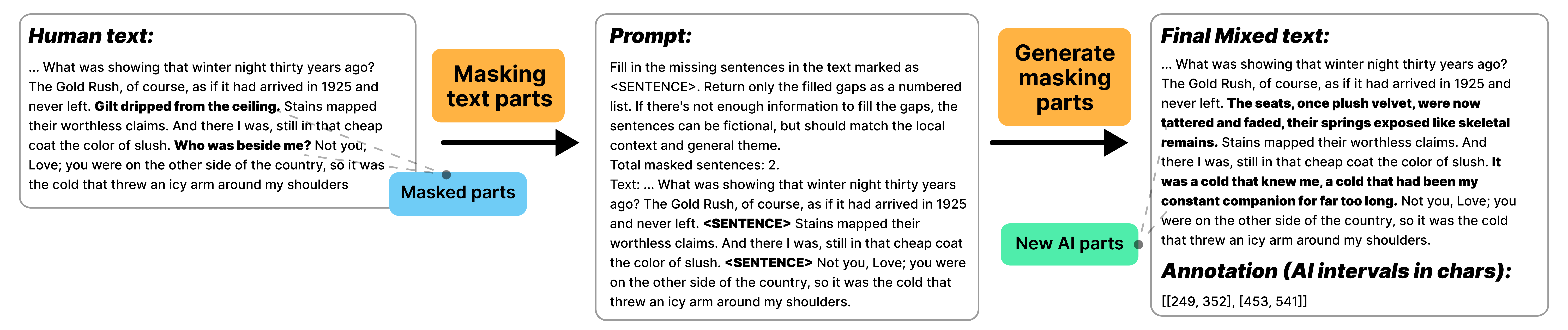}
    \caption{The automated mixed-text generation process: sentences in a human text are masked and then filled in by an LLM, resulting in a text with annotated AI-generated intervals.}
    \label{fig:text_gen_example}
\end{figure*}

We detail the methodology for creating \texttt{\DatasetName{}}, a large-scale dataset structured to support binary classification and segment-level detection, designed to ensure diversity, robustness, and real-world applicability. We instantiate this methodology for English and Russian (motivation in Section~\ref{sec:introduction}), but every stage of the pipeline (human corpus aggregation, prompt-driven AI generation, mixed-text construction, and post-processing) is language-agnostic and can be applied to additional languages given a new human corpus, adapted prompt templates, and language-tuned filters. The overall pipelines for classification and detection datasets generation are shown in Figure~\ref{fig:two_gen_pipelines}.

\subsection{Target Tasks and Annotation Schema}
\label{subsec:tasks}

Our dataset supports two tasks (examples in Appendix Figure~\ref{fig:example}).
First, \textbf{Binary Text Classification} is defined as $\mathcal{D}_{\text{cls}} = \{(x_i, y_i)\}_{i=1}^{N}$, where $x_i$ is the text and $y_i \in \{\text{'human'}, \text{'ai'}\}$ denotes authorship. 
Second, \textbf{AI Interval Detection} follows $\mathcal{D}_{\text{det}} = \{(x_i, y_i, S_i)\}_{i=1}^{M}$, where $S_i = \{(start_j, end_j)\}_{j=1}^{K_i}$ is a set of character-level offsets marking the $K_i$ AI-generated spans. $S_i$ is empty for human texts, covers the full text for AI samples, and marks specific regions in mixed texts. Both datasets are partitioned into standard training, validation, and test sets for the experiments in Section~\ref{sec:exps}.

\subsection{Classification Dataset Curation}
\label{subsec:classification_curation}

The curation of our classification dataset (Figure~\ref{fig:human_ai_pipeline}) aimed to create a comprehensive and challenging resource. The design was guided by three core principles: diversity across domains and lengths, a wide range of generator models (including both modern and legacy), and complex generation scenarios designed to produce subtle and sophisticated examples.

\textbf{Human Corpus Collection.}
The foundation is a large, diverse corpus of human-authored EN/RU texts aggregated from open-source collections (full list in Appendix~\ref{app:sources}). It spans eight shared domains: \textbf{Short-form text}, \textbf{News}, \textbf{Question}, \textbf{Review}, \textbf{Factual text}, \textbf{Poetry}, \textbf{Story}, \textbf{Article}, plus a ninth, \textbf{Paper Abstract}, for English to cover academic writing.

\textbf{Length-Balanced Sampling.}
To ensure structural balance, we sampled human texts uniformly from predefined word-count buckets. This balance was then propagated to the AI corpus by including explicit output-length instructions within our generation prompts. This process ensures the final AI collection mirrors the length distribution of the human corpus, minimizing the risk of models learning to use text length as a simple heuristic for classification.

\textbf{AI Text Generation Scenarios.}
Using the length-balanced and domain-balanced human corpus as a start, we generated a parallel AI corpus with a diverse library of prompt templates. These templates covered four main scenarios (see Appendix~\ref{app:prompt_examples} for examples): generating new text based on a human source (\textbf{Create}), shortening it via summarization or simplifying (\textbf{Delete}), expanding it with more detail (\textbf{Expand}), and modifying \textbf{outputs from these pipelines} through different methods, e.g., humanization or stylization (\textbf{Update}). A key benefit of this process is that every AI text is thematically paired with a human text. This design is crucial as it compels detection models to learn subtle stylistic and structural differences, rather than relying on spurious correlations with the topic.

\textbf{Generator Model Diversity.}
To ensure our dataset is not overfitted to the artifacts of a single model family, we utilized a wide spectrum of LLMs. This included: (a) modern proprietary models (e.g., from the Gemini~\citep{comanici2025gemini} and OpenAI GPT-4~\citep{gpt4} families); (b) a diverse set of modern open-source models (e.g., Qwen3~\citep{yang2025qwen3}, DeepSeek-R1~\citep{guo2025deepseek}); (c) widely-used legacy models (e.g., GPT-3.5~\citep{Ouyang2022instructgpt}); and (d) for the Russian dataset, language-specific models such as GigaChat~\citep{mamedov2025gigachat} and YaGPT\footnote{https://ya.ru/ai/gpt}. We intentionally included models of various sizes (from 760M to 72B parameters) and capabilities (with and without advanced reasoning) to capture a broad range of potential generation signatures.
A complete list of all used LLMs is provided in Appendix \ref{app:generators}.

\begin{figure*}[!h]
    \centering
    \includegraphics[width=0.95\textwidth,keepaspectratio]{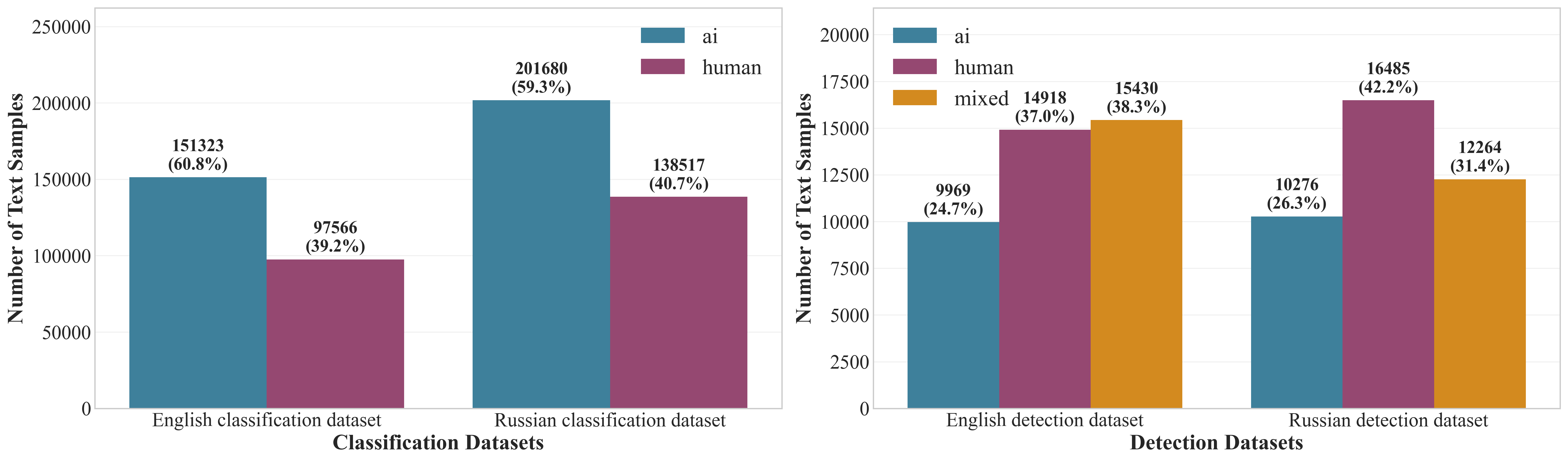}
    \caption{The number of samples for each label in the \DatasetName\ dataset.}
    \label{fig:full_text_stats}
\end{figure*}

\subsection{Detection Dataset Curation}
\label{subsec:detection_curation}

The detection dataset was curated to provide realistic human-AI collaboration examples with a variable number of AI intervals via three distinct pathways (Figure~\ref{fig:human_ai_mixed_pipeline}).

First, \textbf{Human Text Gap-Filling} involved masking randomly selected sentences at arbitrary positions within human texts for LLMs to fill (Figure~\ref{fig:text_gen_example}). Due to the complexity of this task, we exclusively employed SOTA models (Gemini-2.5-flash \citep{comanici2025gemini}, GPT-4/4o/o1 \citep{gpt4, hurst2024gpt, jaech2024openai}) to ensure the highest quality of generated content.

Second, \textbf{Human Text Continuation} adapted "continue" prompts (a variant of \textit{Create}) from our classification set, creating single-interval texts that inherit full generator diversity.

Finally, \textbf{Manual Editing of AI Texts} produced a unique sub-corpus of the most challenging examples. Though smaller due to the expensive nature of this process, these examples are vital as human modifications break consistent AI statistical patterns, forcing models to learn more nuanced and robust features. Crucially, editors focused solely on content modification: they inserted separator tokens only as a convenience marker within the editing interface to delimit working regions, \textit{not} to define final segment boundaries. Ground-truth labels were computed deterministically via character-level alignment of the final edited text against the original AI source, making the resulting interval annotations fully automatic and independent of any subjective human judgment (details in Appendix~\ref{app:manually_modified_texts}).

\subsection{Post-processing and Filtering}
\label{subsec:postprocessing}

To ensure data quality, all human and AI texts underwent a rigorous filtering pipeline that involved strict language filtering (English or Russian), removing duplicates, low-quality content (e.g., incomplete or repetitive), and any text shorter than five words, including short AI-generated continuations. We also stripped LLM-specific artifacts by filtering common refusal phrases (e.g., "As a language model...") and removing Markdown formatting to normalize the data.

As detailed in Figure~\ref{fig:full_text_stats}, the resulting classification datasets comprise $\sim$249k English and $\sim$340k Russian samples ($\sim$60/40 AI-to-human ratio), while the detection datasets feature a three-way split of human, AI, and mixed texts. Text-length coverage is broad across all word-count ranges (0 to 500+ words; Appendix Figure~\ref{fig:two_len_plots}), and domain representation is robust, averaging $\sim$16.8k-25.2k AI texts for classification and over 1.2k AI and 1.6k mixed texts for detection per domain (see Appendix~\ref{app:statistics}). Figure~\ref{fig:interval_nums} further shows that the average number of AI intervals correlates positively with word count, ensuring the dataset contains challenging examples with multiple, non-trivial AI insertions.

\begin{figure*}[!h]
    \centering
    \includegraphics[width=0.8\textwidth,keepaspectratio]{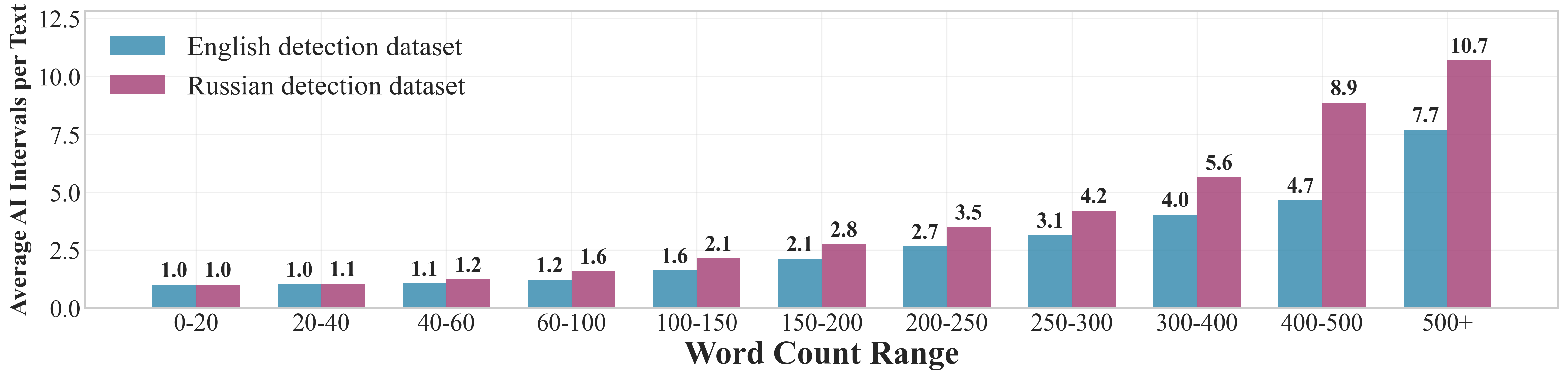}
    \caption{Average number of AI intervals per text in each word count range for the detection datasets.}
    \label{fig:interval_nums}
\end{figure*}

\section{Quality and Complexity Assessment}
\label{sec:quality}

We analyze dataset quality using a dual approach: following \citet{gritsai2024ai} to assess deep structural properties via topological and perturbation metrics, and complementing this with textual similarity metrics from the MultiSocial benchmark \cite{macko2024multisocial} to quantify how closely the output of each LLM resembles the source text it was based on. These analyses collectively demonstrate how indistinguishable our AI-generated texts are from their human counterparts.

\subsection{Topological Statistics}

Recent work has shown that the intrinsic dimensionality of a text's embedding space can serve as a feature to distinguish between human and machine origins \citet{tulchinskii2023intrinsic}. This is often measured using the Persistence Homology Dimension (PHD). While earlier models tended to produce texts with lower PHD than humans, modern LLMs generate text with much more similar topological complexity.

To quantify the similarity between the PHD distributions of human ($h_d$) and AI-generated ($m_d$) texts, we use the symmetric KL-divergence score, $\text{KL}_{\text{TTS}}$, as proposed by \citet{gritsai2024ai}:
$$ \text{KL}_{\text{TTS}} (h_d, m_d) = | D_{\text{KL}}(h_d || m_d) - D_{\text{KL}}(m_d || h_d) | $$
A lower $\text{KL}_{\text{TTS}}$ score indicates that the two distributions are more similar, suggesting that the AI texts are harder to distinguish from human texts based on this topological feature. 
Following \citet{gritsai2024ai}, we extract embeddings using \texttt{roberta-base}\footnote{https://huggingface.co/FacebookAI/roberta-base}.
We acknowledge that using an English-centric encoder (\texttt{roberta-base}) for Russian texts may introduce cross-lingual bias. However, we retained this choice to maintain direct comparability with the baseline values established in prior work \citet{gritsai2024ai} and to isolate dataset-level differences from metric-level variations.

Our English classification dataset achieves a competitive $\text{KL}_{\text{TTS}}$ of 0.0189 when benchmarked against other public datasets reported by \citet{gritsai2024ai} (see Table~\ref{tab:comparison_results}). Notably, our Russian dataset shows an exceptionally low $\text{KL}_{\text{TTS}}$ of 0.0032, one of the lowest among all compared Russian-language datasets, demonstrating the high quality of our generation pipeline.

The analysis of the detection datasets in Table~\ref{tab:detection_results} provides further insights. For both English and Russian, the $\text{KL}_{\text{TTS}}$ scores between \textit{AI} and \textit{human} texts remain low, confirming the quality of the purely AI-generated texts in this subset. Most importantly, the score between \textit{mixed} and \textit{human} texts is exceptionally low (0.0017 for EN, 0.0332 for RU). This quantitatively confirms that our methods for creating mixed texts produce challenging examples that are topologically very similar to purely human-written content. This conclusion is further supported by the closely aligned mean PHD values shown in Tables~\ref{tab:comparison_results} and \ref{tab:detection_results}, especially the near-identical means for human and mixed texts in the English detection dataset. A visual representation of these highly overlapping distributions is provided in Appendix~\ref{app:phd_distributions}.

% --- Table for Comparison with other Datasets ---
\begin{table*}[t!]
    \centering
    \caption{Quality statistics of our classification datasets vs.\ public datasets (other values from \cite{gritsai2024ai}); lower $\text{KL}_{\text{TTS}}$ and $|\Delta_{\text{shift}}|$ are better.}
    \small
    \begin{tabular}{l c c c c}
    \hline
        Dataset & $\text{KL}_{\text{TTS}}$ $\downarrow$ & $\text{PHD}_{\text{human}}$ & $\text{PHD}_{\text{machine}}$ &  |$\Delta_{\text{shift}}$| $\downarrow$ \\
        \hline
        GhostBuster\citet{verma2023ghostbuster} & 0.053 & 9.84 $\pm$ 1.18 & 9.76 $\pm$ 1.15 & 0.024 \\
        MGTBench\citet{he2024mgtbench} & 0.043  & 8.77 $\pm$ 1.31 & 9.97 $\pm$ 1.02 &  0.031 \\
        M4\citet{wang2024m4gt} & 0.036 & 7.26 $\pm$ 1.99 & 8.59 $\pm$ 1.40 & 0.107 \\
        MAGE\citet{li2023mage} & 0.011  & 9.80 $\pm$ 2.14 & 9.38 $\pm$ 3.04 & 0.094 \\
        SemEval24 Multi\citet{wang2024semeval} & 0.001  & 9.65 $\pm$ 1.81 & 9.42 $\pm$ 1.44 &  0.059 \\
        RuATD\citet{shamardina2022findings} & 0.007 & 7.33 $\pm$ 1.40 & 7.46 $\pm$ 1.41 &  0.315 \\
        \hline
        \DatasetName{} (EN) & 0.0189 & 9.19 $\pm$ 1.87 & 9.91 $\pm$ 1.64 & -0.00317 \\
        \DatasetName{} (RU) & 0.0032 & 6.98 $\pm$ 0.82 & 7.06 $\pm$ 0.69 & 0.00074 \\
    \hline
    \end{tabular}
    \label{tab:comparison_results}
\end{table*}

% --- Table for Detection Dataset Analysis ---
\begin{table*}[t!]
    \centering
    \caption{Topological statistics for our detection datasets. The very low $\text{KL}_{\text{TTS}}$ between human and mixed texts (especially EN) reflects the complexity of mixed-authorship examples.}
    \small
    \begin{tabular}{l c c c c c}
    \hline
        Dataset & $\text{PHD}_{\text{human}}$ & $\text{PHD}_{\text{mixed}}$ & $\text{PHD}_{\text{ai}}$ & \makecell{$\text{KL}_{\text{TTS}}$ $\downarrow$ \\ (mixed vs \\human)} & \makecell{$\text{KL}_{\text{TTS}}$ $\downarrow$ \\ (ai vs human)} \\
        \hline
        \DatasetName{} (EN) & 9.27 $\pm$ 1.87 & 9.26 $\pm$ 1.83 & 9.83 $\pm$ 1.65 & 0.0017 & 0.0233 \\
        \DatasetName{} (RU) & 7.12 $\pm$ 0.87 & 7.15 $\pm$ 0.63 & 7.27 $\pm$ 0.58 & 0.0332 & 0.0483 \\
    \hline
    \end{tabular}
    \label{tab:detection_results}
\end{table*}

\begin{table*}[t!]
\centering
\caption{Global average similarity metrics for the English and Russian classification datasets.}
\label{tab:global_similarity_metrics}
\resizebox{\textwidth}{!}{%
\begin{tabular}{lcccccc}
\hline
\textbf{Language} & \textbf{METEOR ↑} & \textbf{BERTScore ↑} & \textbf{n-gram ↑} & \textbf{ED-norm (LD) ↓} & \textbf{LangCheck ↓} & \textbf{MAUVE* ↓} \\
\hline
\DatasetName{} (EN) & 0.2665 & 0.6964 & 0.2164 & 3.0255 & 0.0019 & 0.1713 \\
\DatasetName{} (RU) & 0.1797 & 0.6767 & 0.1630 & 2.4792 & 0.0045 & 0.1696 \\
\hline
\multicolumn{7}{l}{\footnotesize{*MAUVE scores were computed on a random sample of 1k (human, AI) pairs per model. Other metrics were calculated on full data.}}
\end{tabular}
}
\end{table*}

\subsection{Perturbation Statistics}

This metric assesses dataset quality by measuring how text embeddings shift after minor, meaning-preserving adversarial changes, measured by the $\Delta_{\text{shift}}$ score from \citet{gritsai2024ai}. The perturbation is performed by replacing tokens with synonyms (using WordNet~\cite{miller1995wordnet} for English and RuWordNet\footnote{https://pypi.org/project/ruwordnet/} for Russian). The \texttt{multilingual-e5-large}\footnote{https://huggingface.co/intfloat/multilingual-e5-large} encoder is used to generate embeddings. We report the raw $\Delta_{\text{shift}}$ values to indicate the directionality of the shift. However, for quality assessment, a value with a magnitude ($|\Delta_{\text{shift}}|$) close to zero is desirable, as it signifies that both human and AI texts react similarly to the perturbations.

As shown in Table~\ref{tab:comparison_results}, our English classification dataset achieves a near-zero $\Delta_{\text{shift}}$ of -0.00317, indicating a very high degree of similarity to human texts in terms of adversarial robustness and outperforming many existing datasets. A visual comparison of the embedding shift distributions for our classification datasets is provided in Appendix~\ref{app:shift_comparison}.

\subsection{Textual Similarity Metrics}
\label{sec:textual_similarity}

While the preceding metrics assess structural integrity, we also evaluate the direct textual fidelity between AI-generated texts and their human source counterparts. Table~\ref{tab:global_similarity_metrics} summarizes the global average scores for these metrics on our English and Russian classification datasets. We present metrics where higher values indicate greater similarity (METEOR~\cite{banerjee2005meteor}, BERTScore~\cite{zhang2019bertscore}, n-gram) and metrics where lower values signify closer resemblance (Levenshtein Distance(LD), LangCheck, MAUVE~\cite{pillutla2021mauve}). The MAUVE metric is marked with an asterisk (*) to denote that it was calculated on a random sample of 1k text pairs per model. Detailed metric descriptions, as well as per-model and per-prompt type results, are provided in Appendix~\ref{app:similarity_appendix}.

Together, low $\text{KL}_{\text{TTS}}$ and near-zero $|\Delta_{\text{shift}}|$ show our generated texts are structurally indistinguishable from human writing. Lexical fidelity is equally strong: English BERTScore (0.6964), METEOR (0.2665), and n-gram (0.2164) all surpass top models in MultiSocial~\cite{macko2024multisocial}.

\section{Experiments}
\label{sec:exps}

To demonstrate the utility of our dataset, we conduct a series of baseline experiments. While simple token-level baselines exist, recent work suggests they struggle with the long-context coherence of modern LLMs. Therefore, we focus on establishing a strong transformer-based baseline using the GigaCheck~\citep{tolstykh2024gigacheck} framework. We fine-tune two backbones: Mistral-7B-v0.3~\citep{mistral} and the smaller mmBERT-base~\citep{marone2025mmbert} (307M parameters). For binary classification, these are trained as standard classifiers; for the interval localization task, following the GigaCheck methodology, we employ a DN-DAB-DETR~\citep{li2022dn} head on top of the frozen backbones. All training hyperparameters are detailed in Appendix~\ref{app:training_hyperparameters}.

For all experiments, we enforced strict grouping constraints during data partitioning: paired human and AI texts derived from the same topic were assigned to the same split to prevent content-based data leakage.

\subsection{Baseline Performance}
\label{subsec:baseline_experiments}

We first evaluate the models on our standard held-out test sets. 

For the \textbf{binary classification task}, we trained English-only (train/valid/test: 173,511 / 36,949 / 38,429), Russian-only (237,929 / 49,747 / 52,521), and bilingual models. The performance is presented in Table~\ref{tab:classification_results}. Both architectures achieve high F1 scores (\>0.96), establishing a strong baseline for future comparisons. A detailed per-domain breakdown is available in Appendix~\ref{app:classification_results}.

Zero-shot detectors lag dramatically: the best, Fast-DetectGPT~\citep{bao2024fast}, reaches only AI F1 0.69 overall vs.\ 0.99 for our fine-tuned baseline; DetectGPT~\citep{mitchell2023detectgpt} barely exceeds chance (AUROC 0.58); and Binoculars~\citep{hans2024spotting} collapses on Russian (AI F1 0.02). This confirms the task requires in-domain supervision (Appendix~\ref{app:zeroshot_results}).

For the \textbf{localizing AI-generated intervals} task, we trained the DN-DAB-DETR models on dedicated splits for English (train/valid/test: 27,766 / 5,536 / 7,015) and Russian (26,545 / 5,907 / 6,573). The quantitative results are summarized in Table~\ref{tab:detection_res}. We report mean Average Precision (mAP) with IoU thresholds fixed at 0.5 and averaged from 0.5 to 0.95.

\begin{table}[htbp!]
\centering
\caption{Binary classification results for the English-only (ENG), Russian-only (RU), and Bilingual (BIL) fine-tuned models.}
\small
\begin{tabular}{l c c c c}
\hline
\textbf{Model} & \textbf{\makecell{AI\\F1}} & \textbf{\makecell{Human\\F1}} & \textbf{\makecell{Mean\\Acc.}} & \textbf{\makecell{TPR@\\1\%FPR}} \\
\hline
ENG Mistral-7B & 0.986 & 0.979 & 0.985 & 0.979 \\
ENG mmBERT     & 0.972 & 0.958 & 0.969 & 0.941 \\
\hline
RU Mistral-7B  & 0.986 & 0.980 & 0.984 & 0.978 \\
RU mmBERT      & 0.970 & 0.958 & 0.966 & 0.924 \\
\hline
BIL Mistral-7B & 0.986 & 0.980 & 0.985 & 0.979 \\
BIL mmBERT     & 0.969 & 0.955 & 0.966 & 0.930 \\
\hline
\end{tabular}
\label{tab:classification_results}
\end{table}

\begin{table}[htbp!]
\centering
\caption{AI interval detection results for the GigaCheck (DN-DAB-DETR) model with frozen Mistral-7B or mmBERT backbones.}
\label{tab:detection_res}
\small
\begin{tabular}{l c c}
\hline
    \textbf{Model} & {\textbf{\makecell{mAP@0.5}}} & {\textbf{\makecell{mAP@0.5:0.95}}} \\
\hline
    ENG Mistral-7B & 0.8749 & 0.7555 \\
    ENG mmBERT & 0.8645 & 0.7494 \\
    \hline
    RU Mistral-7B & 0.8928 & 0.7839 \\
    RU mmBERT & 0.8758 & 0.7629 \\
    \hline
    BIL Mistral-7B & 0.8976 & 0.7921 \\
    BIL mmBERT & 0.8867 & 0.7864 \\
\hline
\end{tabular}
\end{table}

\subsection{Generalization to Unseen Generators}
\label{subsec:logo_experiments}

To assess cross-architecture generalization, we performed \textbf{Leave-One-Family-Out (LOFO)} experiments: for each run we removed an entire provider family (all variants and sizes) from training and evaluated a bilingual \texttt{mmBERT-base} model on the held-out family's target generator. We held out three families chosen as independent architectural lineages spanning the full size range of our open-source pool: \textbf{Mistral} (target \textit{Ministral-8B}), \textbf{DeepSeek} (target \textit{DeepSeek-R1-Distill-Qwen-32B}), and \textbf{Llama} (target \textit{Llama-3.3-70B}). Removing the whole family, not just the target, prevents same-family variants from leaking signatures into training. Table~\ref{tab:logo_results} shows negligible drops ($<1.5\%$ AI F1) vs.\ the fully-trained baseline, indicating that the dataset's diversity enables learning universal AI markers rather than family-specific artifacts. A complementary population-level analysis across all 14 families in the test set (Section~\ref{sec:error_analysis}; Appendix~\ref{subsec:err_family}) shows the dominant difficulty axis is generator scale, not family identity.

\subsection{Detection Performance by Generation Strategy}
\label{subsec:ablation_strategy}

To verify the complexity of different collaboration scenarios, we evaluated detection baselines across mixed-text subsets (Table~\ref{tab:ablation_strategy}). While models achieve high accuracy on \textit{Gap-Filling} and \textit{Continuation}, performance drops significantly on \textit{Manual Editing}. This confirms that subtle human edits disrupt statistical patterns, validating the importance of including manually edited examples to train robust models capable of detecting sophisticated, ``humanized'' AI content.

\begin{table}[htbp!]
\centering
\caption{Leave-One-Family-Out (LOFO) generalization. Each row holds out the \emph{entire target family} (all variants and sizes) from training and evaluates on its target generator (Human + target only); \textbf{Baseline} = full training. Drops are negligible.}
\label{tab:logo_results}
\resizebox{\columnwidth}{!}{%
\begin{tabular}{l l c c c c}
\hline
\textbf{Held-out family (target)} & \textbf{Config} & \textbf{AI F1} & \textbf{Human F1} & \textbf{Mean Acc} & \textbf{TPR@1\%FPR} \\
\hline
\multirow{2}{*}{\makecell[l]{DeepSeek\\(DeepSeek-R1-Distill, 32B)}}
 & Baseline & 0.839 & 0.987 & 0.968 & 0.941 \\
 & LOFO     & 0.828 & 0.986 & 0.968 & 0.947 \\
\hline
\multirow{2}{*}{\makecell[l]{Mistral\\(Ministral-8B-Instruct, 8B)}}
 & Baseline & 0.811 & 0.985 & 0.938 & 0.859 \\
 & LOFO     & 0.809 & 0.985 & 0.935 & 0.846 \\
\hline
\multirow{2}{*}{\makecell[l]{Llama\\(Llama-3.3-Instruct, 70B)}}
 & Baseline & 0.872 & 0.987 & 0.968 & 0.934 \\
 & LOFO     & 0.877 & 0.987 & 0.966 & 0.932 \\
\hline
\end{tabular}%
}
\end{table}

\begin{table}[h]
\centering
\caption{Ablation study of detection performance across different mixed-authorship text generation strategies. The \textbf{Manual Editing} strategy proves to be the most challenging for both architectures.}
\label{tab:ablation_strategy}
\resizebox{\columnwidth}{!}{%
\begin{tabular}{lcccc}
\toprule
\multirow{2}{*}{\textbf{Mixing Strategy}} & \multicolumn{2}{c}{\textbf{BIL mmBERT}} & \multicolumn{2}{c}{\textbf{BIL Mistral7B}} \\
 & \textbf{mAP@0.5} & \textbf{mAP@0.5:0.95} & \textbf{mAP@0.5} & \textbf{mAP@0.5:0.95} \\ \midrule
Gap-Filling & 0.826 & 0.654 & 0.838 & 0.660 \\
Continuation & 0.788 & 0.613 & 0.855 & 0.692 \\
Manual Editing & 0.656 & 0.458 & 0.733 & 0.532 \\ \bottomrule
\end{tabular}%
}
\end{table}

\section{Error Analysis}
\label{sec:error_analysis}

\begin{figure}[ht!]
    \centering
    \includegraphics[width=\columnwidth]{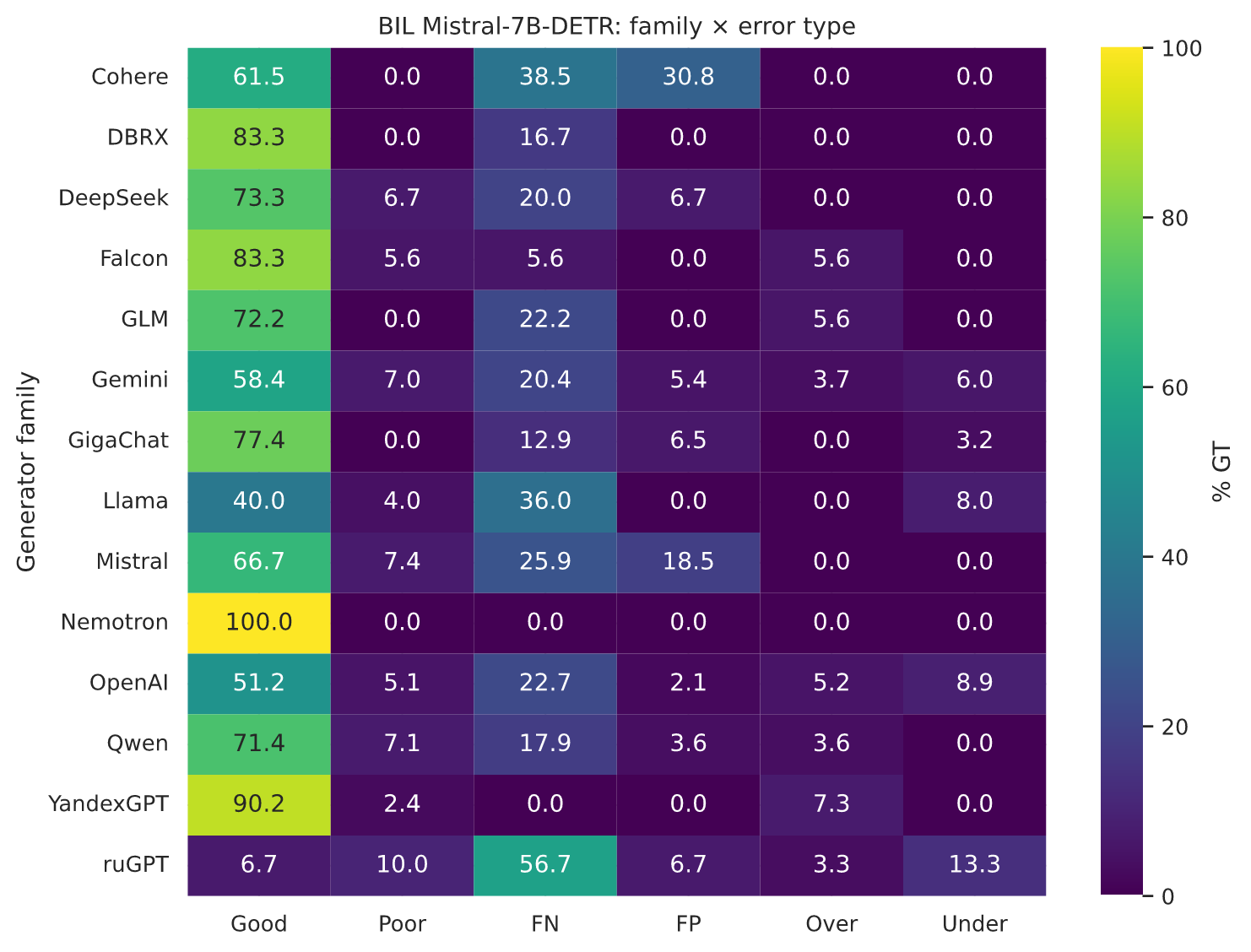}
    \caption{Per-family error-type distribution for BIL Mistral-7B-DETR (\% of within-family GT intervals per category). \texttt{ruGPT} and early Llama are hardest; OpenAI/Gemini are easiest. BIL mmBERT-DETR analysis: Appendix Figure~\ref{fig:err_family_heatmap_mmbert}.}
    \label{fig:err_family_heatmap}
\end{figure}

We categorize each predicted/GT interval pair into six error types: \textit{good} and \textit{poor matches} (1-to-1, separated by an IoU threshold of 0.8), \textit{false negatives}, \textit{false positives}, \textit{over-} and \textit{under-segmentation} (definitions in Appendix~\ref{sec:appendix_errors}). Manual editing yields substantially fewer good matches than the fill-gaps regime (48\%/56\% for BIL Mistral-7B-DETR; 36\%/60\% for BIL mmBERT-DETR), confirming that human edits disrupt detectable patterns (Appendix Figure~\ref{fig:err_taxonomy}).

Slicing errors by generator metadata, the dominant failure mode is on under-represented small open-source models ($\leq$8B; \texttt{llama-7b}, \texttt{ruGPT}), which reach interval-level F1 of 0.29 - 0.50 vs.\ 0.83 - 0.90 for frontier closed-source generators (Figure~\ref{fig:err_family_heatmap}). The apparent positive recency trend ($+0.11$ F1/year, $p = 2.4\!\times\!10^{-4}$ for BIL Mistral-7B-DETR) is an artifact of this same confound: the oldest generators in our test set are also the smallest open-source ones (Appendix~\ref{subsec:err_recency}). Text length is the strongest text-side factor: recall drops from 0.84 ($\leq$200 words) to 0.15 ($\geq$2000 words) while precision stays $\sim$0.95, so long contexts produce silent misses rather than false positives. The two detector backbones agree only moderately on per-generator F1 (Spearman $\rho=0.40$, $p=0.04$, $n=28$), indicating that detection difficulty has both data-intrinsic and architecture-specific components. Zero-shot evaluation on RoFT~\citep{dugan2023real} and RoFT-ChatGPT~\citep{kushnareva2023ai} (Appendix~\ref{subsec:transferability}) reinforces this: transfer is stronger to generators overlapping LLMTrace's training distribution (GPT-3.5) and weaker to unseen architectures (GPT-2, CTRL). Full breakdowns by family, size, recency, Arena Elo proxy, text properties, and cross-detector statistics are in Appendix~\ref{sec:appendix_errors}.

\section{Conclusion}
\label{sec:conclusion}

We presented \texttt{\DatasetName{}}, a large-scale benchmark addressing critical gaps in detecting modern LLMs and mixed-authorship content. Instantiated for English and Russian via an extensible, language-agnostic curation pipeline, and spanning 38 diverse generators across nine domains, it supports both binary classification and precise interval detection, with extensive analysis confirming the high quality and structural realism of the generated texts. By releasing this resource and establishing strong baselines, we aim to facilitate the development of next-generation detectors capable of handling nuanced, real-world AI usage.

\section*{Limitations}
\label{sec:limitations}

While \texttt{\DatasetName{}} establishes a comprehensive benchmark, we acknowledge specific aspects that outline directions for future research.

First, while our Leave-One-Family-Out (LOFO) experiments (Section~\ref{subsec:logo_experiments}) hold out three entire provider families (Mistral, DeepSeek, Llama) spanning the full parameter range, and are complemented by a population-level per-family error analysis across all 14 families in the test set (Appendix~\ref{subsec:err_family}), running the controlled LOFO protocol exhaustively across every family represented among our 38 generators would provide a fuller picture and remains a valuable direction.
Second, regarding evaluation, we established a robust baseline using a detection architecture (DN-DAB-DETR) and standard mAP metrics. Future work could broaden this by benchmarking against sequence-labeling methods and exploring richer span-level metrics (e.g., boundary distance errors) to capture finer localization nuances.
Third, the released dataset currently instantiates the curation pipeline for English and Russian; while the pipeline is language-agnostic by design, extending coverage to typologically distinct languages (e.g., CJK, Arabic, Hindi) and to additional language-specific generator families is a natural next step that would further validate the framework's extensibility.
Finally, the significant performance drop on manually edited texts indicates that identifying subtle human modifications remains an open challenge, and the error analysis (Section~\ref{sec:error_analysis}) further shows that detection quality on legacy and under-represented small open-source generators is markedly lower than on frontier closed-source models; broadening the training distribution towards such generators is a clear next step.

\section*{Ethical Considerations}
 
\textbf{Human Labor.}
All editors who contributed to the mixed-text corpus participated voluntarily, were fully informed of the task, and were compensated at rates significantly above local minimum wage.
 
\textbf{Data Privacy.}
Human-authored texts were sourced from established open-source collections. Standard automated filtering was applied, but incidental PII may remain in web-crawled data; researchers should exercise caution when analyzing individual samples.
 
\textbf{Dual-Use Risks.}
Detection datasets could theoretically help adversaries train less detectable generators, and imperfect detectors risk false positives in academic or professional settings. We believe the benefit of open, high-quality resources for studying these challenges outweighs the risks of withholding them.

\bibliography{custom}

\newpage
\appendix
\onecolumn

% --- APPENDIX: DATA SOURCES ---

\section{Classification Benchmarks Comparison}
\label{app:classification_benchmarks}

Table~\ref{tab:classification_benchmarks} provides a detailed comparison of \DatasetName{} against 17 existing AI text classification benchmarks along the axes of dataset size, domain coverage, language coverage, number of generators, and prompt-type diversity. This table complements Table~\ref{tab:detection_benchmarks} in the main text, which focuses on detection (interval-localization) benchmarks.

\begin{table*}[h!]
    \centering
    \caption{Statistics of the texts in the classification datasets.}
    \label{tab:classification_benchmarks}
    \small
    \begin{tabular}{l c c c c c}
        \hline
        \textbf{Dataset} & \textbf{\makecell[c]{Dataset Size}} & \textbf{\# Domains} & \textbf{\# Lang} & \textbf{\# Gen} & \textbf{\makecell[c]{Various prompt types}} \\
        \hline
        MAGE \citep{li2023mage} & 447.7k & 7 & 1 & 27 & \ding{51} \\
        MGTBench \citep{he2024mgtbench} & 21k & 3 & 1 & 6 & \ding{55} \\
        BUST \citep{cornelius2024bust} & 25.2k & 4 & 1 & 7 & \ding{51} \\
        RAID \citep{dugan2024raid} & 6.2M & 8 & 1 & 11 & \ding{55} \\
        DetectRL \citep{wu2024detectrl} & 235.2k & 4 & 1 & 4 & \ding{51} \\
        HC3 \citep{guo2023close} & 125.2k & 1 & 2 & 1 & \ding{55} \\
        MULTITuDE \citep{macko2023multitude} & 74.1k & 1 & 11 & 8 & \ding{55} \\
        M4GT-Bench \citep{wang2024m4gt} & 152.7k & 8 & 8 & 7 & \ding{51} \\
        MultiSocial \citep{macko2024multisocial} & 472k & 1 & 22 & 7 & \ding{55} \\
        RuATD \citep{shamardina2022findings} & 215k & 6 & 1 & 13 & \ding{51} \\
        WETBench \citep{quaremba2025wetbench} & 101.9k & 1 & 3 & 4 & \ding{51} \\
        Peer Review Detection \citep{yu2025your} & 789k & 1 & 1 & 5 & \ding{55} \\
        GEDE \citep{gehring2025assessing} & 13.4k & 1 & 1 & 2 & \ding{51} \\
        ESPERANTO \citep{ayoobi2024esperanto} & 720k & 4 & 1 & 8 & \ding{51} \\
        SHIELD \citep{ayoobi2025beyond} & 700k & 7 & 1 & 7 & \ding{55} \\
        Beemo \citep{artemova2024beemo} & 19.6k & several & 1 & 10 & \ding{51} \\
        MixSet \citep{zhang2024llm} & 3.6k & 6 & 1 & 2 & \ding{51} \\
        \hline
        \textbf{\DatasetName{} (ours)} & 589,086 & 9 & 2 & 38 & \ding{51} \\
        \hline
    \end{tabular}
\end{table*}

\section{Data Sources}
\label{app:sources}

The tables \ref{tab:ru_sources} and \ref{tab:en_sources} list the open-source datasets and public sources used to construct the human-authored corpora for the Russian and English languages.

\begin{table}[h!]
\centering
\caption{Data sources for the Russian corpus.}
\label{tab:ru_sources}
\small
\begin{tabular}{ll}
\hline
\textbf{Dataset Name} & \textbf{Source / Link} \\ \hline
SiberianDatasetXL & \url{https://huggingface.co/datasets/SiberiaSoft/SiberianDatasetXL} \\
medical\_qa\_ru\_data & \url{https://huggingface.co/datasets/blinoff/medical_qa_ru_data} \\
mailruQA-big & \url{https://huggingface.co/datasets/Den4ikAI/mailruQA-big} \\
mfaq & \url{https://huggingface.co/datasets/clips/mfaq} \\
miracl-ru-corpus & \url{https://huggingface.co/datasets/Cohere/miracl-ru-corpus-22-12} \\
wiki\_lingua & \url{https://huggingface.co/datasets/GEM/wiki_lingua} \\
taiga & \url{https://huggingface.co/datasets/danasone/taiga} \\
IlyaGusev's datasets & \url{https://huggingface.co/IlyaGusev/datasets} \\
RussianSuperGLUE & \url{https://github.com/RussianNLP/RussianSuperGLUE} \\
xlsum & \url{https://huggingface.co/datasets/csebuetnlp/xlsum} \\
mlsum & \url{https://huggingface.co/datasets/reciTAL/mlsum} \\
xquad & \url{https://huggingface.co/datasets/google/xquad} \\
NestQuad & \url{https://huggingface.co/datasets/NeSTudio/NestQuad} \\
sberquad & \url{https://huggingface.co/datasets/kuznetsoffandrey/sberquad} \\
ru\_sentiment\_dataset & \url{https://huggingface.co/datasets/MonoHime/ru_sentiment_dataset} \\
restaurants\_reviews & \url{https://huggingface.co/datasets/blinoff/restaurants_reviews} \\
cedr & \url{https://huggingface.co/datasets/sagteam/cedr_v1} \\ \hline
\end{tabular}
\end{table}

\begin{table}[!htbp]
\centering
\caption{Data sources for the English corpus.}
\small
\begin{tabularx}{\textwidth}{l >{\raggedright\arraybackslash}X}
\hline
\textbf{Dataset Name / Source} & \textbf{Link / Reference} \\ \hline
medium-articles & https://www.kaggle.com/datasets/fabiochiusano/medium-articles \\
wiki-22-12 & https://huggingface.co/datasets/Cohere/wikipedia-22-12 \\
cnn-dailymail & https://www.kaggle.com/datasets/gowrishankarp/newspaper-text-summarization-cnn-dailymail \\
plot-synopses & https://www.kaggle.com/datasets/cryptexcode/mpst-movie-plot-synopses-with-tags \\
foundation-poems & https://www.kaggle.com/datasets/tgdivy/poetry-foundation-poems \\
amazon-questions & https://www.kaggle.com/datasets/praneshmukhopadhyay/amazon-questionanswer-dataset \\
yahoo-answers & https://www.kaggle.com/datasets/yacharki/yahoo-answers-10-categories-for-nlp-csv \\
amazon-reviews & https://www.kaggle.com/datasets/kritanjalijain/amazon-reviews \\
sentiment-tweets & https://www.kaggle.com/datasets/tariqsays/sentiment-dataset-with-1-million-tweets \\
ask-reddit & https://www.kaggle.com/datasets/gpreda/ask-reddit \\
arxiv-abstracts & https://www.kaggle.com/datasets/spsayakpaul/arxiv-paper-abstracts \\
Common Crawl & https://commoncrawl.org/ \\
SciXGen & \cite{chen2021scixgen} \\ 
XSum & \cite{narayan2018don} \\ 
TLDR\_news & \url{https://huggingface.co/datasets/JulesBelveze/TLDR_news} \\
Reddit WritingPrompts & \cite{fan2018hierarchical} \\ 
ROCStories Corpora & \cite{mostafazadeh2016corpus} \\ 
Yelp dataset & \cite{zhang2015character} \\ 
/r/ChangeMyView (CMV) & \cite{tan2016winning} \\ 
News Outlets & CNN, Washington Post, New York Times \\
Academic Sources & arXiv (CS, physics), Springer's SSRN (HHS) \\ 
\hline
\end{tabularx}
\label{tab:en_sources}
\end{table}

% --- APPENDIX: Generators ---
\section{Used generation models}
\label{app:generators}

The table~\ref{tab:llms} lists the open-source and proprietary Large Language Models (LLMs) used to construct the AI-written corpora for the Russian and English languages.

\clearpage
\begin{table*}[h!]
    \centering
    \caption{List of LLMs used for AI text generation.}
    \label{tab:llms}
    \small
    \begin{tabularx}{\textwidth}{l c c >{\raggedright\arraybackslash}X}
    \hline
        \textbf{Model} & \textbf{Languages} & \textbf{Proprietary} & \textbf{Source} \\
    \hline
    \multicolumn{4}{l}{\textit{Open-Source Models}} \\ % Use 'l' for left alignment of the subtitle
    \hline
        AI21-Jamba-Mini-1.5 & RU & No & https://huggingface.co/ai21labs/AI21-Jamba-Mini-1.5 \\
        c4ai-command-r-08-2024 & RU+EN & No & https://huggingface.co/CohereLabs/c4ai-command-r-08-2024 \\
        dbrx-instruct & RU+EN & No & https://huggingface.co/databricks/dbrx-instruct \\
        DeepSeek-R1-Distill-Qwen-32B & RU+EN & No & https://huggingface.co/deepseek-ai/DeepSeek-R1-Distill-Qwen-32B \\
        Falcon3-10B-Instruct & EN & No & https://huggingface.co/tiiuae/Falcon3-10B-Instruct \\
        FRED-T5-1.7B & RU & No & https://huggingface.co/ai-forever/FRED-T5-1.7B \\
        gemma-1.1-7b-it & RU & No & https://huggingface.co/google/gemma-1.1-7b-it \\
        gemma-2-27b-it & RU & No & https://huggingface.co/google/gemma-2-27b-it \\
        GLM-4-32B-0414 & EN & No & https://huggingface.co/zai-org/GLM-4-32B-0414 \\
        Jamba-v0.1 & RU & No & https://huggingface.co/ai21labs/Jamba-v0.1 \\
        Llama-3.1-Nemotron-70B-Instruct-HF & EN & No & https://huggingface.co/nvidia/Llama-3.1-Nemotron-70B-Instruct-HF \\
        Llama-3.3-70B-Instruct & RU+EN & No & https://huggingface.co/unsloth/Llama-3.3-70B-Instruct \\
        llama-7b & RU & No & https://huggingface.co/baffo32/decapoda-research-llama-7B-hf \\
        Magistral-Small-2506 & EN & No & https://huggingface.co/mistralai/Magistral-Small-2506 \\
        Meta-Llama-3-8B-Instruct & RU & No & https://huggingface.co/meta-llama/Meta-Llama-3-8B-Instruct \\
        Ministral-8B-Instruct-2410 & RU+EN & No & https://huggingface.co/mistralai/Ministral-8B-Instruct-2410 \\
        Phi-3-medium-128k-instruct & RU & No & https://huggingface.co/microsoft/Phi-3-medium-128k-instruct \\
        Phi-3-mini-128k-instruct & RU & No & https://huggingface.co/microsoft/Phi-3-mini-128k-instruct \\
        Qwen2-7B-Instruct & RU & No & https://huggingface.co/Qwen/Qwen2-7B-Instruct \\
        Qwen2.5-72B-Instruct & RU+EN & No & https://huggingface.co/Qwen/Qwen2.5-72B-Instruct \\
        Qwen3-32B & EN & No & https://huggingface.co/Qwen/Qwen3-32B \\
        QwQ-32B & RU & No & https://huggingface.co/Qwen/QwQ-32B \\
        ruGPT family (rugpt3large, rugpt2large) & RU & No & \cite{zmitrovich2023family} \\
        WizardLM-2-7B & RU & No & https://huggingface.co/dreamgen/WizardLM-2-7B \\
        YandexGPT-5-Lite-8B-instruct & RU & No & https://huggingface.co/yandex/YandexGPT-5-Lite-8B-instruct \\
        Yi-1.5-34B-Chat & RU & No & https://huggingface.co/01-ai/Yi-1.5-34B-Chat \\
    \hline
    \multicolumn{4}{l}{\textit{Proprietary Models}} \\
    \hline
        GigaChat series (Pro, Max) & RU & Yes & https://giga.chat/ \\
        Google Gemini series (2.0/2.5 flash) & RU+EN & Yes & https://gemini.google.com/ \\
        OpenAI GPT series (3.5, 4, 4o, o1, o1-mini, o3) & RU+EN & Yes & https://openai.com/ \\
        YaGPT 2 & RU & Yes & https://ya.ru/ai/gpt \\
    \hline
    \end{tabularx}
\end{table*}
\clearpage

\begin{figure*}[!ht]
    \centering
    \includegraphics[width=0.8\textwidth]{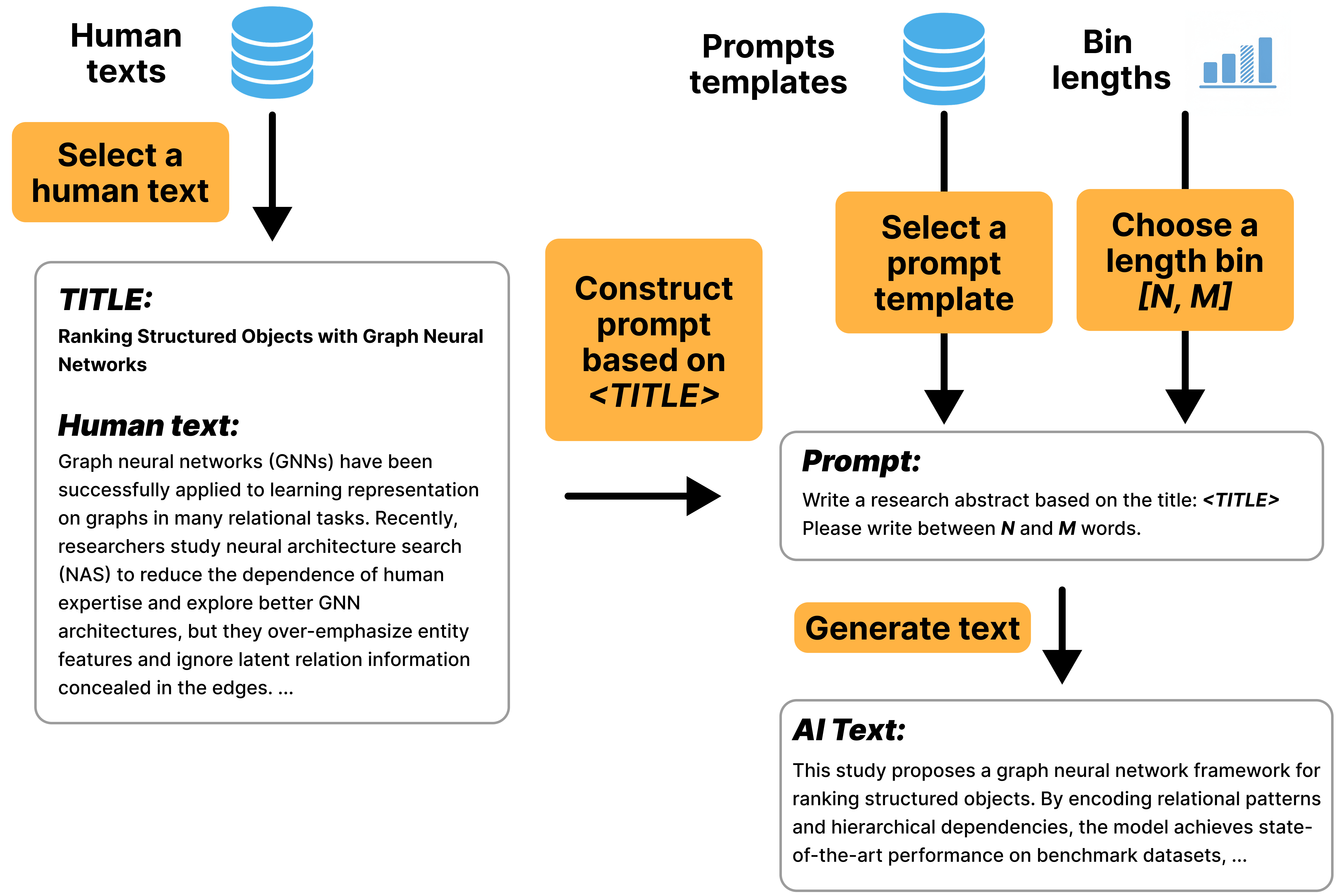}
    \caption{Example of the \textbf{Create} generation pipeline, where a research abstract is generated from a title.}
    \label{fig:prompt_create}
\end{figure*}

\begin{figure*}[h!]
    \centering
    \includegraphics[width=0.8\textwidth]{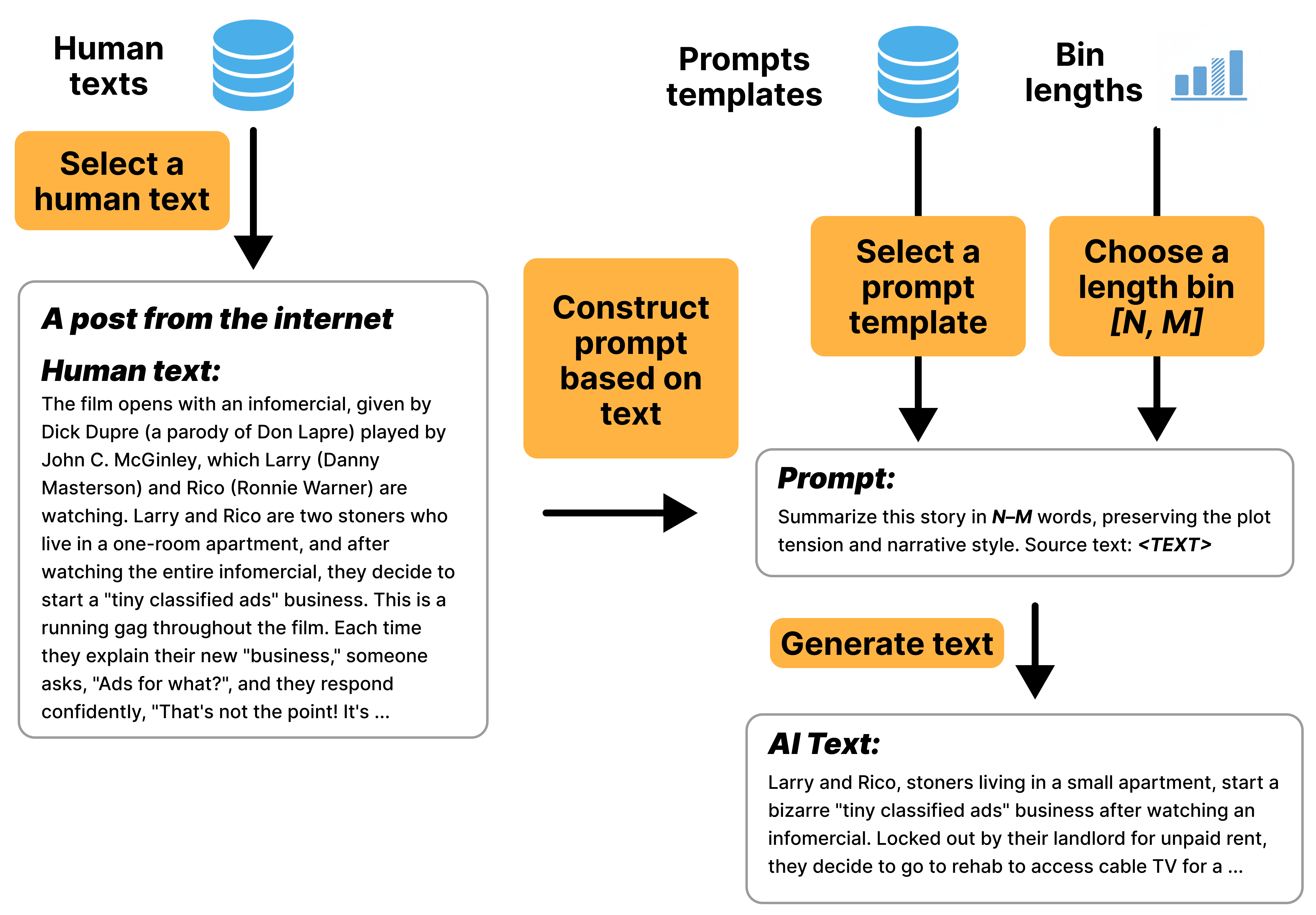}
    \caption{Example of the \textbf{Delete} generation pipeline, where a long post is summarized.}
    \label{fig:prompt_delete}
\end{figure*}

\clearpage

\begin{figure*}[ht!]
    \centering
    \includegraphics[width=0.8\textwidth]{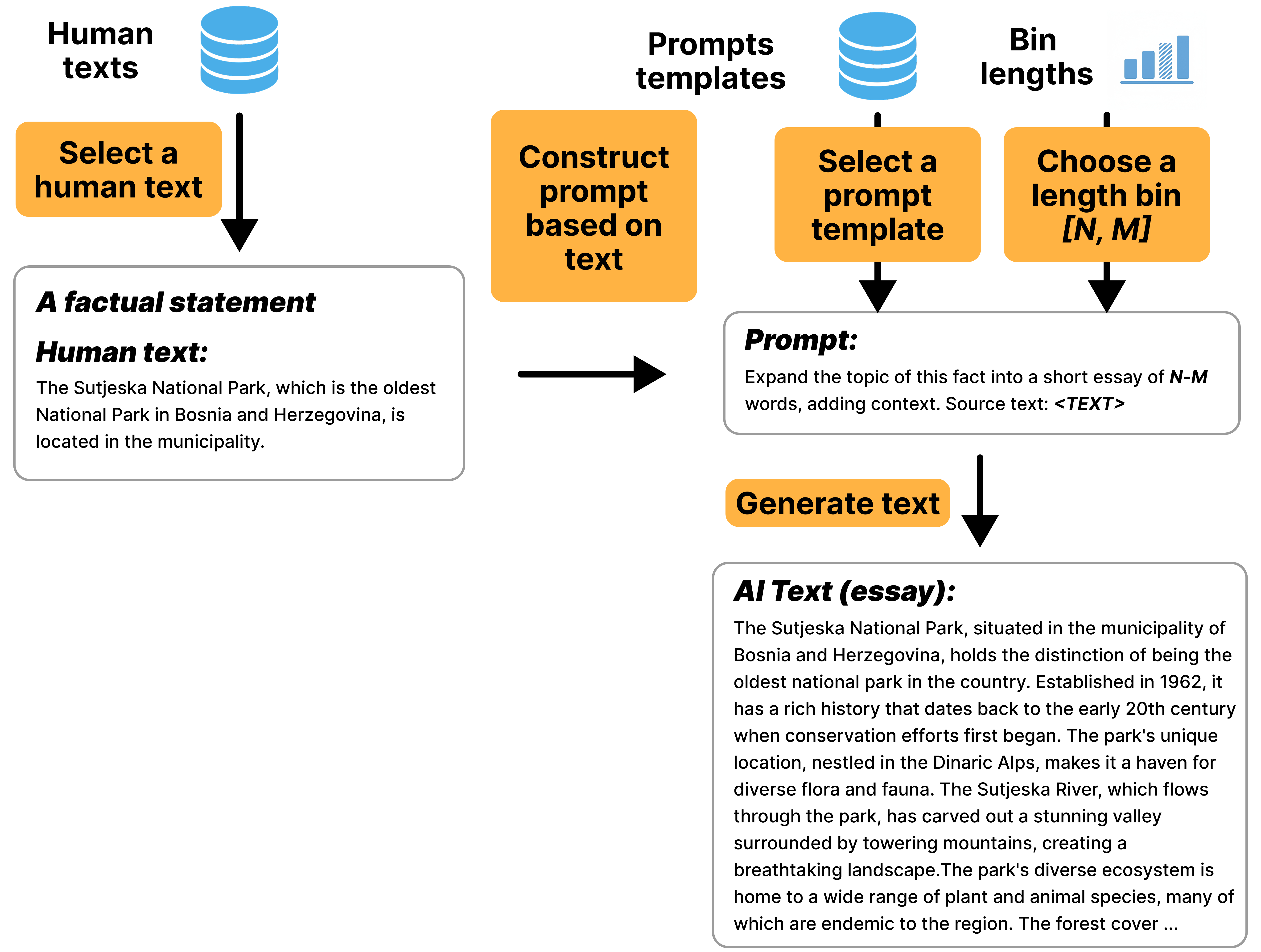}
    \caption{Example of the \textbf{Expand} generation pipeline, where a factual statement is expanded into an essay.}
    \label{fig:prompt_expand}
\end{figure*}

\begin{figure*}[h!]
    \centering
    \includegraphics[width=0.8\textwidth]{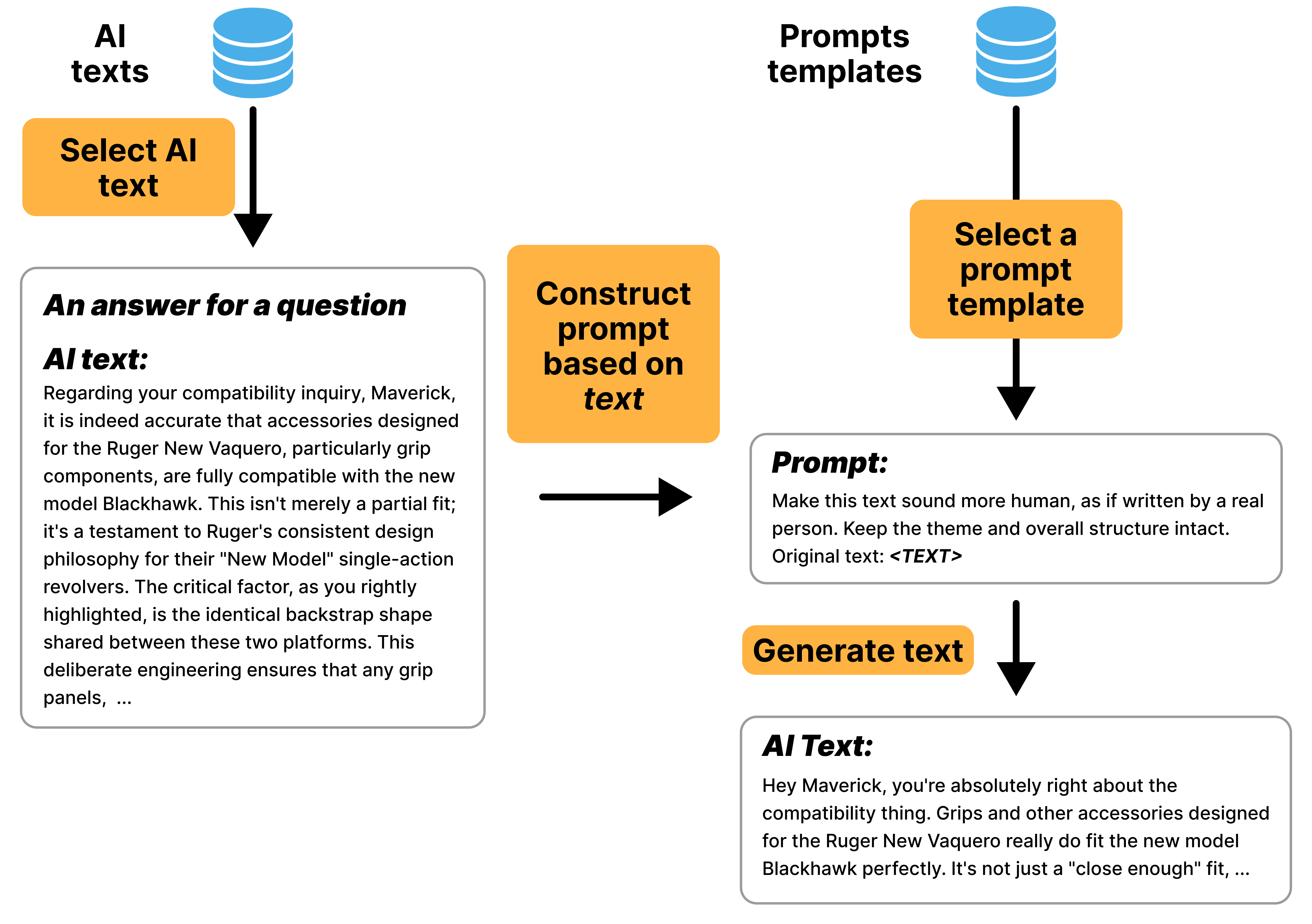}
    \caption{Example of the \textbf{Update} generation pipeline, where an AI-generated text is modified to sound more human.}
    \label{fig:prompt_update}
\end{figure*}

\clearpage

% --- APPENDIX: PROMPT EXAMPLES ---
\section{Prompt Generation Examples}
\label{app:prompt_examples}

This section provides detailed visual diagrams illustrating the four primary categories of prompt templates used to generate the AI text corpus: \textit{Create}, \textit{Delete}, \textit{Expand}, and \textit{Update}. Each figure demonstrates the end-to-end process, from the selection of a source text to the final AI-generated output.

\paragraph{}The \textbf{Create} pipeline (Figure~\ref{fig:prompt_create}) uses a specific attribute from a human text, such as its title, to construct a prompt for generating a new, topically-related AI text within a target length bin.

\paragraph{}The \textbf{Delete} pipeline (Figure~\ref{fig:prompt_delete}) takes an entire human text as input and uses a prompt to generate a condensed, summary version of it, again adhering to a length constraint.

\paragraph{}The \textbf{Expand} pipeline (Figure~\ref{fig:prompt_expand}) operates similarly but with the opposite goal: it uses a prompt to elaborate on a short human text, generating a more detailed and verbose AI text.

\paragraph{}Finally, the \textbf{Update} pipeline, depicted in Figure~\ref{fig:prompt_update}, is unique. It begins with an already-generated AI text and applies a modification prompt, for instance, to "humanize" the writing style or change its tone, to create a more complex and subtle AI artifact. Note that this pipeline does not typically require a length constraint, as the goal is stylistic transformation rather than content resizing.

% --- APPENDIX: STATISTICS ---
\section{Dataset Statistics and Analysis}
\label{app:statistics}

\begin{figure}[h!]
    \centering

    % --- Classification (AI vs Human) ---
    
    \begin{subfigure}{0.48\textwidth}
        \centering
        \includegraphics[width=\linewidth]{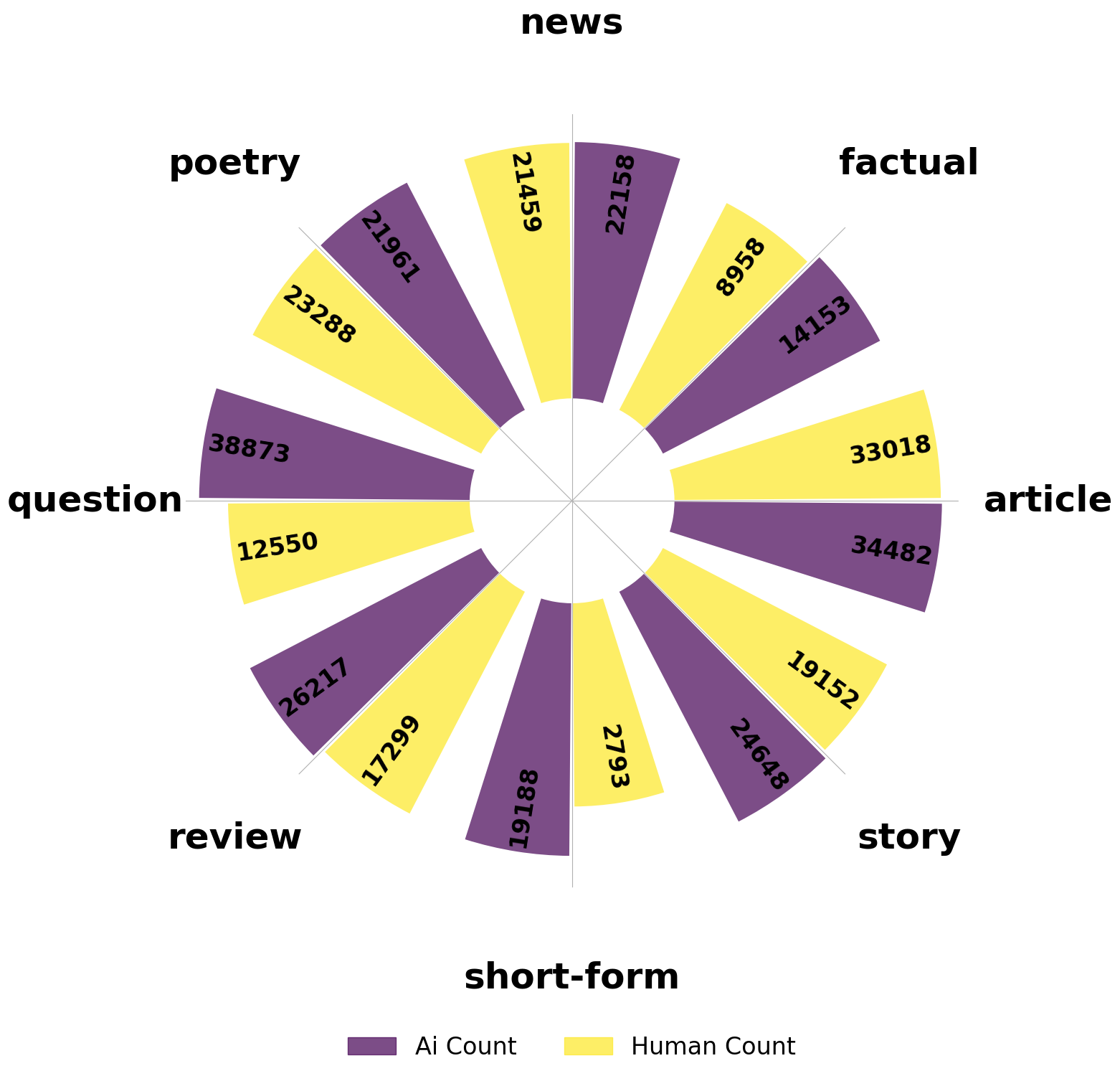}
        \caption{Russian classification dataset}
        \label{fig:rus_class}
    \end{subfigure}
    \hfill 
    \begin{subfigure}{0.48\textwidth}
        \centering
        \includegraphics[width=\linewidth]{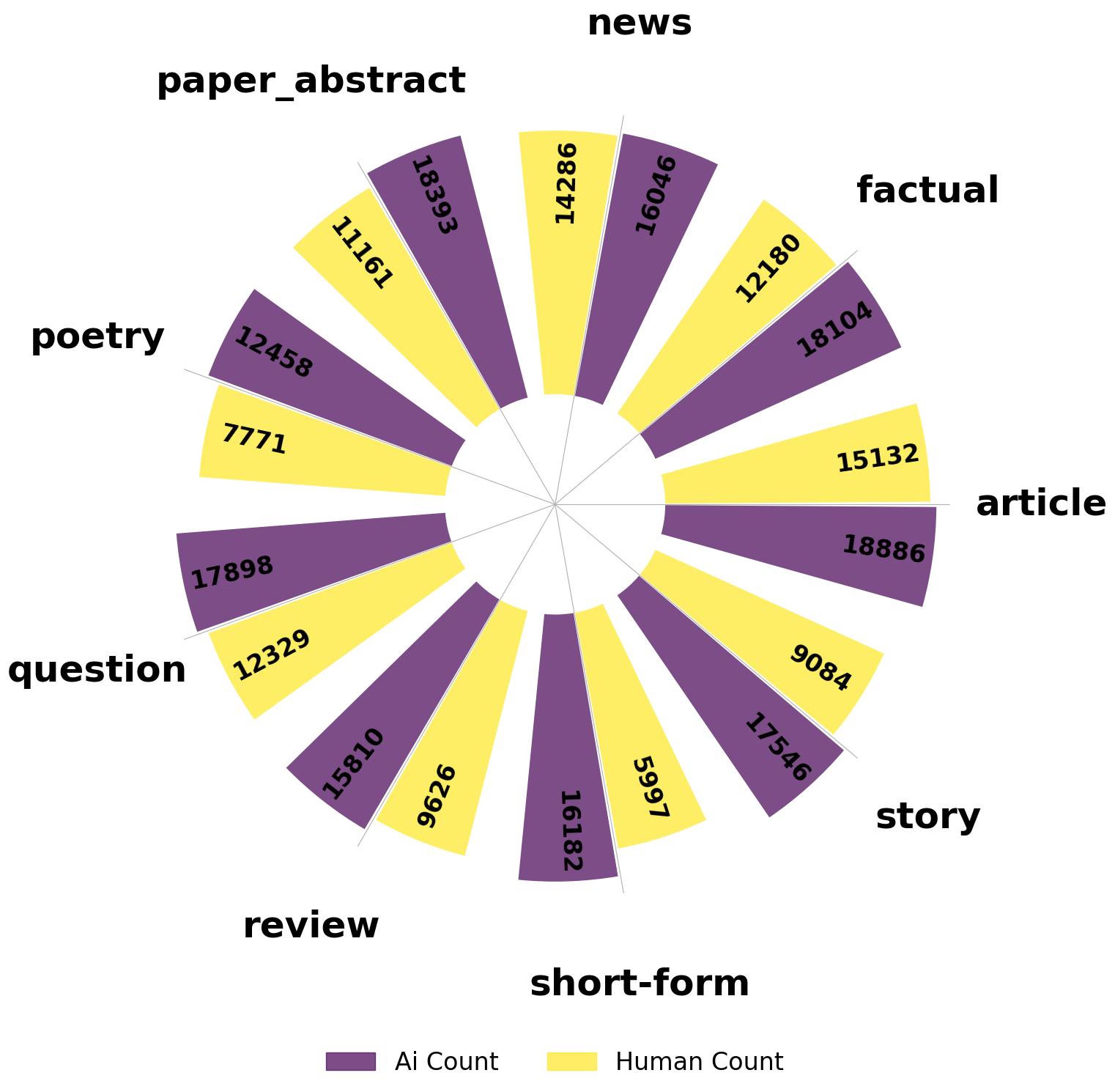}
        \caption{English classification dataset}
        \label{fig:eng_class}
    \end{subfigure}
\caption{Distribution of labels across domains for classification datasets. Each plot shows the number of samples for each label type within a specific domain, for both Russian and English languages.}
    \label{fig:main_dataset_comparison}
\end{figure}

\begin{figure}[h!]
    % --- Detection (AI vs Human vs Mixed) ---

    \begin{subfigure}{0.48\textwidth}
        \centering
        \includegraphics[width=\linewidth]{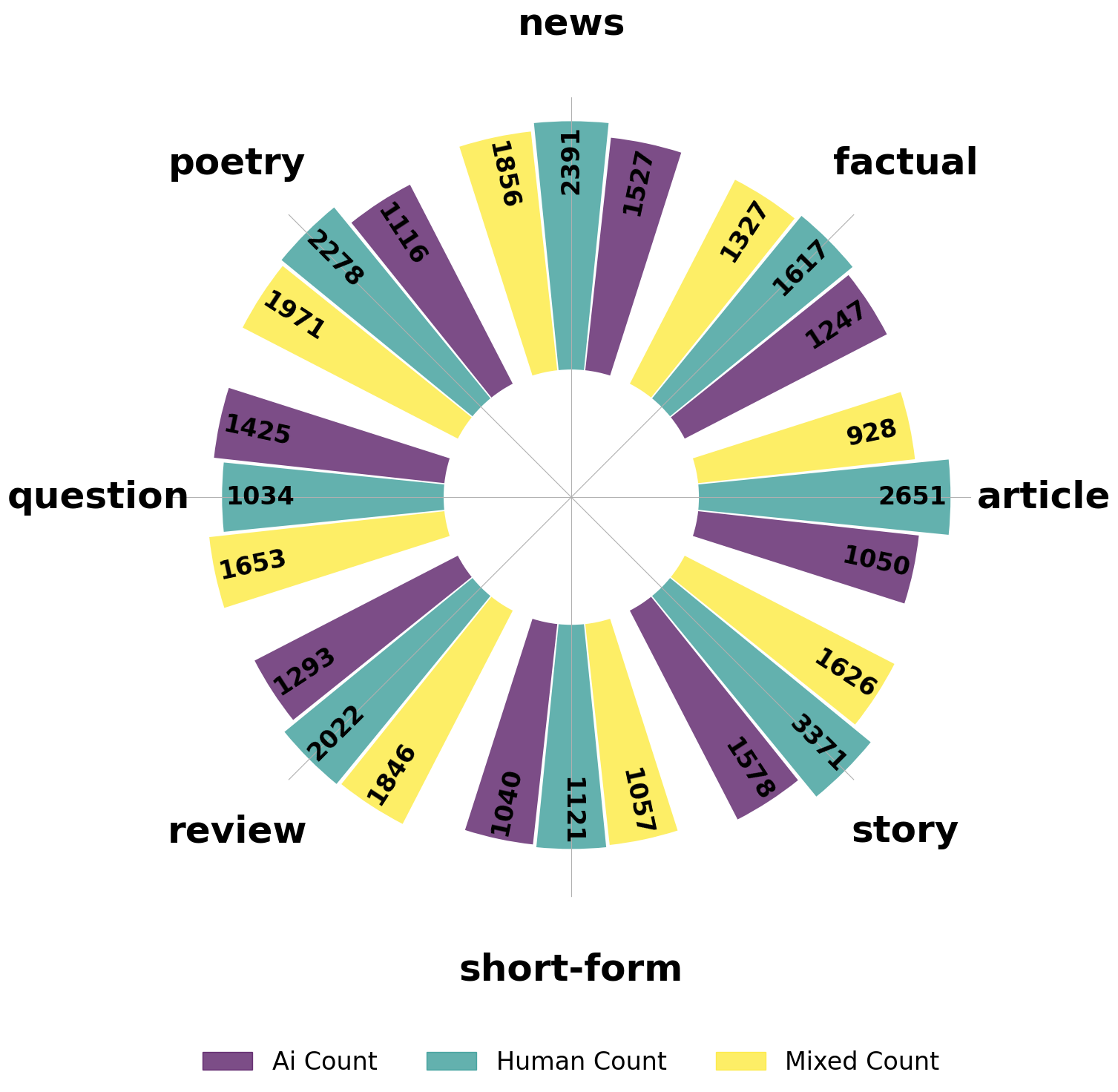}
        \caption{Russian detection dataset}
        \label{fig:rus_detect}
    \end{subfigure}
    \hfill 
    \begin{subfigure}{0.48\textwidth}
        \centering
        \includegraphics[width=\linewidth]{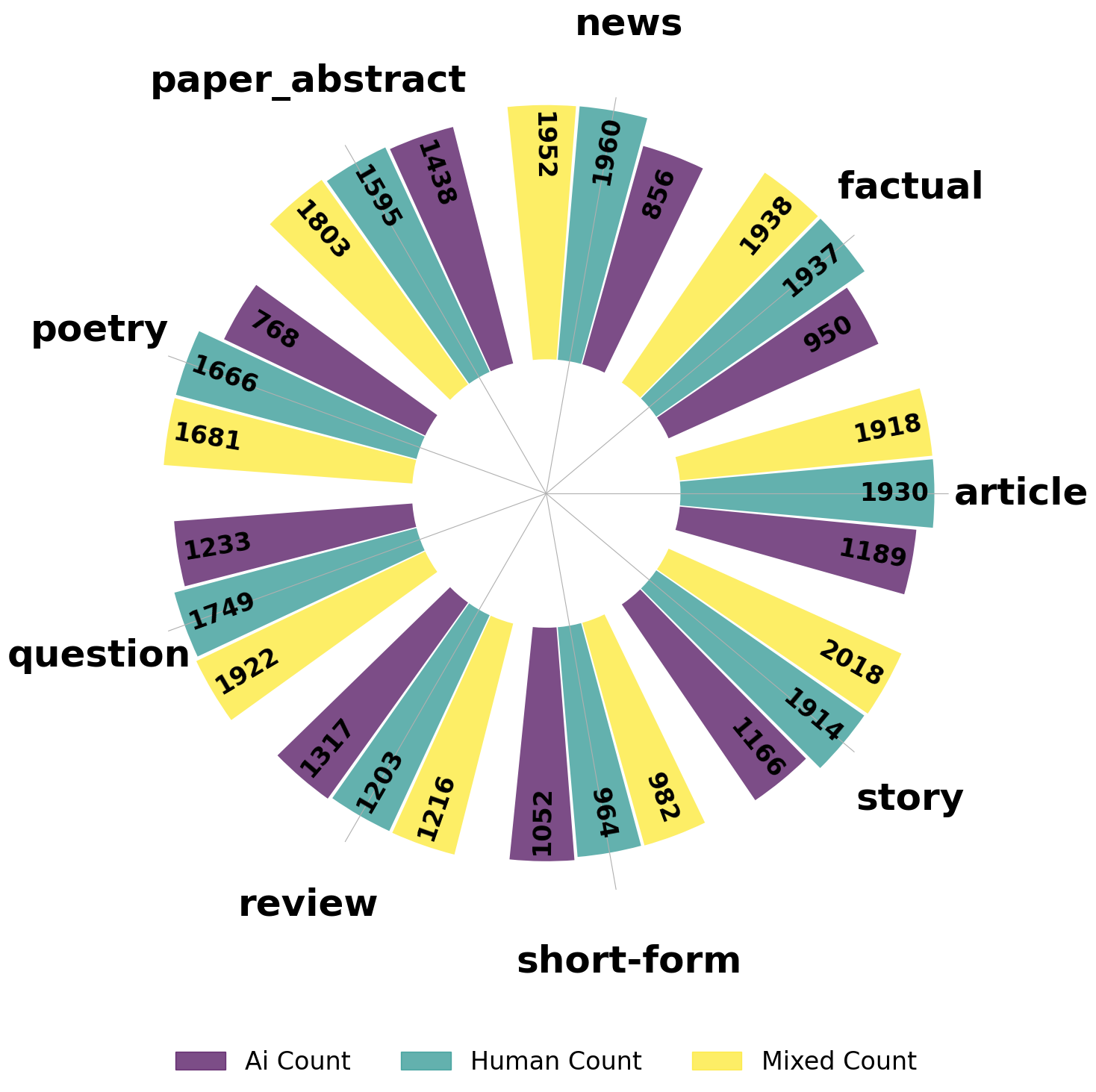}
        \caption{English detection dataset}
        \label{fig:eng_detect}
    \end{subfigure}

    \caption{Distribution of labels across domains for detection datasets. Each plot shows the number of samples for each label type within a specific domain, for both Russian and English languages.}
    \label{fig:main_dataset_comparison_det}
\end{figure}

This section provides a detailed breakdown of the sample distribution across the various domains for both the classification and detection datasets. Figures~\ref{fig:main_dataset_comparison} and \ref{fig:main_dataset_comparison_det} illustrates the count of human, AI, and mixed texts within each domain for the English and Russian corpora. These plots visually confirm the broad and balanced coverage of genres, which is a key feature of our dataset designed to enhance the generalizability of detection models.

\begin{figure*}[!ht]
    \centering
    \begin{subfigure}{0.48\textwidth}
        \centering
        \includegraphics[width=\linewidth]{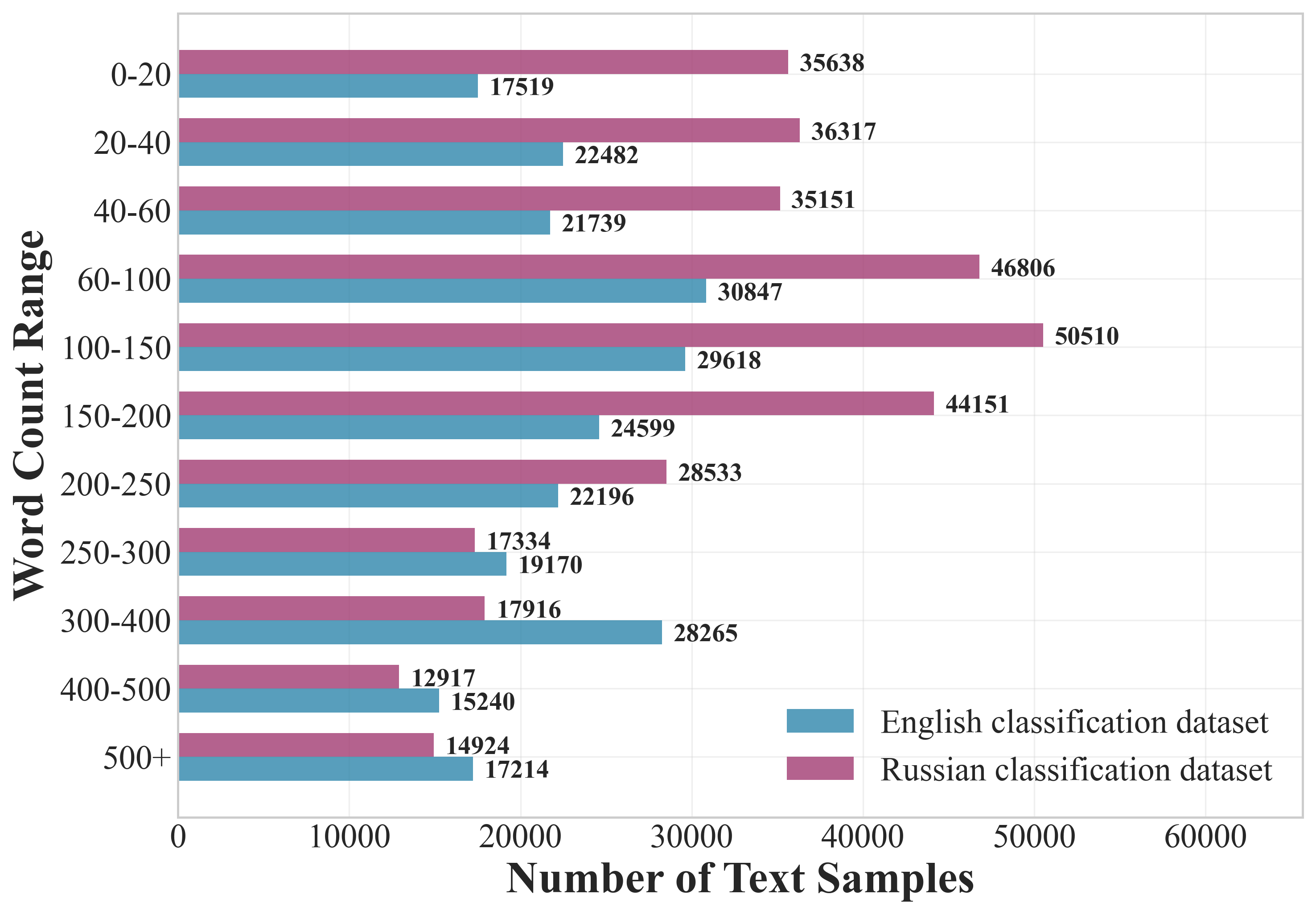}
        \caption{Classification datasets}
        \label{fig:class_text_lens}
    \end{subfigure}
    \hfill
    \begin{subfigure}{0.48\textwidth}
        \centering
        \includegraphics[width=\linewidth]{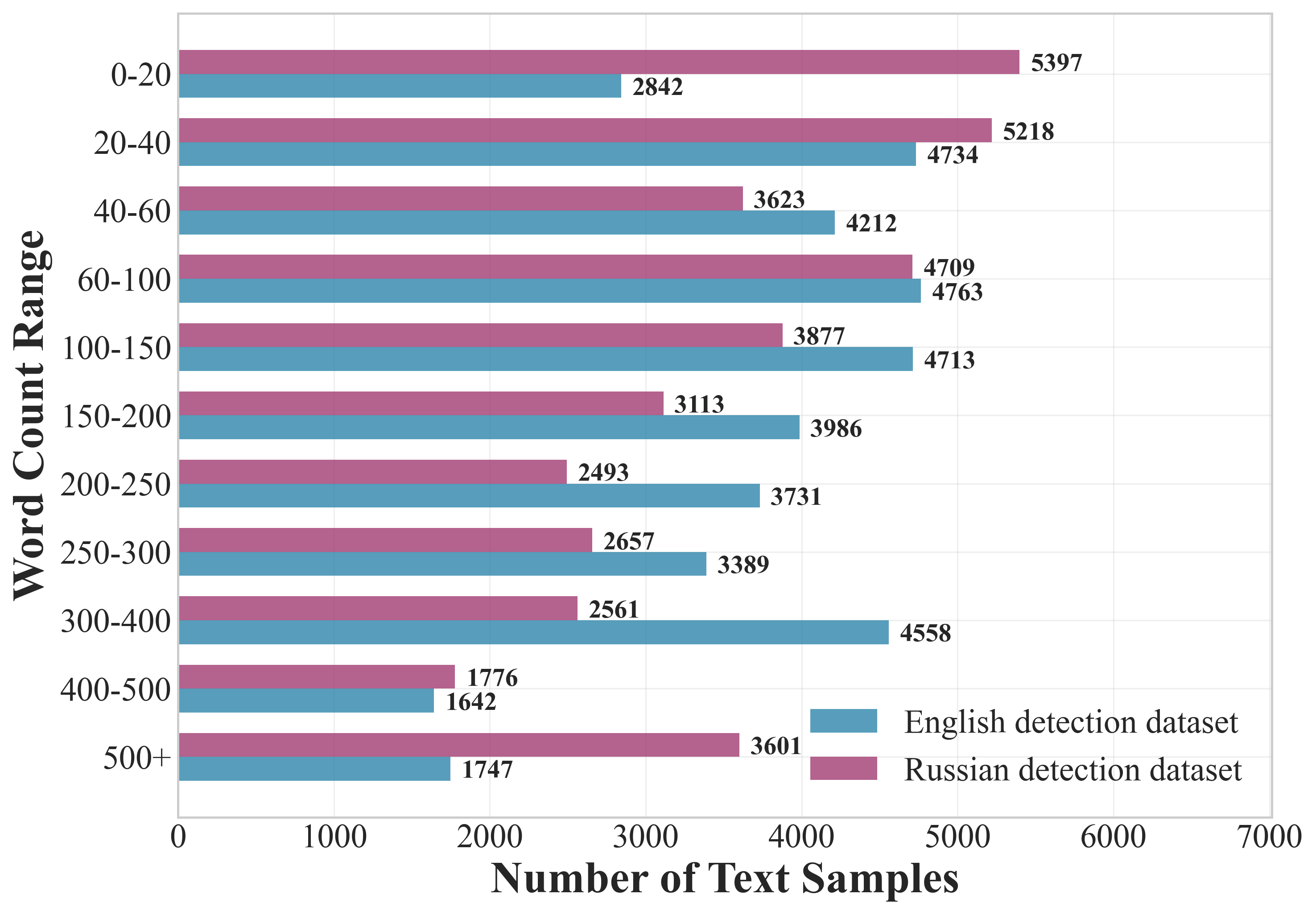}
        \caption{Detection datasets}
        \label{fig:detect_text_lens}
    \end{subfigure}

    \caption{The distribution of text samples across different word count ranges for each dataset.}
    \label{fig:two_len_plots}
\end{figure*}

\section{PHD Distributions}
\label{app:phd_distributions}

\begin{figure}[]
    \centering

    % --- Classification (AI vs Human) ---
    
    \begin{subfigure}{0.48\textwidth}
        \centering
        \includegraphics[width=\linewidth]{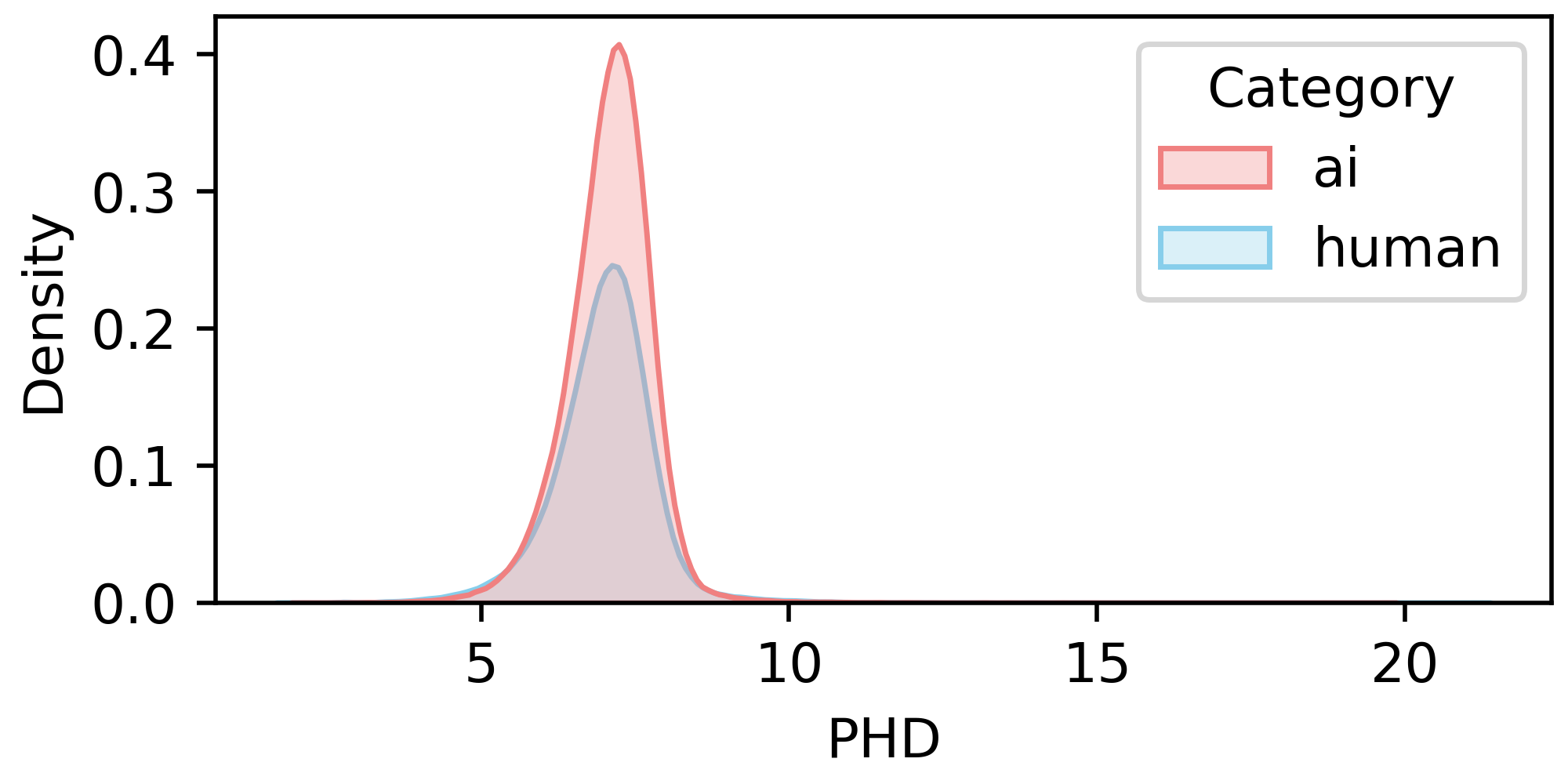}
        \caption{Russian classification dataset}
        \label{fig:phd_rus_class}
    \end{subfigure}
    \hfill 
    \begin{subfigure}{0.48\textwidth}
        \centering
        \includegraphics[width=\linewidth]{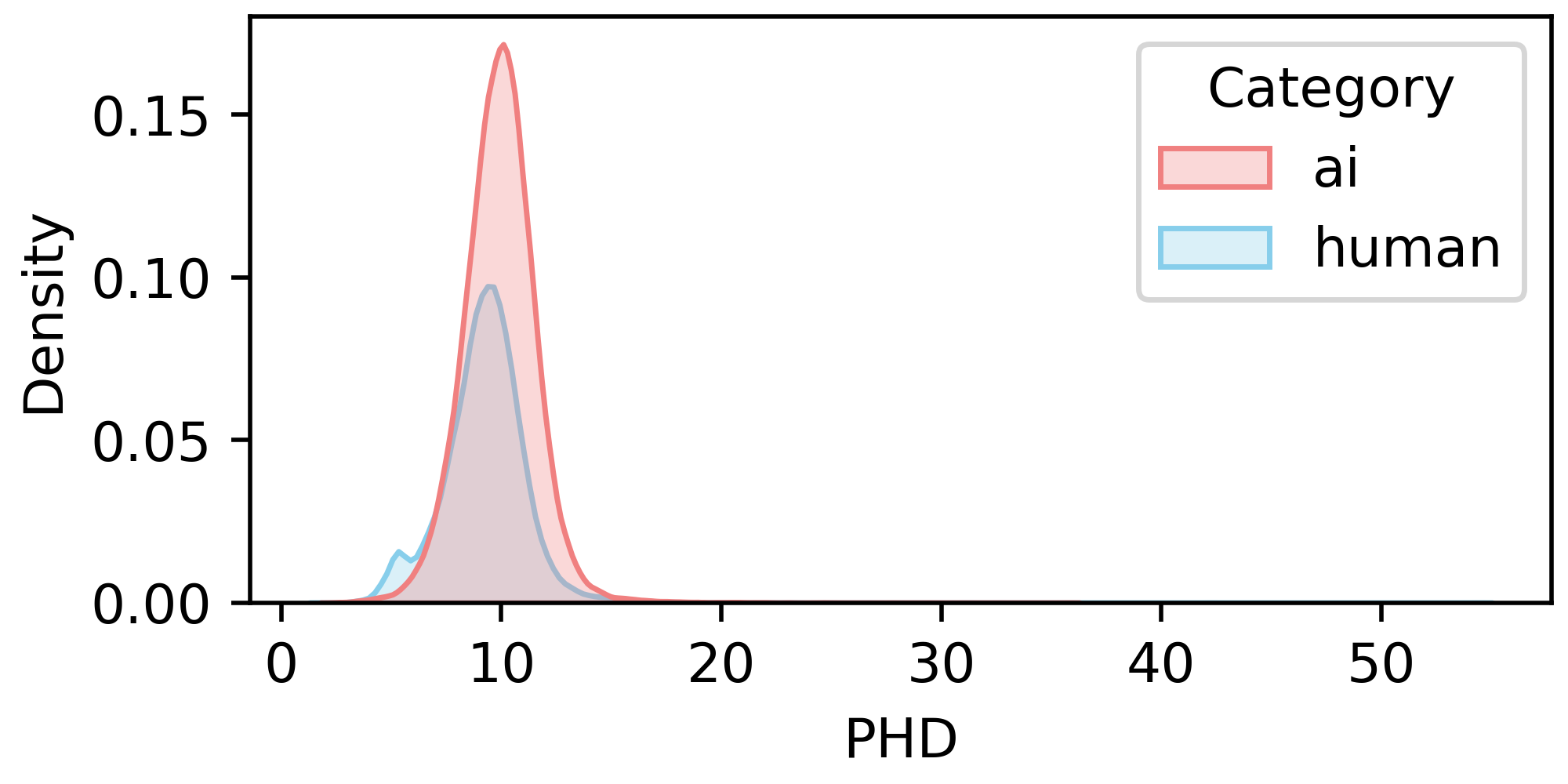}
        \caption{English classification dataset}
        \label{fig:phd_eng_class}
    \end{subfigure}

    % --- Detection (AI vs Human vs Mixed) ---

    \begin{subfigure}{0.48\textwidth}
        \centering
        \includegraphics[width=\linewidth]{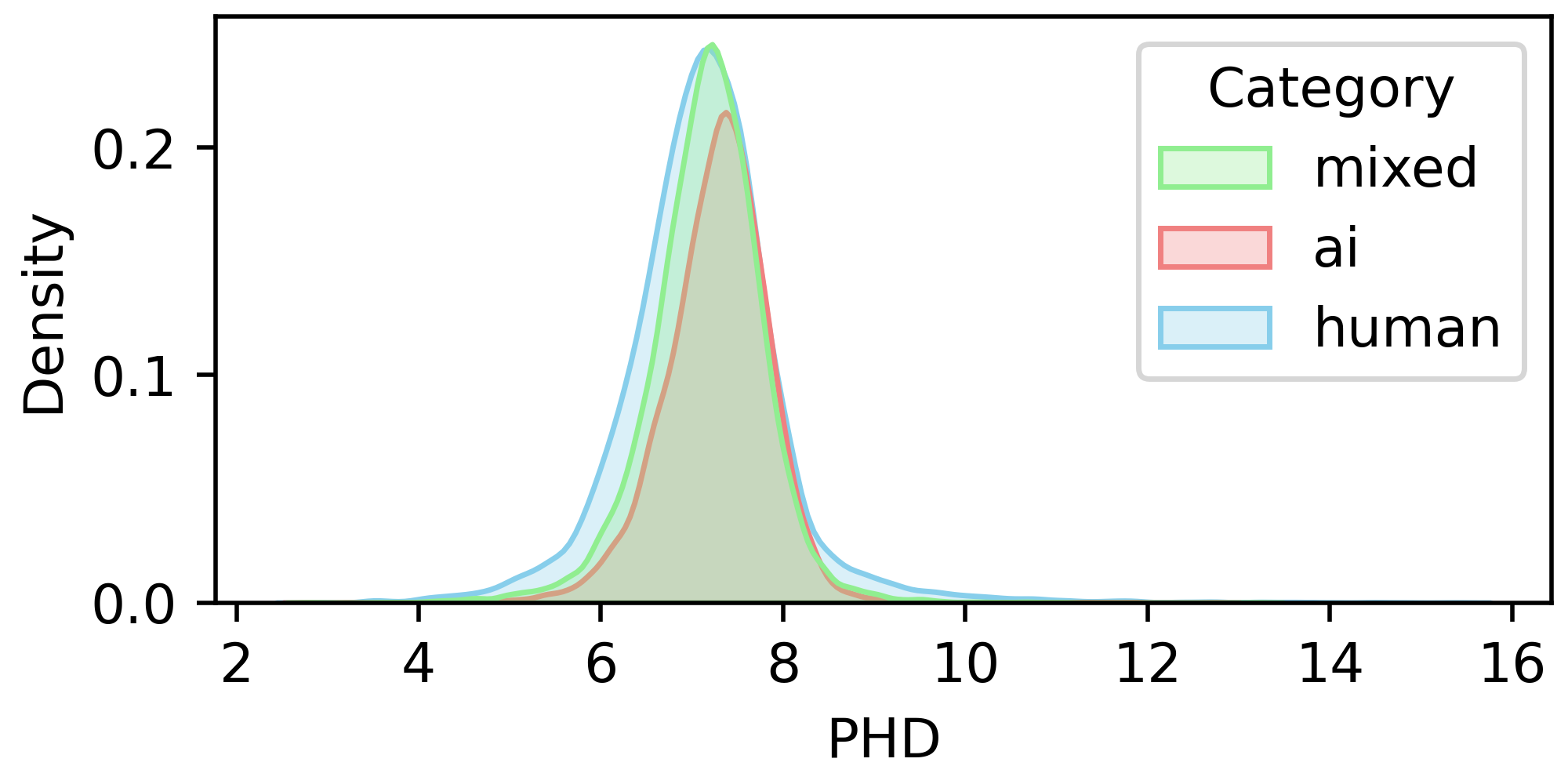}
        \caption{Russian detection dataset}
        \label{fig:phd_rus_detect}
    \end{subfigure}
    \hfill 
    \begin{subfigure}{0.48\textwidth}
        \centering
        \includegraphics[width=\linewidth]{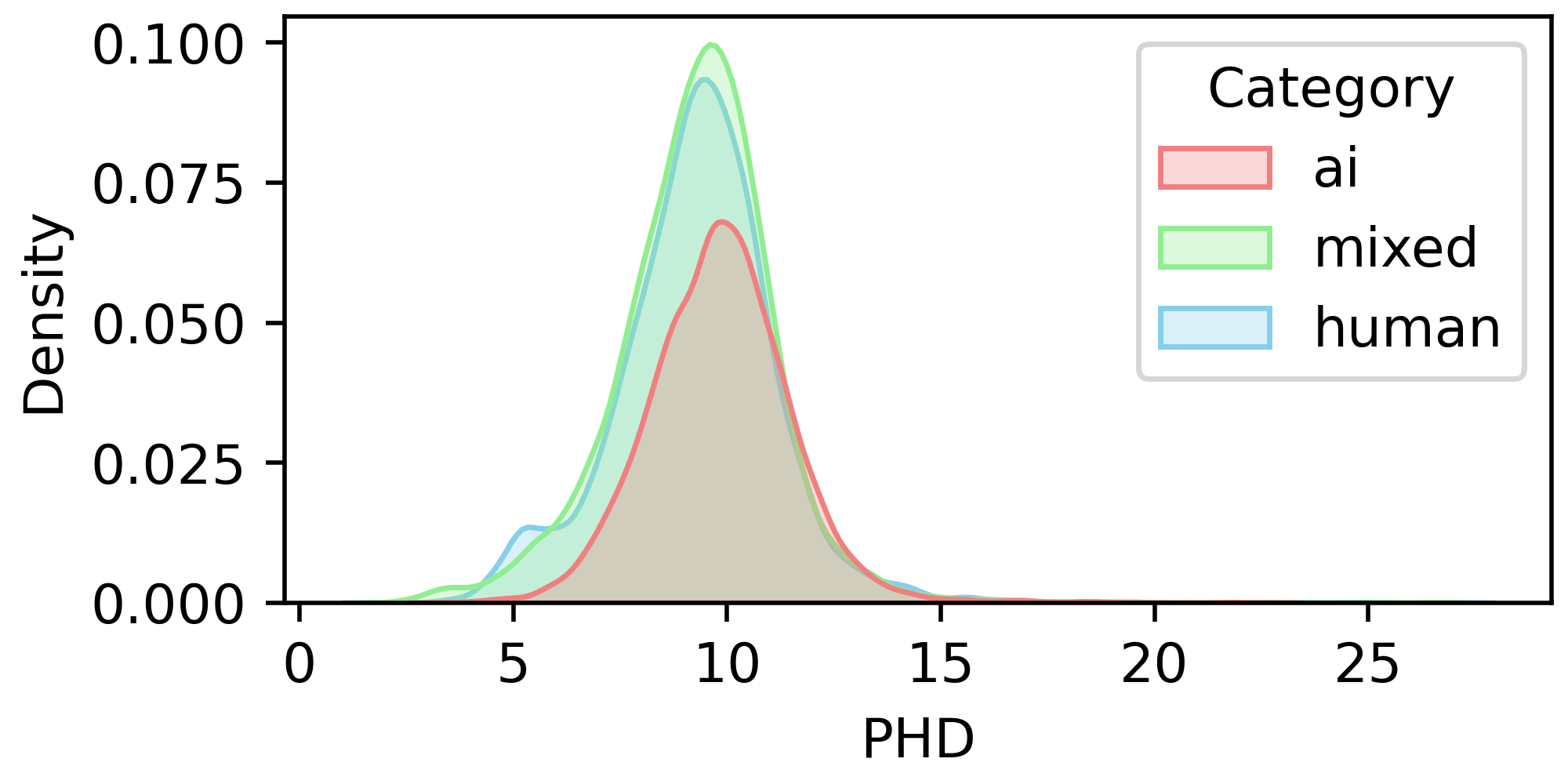}
        \caption{English detection dataset}
        \label{fig:phd_eng_detect}
    \end{subfigure}

    \caption{PHD comparison for classification (AI vs. Human) and detection (AI vs. Human vs. Mixed) datasets for Russian and English languages.}
    \label{fig:phd_dataset_comparison}
\end{figure}

Figure~\ref{fig:phd_dataset_comparison} provides a visual representation of the Persistence Homology Dimension (PHD) distributions for all categories (human, AI, and mixed) across our classification and detection datasets for both languages. These plots serve as a visual confirmation of the low $\text{KL}_{\text{TTS}}$ scores and closely aligned mean PHD values reported in Section~\ref{sec:quality}. The significant overlap between the distributions for all text types highlights the structural and topological similarity between our AI-generated, mixed, and human-authored texts, underscoring the challenge our dataset presents for detection models.

\section{Embedding Shift Distributions after Perturbation}
\label{app:shift_comparison}

\begin{figure*}[t]
    \centering

    % --- Classification (AI vs Human) ---
    
    \begin{subfigure}{0.48\textwidth}
        \centering
        \includegraphics[width=\linewidth]{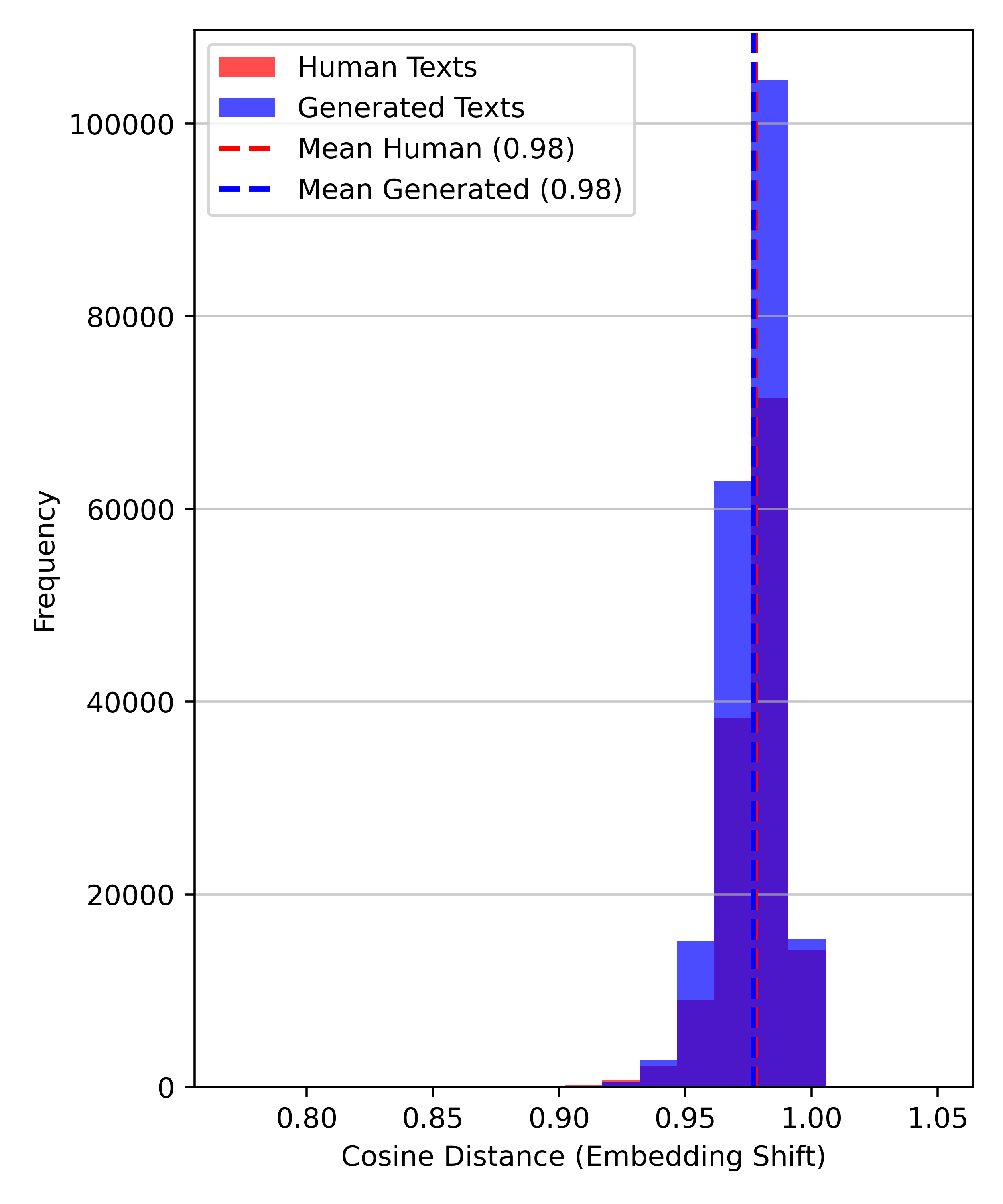}
        \caption{Embedding shifts for the Russian classification dataset.}
        \label{fig:shift_dist_ru}
    \end{subfigure}
    \hfill 
    \begin{subfigure}{0.48\textwidth}
        \centering
        \includegraphics[width=\linewidth]{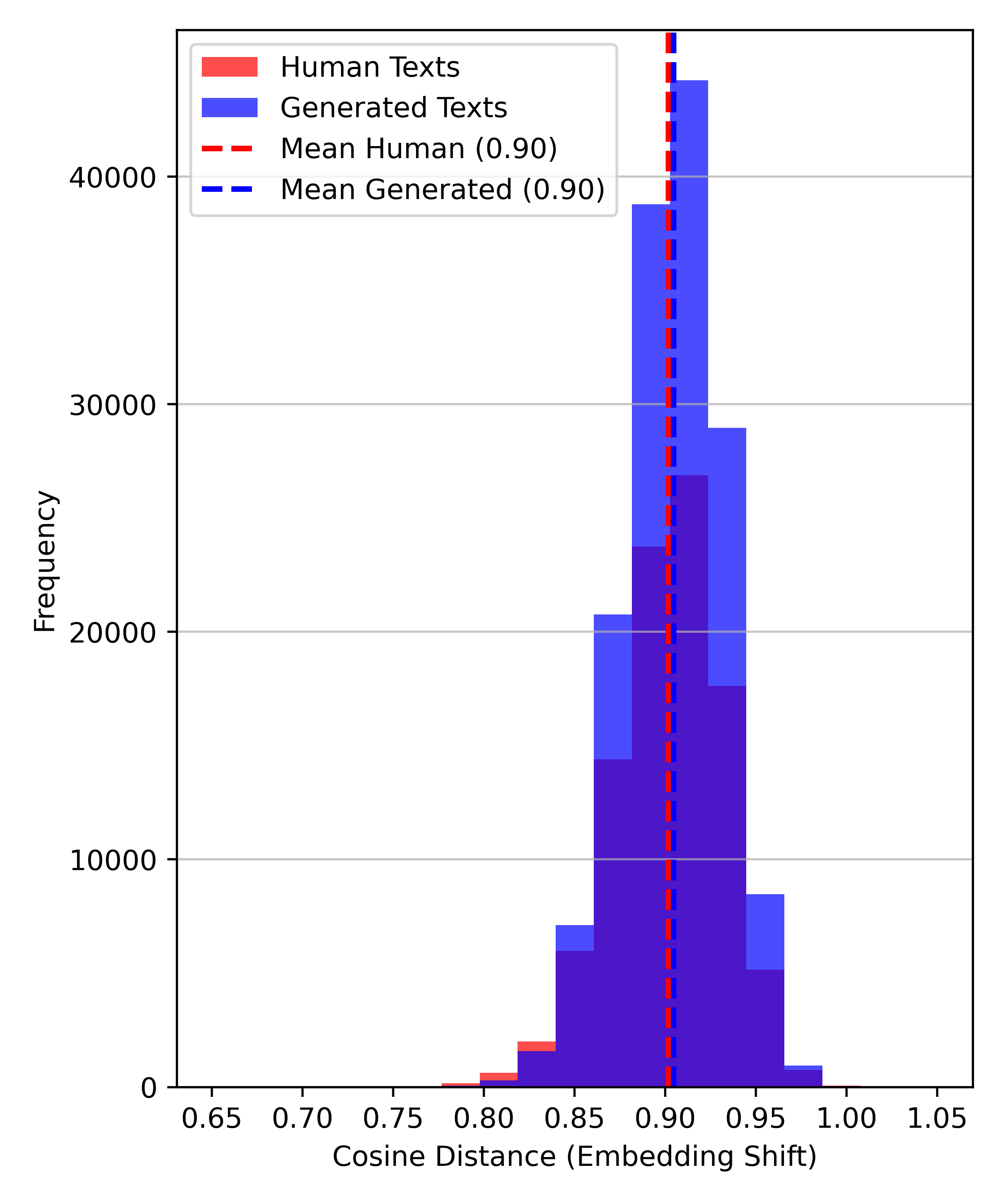}
        \caption{Embedding shifts for the English classification dataset.}
        \label{fig:shift_dist_en}
    \end{subfigure}

    \caption{Distributions of cosine distances between original and perturbed text embeddings for human and AI classes. The significant overlap demonstrates similar robustness to synonym-based perturbations for both classes.}
    \label{fig:cos_sim_dist}
\end{figure*}

This section provides a visual comparison of the embedding shifts for human and AI texts from our classification datasets after undergoing adversarial token perturbation (synonym replacement). The plots in Figure~\ref{fig:cos_sim_dist} show the distributions of cosine distances between the embeddings of original texts and their perturbed versions.

The close alignment and significant overlap between the distributions for human and AI texts in both the English (Figure~\ref{fig:shift_dist_en}) and Russian (Figure~\ref{fig:shift_dist_ru}) datasets visually confirm the low $|\Delta_{\text{shift}}|$ scores reported in Section~\ref{sec:quality}. This indicates that the AI-generated texts in our dataset exhibit a level of robustness to semantic perturbations that is highly comparable to that of human-authored texts, highlighting the quality and challenge of the dataset.

\section{Textual Similarity Metric Descriptions and Detailed Results}
\label{app:similarity_appendix}

\subsection{Similarity Metric Descriptions}
\label{app:similarity_metric_descriptions}

To provide a comprehensive assessment of text similarity, we employ a suite of metrics, each capturing a different aspect of the relationship between the original human text and the machine-generated paraphrase.

\begin{itemize}
    \item \textbf{METEOR} \cite{banerjee2005meteor} (Metric for Evaluation of Translation with Explicit ORdering) is a standard metric in machine translation that measures similarity based on unigram alignments between a reference and a hypothesis text, considering precision, recall, and fragmentation.
    
    \item \textbf{BERTScore} \cite{zhang2019bertscore} computes similarity by comparing the contextual embeddings of tokens from a reference and a hypothesis text. It uses a multilingual BERT model, making it robust for cross-lingual comparisons and sensitive to semantic similarity beyond exact word matches.
    
    \item \textbf{n-gram Similarity}\footnote{\url{https://pypi.org/project/ngram/}} is a language-independent metric that calculates the ratio of shared n-grams (in our case, 3-grams) between two strings, providing a measure of surface-level lexical overlap.
    
    \item \textbf{Levenshtein Distance (LD)}, reported as \textbf{ED-norm}, is a character-level metric that counts the minimum number of single-character edits (insertions, deletions, or substitutions) required to change one string into the other. We normalize this value by the character length of the human text. This calculation is performed using the \texttt{editdistance}\footnote{\url{https://github.com/roy-ht/editdistance}} library, consistent with the methodology of the MultiSocial benchmark \cite{macko2024multisocial}.
    
    \item \textbf{LangCheck} measures the percentage of generated texts for which the language differs from the language of the original human text, where both languages are detected using FastText\footnote{\url{https://fasttext.cc/}}. It serves as an indicator of potential style-imitation artifacts or language inconsistencies.
    
    \item \textbf{MAUVE} \cite{pillutla2021mauve} is a distributional metric that measures the gap between the distribution of machine-generated and human-written texts. It uses clustering on text embeddings, obtained from the \texttt{google-bert/bert-base-multilingual-cased} model, to quantify the divergence between the two text sets. A lower score indicates that the AI-generated distribution is closer to the human one. 
    Due to its computational intensity, all MAUVE scores were calculated on random data samples. For the per-model analysis (Appendix~\ref{app:per_model_results}), our sample consisted of 1,000 pairs per model, for a total of 15,000 English and 26,162 Russian pairs. For the per-prompt type analysis (Appendix~\ref{app:per_prompt_results}), we sampled 3,000 pairs for each of the four prompt types, yielding a total of 12,000 pairs per language.
\end{itemize}

\subsection{Per-Prompt Type Similarity Results}
\label{app:per_prompt_results}

To further analyze the characteristics of our dataset, we also computed the textual similarity metrics grouped by the type of prompt used for generation. This allows us to understand how different types of prompts influence the similarity between the AI-generated output and the human source text. The results for the English and Russian datasets are presented in Table~\ref{tab:eng_per_prompt} and Table~\ref{tab:rus_per_prompt}, respectively.

% --- ENGLISH PER-PROMPT TYPE TABLE ---
\begin{table*}[htbp!]
\centering
\caption{Per-prompt type similarity metrics for the English dataset. Asterisk (*) denotes that the MAUVE metric was calculated on a random sample of 3,000 text pairs per prompt type.}
\label{tab:eng_per_prompt}
\small
\begin{tabular}{lcccccc}
\hline
\textbf{\makecell{Prompt\\ Type}} & {\makecell{\textbf{METEOR ↑}}} & {\makecell{\textbf{BERTScore ↑}}} & {\makecell{\textbf{n-gram ↑}}} & {\makecell{\textbf{ED-norm ↓}}} & {\makecell{\textbf{LangCheck ↓}}} & {\makecell{\textbf{MAUVE* ↓}}} \\
\hline
create & \makecell{0.3955 \\ ($\pm$ 0.3010)} & \makecell{0.7457 \\ ($\pm$ 0.1274)} & \makecell{0.3070 \\ ($\pm$ 0.2313)} & \makecell{2.1116 \\ ($\pm$ 5.0455)} & \makecell{0.0023 \\ ($\pm$ 0.0483)} & 0.3307 \\
delete & \makecell{0.1542 \\ ($\pm$ 0.1177)} & \makecell{0.6967 \\ ($\pm$ 0.0614)} & \makecell{0.2100 \\ ($\pm$ 0.1308)} & \makecell{0.7986 \\ ($\pm$ 0.1554)} & \makecell{0.0008 \\ ($\pm$ 0.0290)} & 0.0645 \\
expand & \makecell{0.3090 \\ ($\pm$ 0.1447)} & \makecell{0.6877 \\ ($\pm$ 0.0610)} & \makecell{0.1706 \\ ($\pm$ 0.1094)} & \makecell{5.5252 \\ ($\pm$ 9.2193)} & \makecell{0.0025 \\ ($\pm$ 0.0501)} & 0.0673 \\
update & \makecell{0.2022 \\ ($\pm$ 0.1045)} & \makecell{0.6578 \\ ($\pm$ 0.0526)} & \makecell{0.1792 \\ ($\pm$ 0.1017)} & \makecell{3.4772 \\ ($\pm$ 7.8540)} & \makecell{0.0018 \\ ($\pm$ 0.0426)} & 0.1389 \\
\hline
\textbf{Global} & \makecell{0.2665 \\ ($\pm$ 0.2071)} & \makecell{0.6964 \\ ($\pm$ 0.0876)} & \makecell{0.2164 \\ ($\pm$ 0.1618)} & \makecell{3.0255 \\ ($\pm$ 6.8643)} & \makecell{0.0019 \\ ($\pm$ 0.0435)} & 0.1675 \\
\hline
\end{tabular}
\end{table*}

% --- RUSSIAN PER-PROMPT TYPE TABLE ---
\begin{table*}[htbp!]
\centering
\caption{Per-prompt type similarity metrics for the Russian dataset. Asterisk (*) denotes that the MAUVE metric was calculated on a random sample of 3,000 text pairs per prompt type.}
\label{tab:rus_per_prompt}
\small
\begin{tabular}{lcccccc}
\hline
\textbf{\makecell{Prompt\\ Type}} & {\makecell{\textbf{METEOR ↑}}} & {\makecell{\textbf{BERTScore ↑}}} & {\makecell{\textbf{n-gram ↑}}} & {\makecell{\textbf{ED-norm ↓}}} & {\makecell{\textbf{LangCheck ↓}}} & {\makecell{\textbf{MAUVE* ↓}}} \\
\hline
create & \makecell{0.1616 \\ ($\pm$ 0.1217)} & \makecell{0.6568 \\ ($\pm$ 0.0577)} & \makecell{0.1307 \\ ($\pm$ 0.0884)} & \makecell{2.5010 \\ ($\pm$ 4.6219)} & \makecell{0.0037 \\ ($\pm$ 0.0608)} & 0.1719 \\
delete & \makecell{0.1235 \\ ($\pm$ 0.1186)} & \makecell{0.6894 \\ ($\pm$ 0.0602)} & \makecell{0.1684 \\ ($\pm$ 0.1226)} & \makecell{1.2259 \\ ($\pm$ 2.7432)} & \makecell{0.0060 \\ ($\pm$ 0.0770)} & 0.0829 \\
expand & \makecell{0.3223 \\ ($\pm$ 0.1780)} & \makecell{0.7053 \\ ($\pm$ 0.0688)} & \makecell{0.2179 \\ ($\pm$ 0.1372)} & \makecell{4.0885 \\ ($\pm$ 21.1010)} & \makecell{0.0068 \\ ($\pm$ 0.0823)} & 0.1175 \\
update & \makecell{0.1648 \\ ($\pm$ 0.1303)} & \makecell{0.6786 \\ ($\pm$ 0.0551)} & \makecell{0.1752 \\ ($\pm$ 0.1062)} & \makecell{2.5295 \\ ($\pm$ 7.7285)} & \makecell{0.0028 \\ ($\pm$ 0.0529)} & 0.2235 \\
\midrule
\textbf{Global} & \makecell{0.1797 \\ ($\pm$ 0.1482)} & \makecell{0.6767 \\ ($\pm$ 0.0623)} & \makecell{0.1630 \\ ($\pm$ 0.1135)} & \makecell{2.4792 \\ ($\pm$ 9.7793)} & \makecell{0.0045 \\ ($\pm$ 0.0670)} & 0.1887 \\
\hline
\end{tabular}
\end{table*}

\subsection{Per-Model Similarity Results}
\label{app:per_model_results}

The following tables provide a detailed breakdown of performance for each LLM, evaluated separately on the English (Table~\ref{tab:eng_per_model}) and Russian (Table~\ref{tab:rus_per_model}) datasets.

% --- ENGLISH PER-MODEL TABLE ---
\begin{table*}[h!]
\centering
\caption{Per-model similarity metrics for the English dataset. Asterisk (*) denotes that the MAUVE metric was calculated on a random sample of 1,000 text pairs per model. Other metrics were calculated on the full English classification dataset.}
\label{tab:eng_per_model}
\resizebox{0.99\textwidth}{!}{
\begin{tabular}{lcccccc}
\hline
\textbf{Model} & {\makecell{\textbf{METEOR ↑}}} & {\makecell{\textbf{BERTScore ↑}}} & {\makecell{\textbf{n-gram ↑}}} & {\makecell{\textbf{ED-norm ↓}}} & {\makecell{\textbf{LangCheck ↓}}} & {\makecell{\textbf{MAUVE* ↓}}} \\
\hline
c4ai-command-r-08-2024     & \makecell{0.2515 \\ ($\pm$ 0.2003)} & \makecell{0.6903 \\ ($\pm$ 0.0881)} & \makecell{0.2119 \\ ($\pm$ 0.1674)} & \makecell{2.8851 \\ ($\pm$ 7.7371)} & \makecell{0.0017 \\ ($\pm$ 0.0417)} & 0.1765 \\
Qwen2.5-72B-Instruct              & \makecell{0.2782 \\ ($\pm$ 0.2083)} & \makecell{0.7033 \\ ($\pm$ 0.0849)} & \makecell{0.2248 \\ ($\pm$ 0.1661)} & \makecell{3.0838 \\ ($\pm$ 6.8450)} & \makecell{0.0013 \\ ($\pm$ 0.0360)} & 0.3581 \\
Qwen3-32B                         & \makecell{0.2499 \\ ($\pm$ 0.1859)} & \makecell{0.6866 \\ ($\pm$ 0.0816)} & \makecell{0.2053 \\ ($\pm$ 0.1462)} & \makecell{3.2182 \\ ($\pm$ 6.9384)} & \makecell{0.0023 \\ ($\pm$ 0.0479)} & 0.1893 \\
dbrx-instruct               & \makecell{0.2843 \\ ($\pm$ 0.1877)} & \makecell{0.7043 \\ ($\pm$ 0.0828)} & \makecell{0.2302 \\ ($\pm$ 0.1591)} & \makecell{3.0995 \\ ($\pm$ 6.3418)} & \makecell{0.0039 \\ ($\pm$ 0.0625)} & 0.2929 \\
DeepSeek-R1-Distill-Qwen-32B & \makecell{0.2656 \\ ($\pm$ 0.1969)} & \makecell{0.6984 \\ ($\pm$ 0.0833)} & \makecell{0.2183 \\ ($\pm$ 0.1591)} & \makecell{3.0828 \\ ($\pm$ 6.5980)} & \makecell{0.0017 \\ ($\pm$ 0.0412)} & 0.2872 \\
gemini-2.0-flash                       & \makecell{0.2529 \\ ($\pm$ 0.2212)} & \makecell{0.6921 \\ ($\pm$ 0.0930)} & \makecell{0.2105 \\ ($\pm$ 0.1652)} & \makecell{2.7950 \\ ($\pm$ 6.5796)} & \makecell{0.0014 \\ ($\pm$ 0.0379)} & 0.3915 \\
gemini-2.5-flash                       & \makecell{0.2644 \\ ($\pm$ 0.2218)} & \makecell{0.6878 \\ ($\pm$ 0.0907)} & \makecell{0.2064 \\ ($\pm$ 0.1510)} & \makecell{3.3609 \\ ($\pm$ 9.2018)} & \makecell{0.0016 \\ ($\pm$ 0.0399)} & 0.2379 \\
gpt-4.1-2025-04-14                     & \makecell{0.2601 \\ ($\pm$ 0.2041)} & \makecell{0.6975 \\ ($\pm$ 0.0876)} & \makecell{0.2164 \\ ($\pm$ 0.1485)} & \makecell{2.7985 \\ ($\pm$ 5.7855)} & \makecell{0.0012 \\ ($\pm$ 0.0351)} & 0.2780 \\
Magistral-Small-2507         & \makecell{0.3149 \\ ($\pm$ 0.2476)} & \makecell{0.7242 \\ ($\pm$ 0.1006)} & \makecell{0.2542 \\ ($\pm$ 0.2036)} & \makecell{2.5281 \\ ($\pm$ 5.7650)} & \makecell{0.0026 \\ ($\pm$ 0.0509)} & 0.4683 \\
Ministral-8B-Instruct-2410   & \makecell{0.2831 \\ ($\pm$ 0.2414)} & \makecell{0.7135 \\ ($\pm$ 0.0945)} & \makecell{0.2329 \\ ($\pm$ 0.1917)} & \makecell{2.2827 \\ ($\pm$ 5.1872)} & \makecell{0.0018 \\ ($\pm$ 0.0423)} & 0.3413 \\
Llama-3.1-Nemotron-70B-Instruct-HF & \makecell{0.2575 \\ ($\pm$ 0.1717)} & \makecell{0.6796 \\ ($\pm$ 0.0768)} & \makecell{0.2019 \\ ($\pm$ 0.1356)} & \makecell{3.7079 \\ ($\pm$ 7.5270)} & \makecell{0.0018 \\ ($\pm$ 0.0425)} & 0.1534 \\
o3-2025-04-16                          & \makecell{0.2257 \\ ($\pm$ 0.1816)} & \makecell{0.6798 \\ ($\pm$ 0.0808)} & \makecell{0.1870 \\ ($\pm$ 0.1279)} & \makecell{3.0114 \\ ($\pm$ 6.2277)} & \makecell{0.0022 \\ ($\pm$ 0.0467)} & 0.2437 \\
Falcon3-10B-Instruct            & \makecell{0.2729 \\ ($\pm$ 0.2160)} & \makecell{0.7033 \\ ($\pm$ 0.0895)} & \makecell{0.2224 \\ ($\pm$ 0.1716)} & \makecell{2.7260 \\ ($\pm$ 5.7081)} & \makecell{0.0013 \\ ($\pm$ 0.0355)} & 0.3005 \\
Llama-3.3-70B-Instruct         & \makecell{0.2611 \\ ($\pm$ 0.1905)} & \makecell{0.6895 \\ ($\pm$ 0.0835)} & \makecell{0.2107 \\ ($\pm$ 0.1597)} & \makecell{3.4868 \\ ($\pm$ 8.4771)} & \makecell{0.0017 \\ ($\pm$ 0.0408)} & 0.2358 \\
GLM-4-32B-0414                 & \makecell{0.2923 \\ ($\pm$ 0.2193)} & \makecell{0.7065 \\ ($\pm$ 0.0865)} & \makecell{0.2268 \\ ($\pm$ 0.1653)} & \makecell{3.0156 \\ ($\pm$ 5.8937)} & \makecell{0.0022 \\ ($\pm$ 0.0464)} & 0.3191 \\
\hline
\end{tabular}%
}
\end{table*}

% --- RUSSIAN PER-MODEL TABLE ---
\begin{table*}[htbp!]
\centering
\caption{Per-model similarity metrics for the Russian dataset. Asterisk (*) denotes that the MAUVE metric was calculated on a random sample of 1,000 text pairs per model. Other metrics were calculated on the full Russian classification dataset.}
\label{tab:rus_per_model}
\resizebox{0.99\textwidth}{!}{
\begin{tabular}{lcccccc}
\hline
\textbf{Model} & {\makecell{\textbf{METEOR ↑}}} & {\makecell{\textbf{BERTScore ↑}}} & {\makecell{\textbf{n-gram ↑}}} & {\makecell{\textbf{ED-norm ↓}}} & {\makecell{\textbf{LangCheck ↓}}} & {\makecell{\textbf{MAUVE* ↓}}} \\
\hline
Yi-1.5-34B-Chat                   & \makecell{0.1957 \\ ($\pm$ 0.1529)} & \makecell{0.6961 \\ ($\pm$ 0.0596)} & \makecell{0.1965 \\ ($\pm$ 0.1160)} & \makecell{1.9419 \\ ($\pm$ 3.5813)} & \makecell{0.0065 \\ ($\pm$ 0.0804)} & 0.2268 \\
c4ai-command-r-08-2024     & \makecell{0.1616 \\ ($\pm$ 0.1074)} & \makecell{0.6787 \\ ($\pm$ 0.0466)} & \makecell{0.1776 \\ ($\pm$ 0.0973)} & \makecell{2.2883 \\ ($\pm$ 4.0365)} & \makecell{0.0014 \\ ($\pm$ 0.0372)} & 0.1627 \\
GigaChat-Max                           & \makecell{0.1497 \\ ($\pm$ 0.1215)} & \makecell{0.6781 \\ ($\pm$ 0.0482)} & \makecell{0.1693 \\ ($\pm$ 0.0933)} & \makecell{2.1745 \\ ($\pm$ 3.5091)} & \makecell{0.0010 \\ ($\pm$ 0.0314)} & 0.2129 \\
Jamba-v0.1                             & \makecell{0.1093 \\ ($\pm$ 0.1436)} & \makecell{0.6360 \\ ($\pm$ 0.0716)} & \makecell{0.1035 \\ ($\pm$ 0.1033)} & \makecell{1.7394 \\ ($\pm$ 3.0358)} & \makecell{0.0000 \\ ($\pm$ 0.0000)} & 0.1391 \\
Meta-Llama-3-8B-Instruct               & \makecell{0.2137 \\ ($\pm$ 0.1346)} & \makecell{0.6728 \\ ($\pm$ 0.0640)} & \makecell{0.1695 \\ ($\pm$ 0.0916)} & \makecell{3.3092 \\ ($\pm$ 5.8424)} & \makecell{0.0012 \\ ($\pm$ 0.0346)} & 0.0539 \\
Phi-3-mini-128k-instruct               & \makecell{0.2120 \\ ($\pm$ 0.1312)} & \makecell{0.6461 \\ ($\pm$ 0.0607)} & \makecell{0.1456 \\ ($\pm$ 0.0808)} & \makecell{3.5791 \\ ($\pm$ 5.9486)} & \makecell{0.0055 \\ ($\pm$ 0.0741)} & 0.0513 \\
QwQ-32B                          & \makecell{0.1496 \\ ($\pm$ 0.1007)} & \makecell{0.6787 \\ ($\pm$ 0.0448)} & \makecell{0.1640 \\ ($\pm$ 0.0892)} & \makecell{2.7626 \\ ($\pm$ 17.3150)} & \makecell{0.0017 \\ ($\pm$ 0.0413)} & 0.1776 \\
Qwen2.5-72B-Instruct              & \makecell{0.2117 \\ ($\pm$ 0.1703)} & \makecell{0.7000 \\ ($\pm$ 0.0592)} & \makecell{0.1956 \\ ($\pm$ 0.1245)} & \makecell{2.2224 \\ ($\pm$ 4.1846)} & \makecell{0.0014 \\ ($\pm$ 0.0375)} & 0.3678 \\
Qwen2-7B-Instruct                      & \makecell{0.1831 \\ ($\pm$ 0.1114)} & \makecell{0.6653 \\ ($\pm$ 0.0585)} & \makecell{0.1536 \\ ($\pm$ 0.0920)} & \makecell{4.8198 \\ ($\pm$ 8.1632)} & \makecell{0.0008 \\ ($\pm$ 0.0275)} & 0.0523 \\
T5                                     & \makecell{0.1775 \\ ($\pm$ 0.1430)} & \makecell{0.6979 \\ ($\pm$ 0.0642)} & \makecell{0.1388 \\ ($\pm$ 0.1022)} & \makecell{2.6308 \\ ($\pm$ 3.3524)} & \makecell{0.0000 \\ ($\pm$ 0.0000)} & 0.9973 \\
WizardLM-2-7B                          & \makecell{0.2030 \\ ($\pm$ 0.1338)} & \makecell{0.6676 \\ ($\pm$ 0.0581)} & \makecell{0.1642 \\ ($\pm$ 0.0928)} & \makecell{3.1223 \\ ($\pm$ 5.5773)} & \makecell{0.0043 \\ ($\pm$ 0.0652)} & 0.0453 \\
dbrx-instruct               & \makecell{0.2167 \\ ($\pm$ 0.1837)} & \makecell{0.6999 \\ ($\pm$ 0.0659)} & \makecell{0.1952 \\ ($\pm$ 0.1366)} & \makecell{2.1221 \\ ($\pm$ 5.2134)} & \makecell{0.0206 \\ ($\pm$ 0.1419)} & 0.2743 \\
DeepSeek-R1-Distill-Qwen-32B & \makecell{0.1732 \\ ($\pm$ 0.1578)} & \makecell{0.6879 \\ ($\pm$ 0.0621)} & \makecell{0.1738 \\ ($\pm$ 0.1207)} & \makecell{3.2682 \\ ($\pm$ 40.5175)} & \makecell{0.0223 \\ ($\pm$ 0.1478)} & 0.2339 \\
gemma-1.1-7b-it                        & \makecell{0.1493 \\ ($\pm$ 0.1038)} & \makecell{0.6363 \\ ($\pm$ 0.0542)} & \makecell{0.1278 \\ ($\pm$ 0.0718)} & \makecell{3.9172 \\ ($\pm$ 6.7532)} & \makecell{0.0051 \\ ($\pm$ 0.0714)} & 0.0082 \\
GigaChat                               & \makecell{0.1506 \\ ($\pm$ 0.1314)} & \makecell{0.6573 \\ ($\pm$ 0.0663)} & \makecell{0.1272 \\ ($\pm$ 0.1007)} & \makecell{3.2317 \\ ($\pm$ 5.1930)} & \makecell{0.0024 \\ ($\pm$ 0.0492)} & 0.1378 \\
gemma-2-27b-it                  & \makecell{0.1730 \\ ($\pm$ 0.1381)} & \makecell{0.6810 \\ ($\pm$ 0.0494)} & \makecell{0.1724 \\ ($\pm$ 0.1003)} & \makecell{1.9631 \\ ($\pm$ 3.1443)} & \makecell{0.0015 \\ ($\pm$ 0.0388)} & 0.2404 \\
gpt-3.5                                & \makecell{0.1766 \\ ($\pm$ 0.1273)} & \makecell{0.6646 \\ ($\pm$ 0.0573)} & \makecell{0.1474 \\ ($\pm$ 0.0934)} & \makecell{3.0603 \\ ($\pm$ 4.3994)} & \makecell{0.0015 \\ ($\pm$ 0.0393)} & 0.1169 \\
gpt-4-0125-preview                     & \makecell{0.1076 \\ ($\pm$ 0.0513)} & \makecell{0.6201 \\ ($\pm$ 0.0379)} & \makecell{0.0990 \\ ($\pm$ 0.0647)} & \makecell{5.7928 \\ ($\pm$ 8.0727)} & \makecell{0.0047 \\ ($\pm$ 0.0682)} & 0.0846 \\
gpt-4-1106-preview                     & \makecell{0.1053 \\ ($\pm$ 0.0503)} & \makecell{0.6182 \\ ($\pm$ 0.0371)} & \makecell{0.0934 \\ ($\pm$ 0.0619)} & \makecell{5.5213 \\ ($\pm$ 7.7930)} & \makecell{0.0021 \\ ($\pm$ 0.0460)} & 0.0629 \\
gpt-4o                                 & \makecell{0.1178 \\ ($\pm$ 0.0795)} & \makecell{0.6455 \\ ($\pm$ 0.0463)} & \makecell{0.0921 \\ ($\pm$ 0.0570)} & \makecell{1.0561 \\ ($\pm$ 0.6252)} & \makecell{0.0060 \\ ($\pm$ 0.0771)} & 0.1094 \\
llama-7b                               & \makecell{0.2260 \\ ($\pm$ 0.1361)} & \makecell{0.6798 \\ ($\pm$ 0.0577)} & \makecell{0.1753 \\ ($\pm$ 0.0953)} & \makecell{2.7548 \\ ($\pm$ 4.2883)} & \makecell{0.0018 \\ ($\pm$ 0.0422)} & 0.3801 \\
Phi-3-medium-128k-instruct   & \makecell{0.2053 \\ ($\pm$ 0.1200)} & \makecell{0.7183 \\ ($\pm$ 0.0558)} & \makecell{0.2652 \\ ($\pm$ 0.1203)} & \makecell{5.3981 \\ ($\pm$ 21.8489)} & \makecell{0.0095 \\ ($\pm$ 0.0976)} & 0.1389 \\
Ministral-8B-Instruct-2410   & \makecell{0.2689 \\ ($\pm$ 0.2380)} & \makecell{0.7254 \\ ($\pm$ 0.0799)} & \makecell{0.2351 \\ ($\pm$ 0.1815)} & \makecell{1.8615 \\ ($\pm$ 8.5790)} & \makecell{0.0024 \\ ($\pm$ 0.0485)} & 0.3012 \\
o1-mini-2024-09-12                     & \makecell{0.1881 \\ ($\pm$ 0.1108)} & \makecell{0.6748 \\ ($\pm$ 0.0507)} & \makecell{0.1849 \\ ($\pm$ 0.0884)} & \makecell{1.5011 \\ ($\pm$ 2.3402)} & \makecell{0.0000 \\ ($\pm$ 0.0000)} & 0.2687 \\
o1-preview-2024-09-12                  & \makecell{0.1785 \\ ($\pm$ 0.1134)} & \makecell{0.6789 \\ ($\pm$ 0.0540)} & \makecell{0.2330 \\ ($\pm$ 0.1115)} & \makecell{1.9907 \\ ($\pm$ 5.2266)} & \makecell{0.0000 \\ ($\pm$ 0.0000)} & 0.3393 \\
o3-2025-04-16                          & \makecell{0.1523 \\ ($\pm$ 0.1003)} & \makecell{0.6680 \\ ($\pm$ 0.0464)} & \makecell{0.1549 \\ ($\pm$ 0.0849)} & \makecell{3.0001 \\ ($\pm$ 8.0777)} & \makecell{0.0042 \\ ($\pm$ 0.0649)} & 0.1021 \\
ruGPT                                  & \makecell{0.2053 \\ ($\pm$ 0.1250)} & \makecell{0.6605 \\ ($\pm$ 0.0595)} & \makecell{0.1416 \\ ($\pm$ 0.0901)} & \makecell{5.8582 \\ ($\pm$ 9.0776)} & \makecell{0.0032 \\ ($\pm$ 0.0566)} & 0.2552 \\
Llama-3.3-70B-Instruct         & \makecell{0.2263 \\ ($\pm$ 0.1788)} & \makecell{0.7000 \\ ($\pm$ 0.0622)} & \makecell{0.2033 \\ ($\pm$ 0.1273)} & \makecell{2.4090 \\ ($\pm$ 5.8166)} & \makecell{0.0012 \\ ($\pm$ 0.0345)} & 0.1998 \\
YaGPT                                 & \makecell{0.1777 \\ ($\pm$ 0.1371)} & \makecell{0.6646 \\ ($\pm$ 0.0639)} & \makecell{0.1520 \\ ($\pm$ 0.0964)} & \makecell{2.6590 \\ ($\pm$ 3.7624)} & \makecell{0.0006 \\ ($\pm$ 0.0239)} & 0.1359 \\
YandexGPT-5-Lite-8B-instruct    & \makecell{0.2241 \\ ($\pm$ 0.1791)} & \makecell{0.7025 \\ ($\pm$ 0.0630)} & \makecell{0.2009 \\ ($\pm$ 0.1269)} & \makecell{2.0329 \\ ($\pm$ 3.4634)} & \makecell{0.0029 \\ ($\pm$ 0.0536)} & 0.2323 \\
\hline
\end{tabular}
}
\end{table*}

\section{GigaCheck Training Hyperparameters}
\label{app:training_hyperparameters}

All models were trained using the \texttt{transformers}\footnote{https://github.com/huggingface/transformers} library.

\paragraph{Classification Models (Mistral-7B and mmBERT-base).}
The classification models were trained with the following key hyperparameters:

\begin{itemize}
    \item \textbf{Pretrained model:} Mistral-7B-v0.3 / mmBERT-base
    \item \textbf{Sequence length:} max 1024, min 100, random sequence length enabled
    \item \textbf{LoRA:} rank=8, alpha=16
    \item \textbf{Precision:} bf16
    \item \textbf{Batch size per GPU:} 64 (train), 1 (eval)
    \item \textbf{Number of GPUs:} 8
    \item \textbf{Gradient accumulation steps:} 1
    \item \textbf{Optimizer:} AdamW
    \item \textbf{Learning rate:} $3\cdot10^{-5}$, cosine scheduler with min LR scaled by 0.5
    \item \textbf{Warmup steps:} 20
    \item \textbf{Number of epochs:} 20
    \item \textbf{Random seed:} 8888
    \item \textbf{Classifier pooling for mmBERT-base:} CLS
\end{itemize}

\paragraph{Detection Model (GigaCheck DN-DAB-DETR).}
The DN-DAB-DETR detection models were trained with the following key hyperparameters:
\begin{itemize}
    \item \textbf{Feature extractor:} Mistral-7B-v0.3 (frozen) / mmBERT-base (frozen) 
    \item \textbf{Sequence length:} max 1024, min 100, random sequence length enabled
    \item \textbf{Precision:} bf16 for the frozen feature extractor, fp32 for the trained DN-DAB-DETR
    \item \textbf{Batch size per GPU:} 64 (train), 1 (eval)
    \item \textbf{Gradient accumulation steps:} 1
    \item \textbf{Number of GPUs:} 8
    \item \textbf{Optimizer:} AdamW
    \item \textbf{Weight decay:} $1\cdot10^{-4}$
    \item \textbf{Learning rate:} $2\cdot10^{-4}$, cosine scheduler with min LR scaled by 0.5
    \item \textbf{Warmup steps:} 100
    \item \textbf{Number of epochs:} 150
    \item \textbf{DETR parameters:} 45 queries, 3 encoder and decoder layers, input embedding dimension 256
\end{itemize}

\section{Detailed Classification Results}
\label{app:classification_results}

To provide a more granular analysis of our binary classification models' performance, we report the results disaggregated by several key factors. Table~\ref{tab:classification_results_data_type} shows the performance across different text domains. Table~\ref{tab:classification_results_word_bin} presents the results broken down by text length, grouped into bins based on word count. Finally, Table~\ref{tab:classification_results_prompt_types} details the performance for each of the prompt types used during data generation. This detailed breakdown demonstrates the model's robust performance across various conditions and data subsets.

% --- Table for Classification Results PER-DATA-TYPE ---
\begin{table*}[htbp!]
    \centering
    \caption{Metrics for each dataset and data type for the Mistral-7B backbone.}
    \begin{tabular}{l l c c c c}
        \hline
        \textbf{Dataset} & \textbf{Data Type} & \textbf{AI F1} & \textbf{Human F1} & \textbf{Mean Accuracy} & \textbf{TPR@FPR=0.01} \\
        \hline
        \multirow{9}{*}{English-only} 
            & Article & 0.9864 & 0.9829 & 0.9857 & 0.9798 \\
            & Factual text & 0.9867 & 0.9803 & 0.9852 & 0.9796 \\
            & News & 0.9881 & 0.9866 & 0.9877 & 0.9827 \\
            & Paper abstract & 0.9934 & 0.9894 & 0.9923 & 0.9919 \\
            & Poetry & 0.9886 & 0.9820 & 0.9863 & 0.9830 \\
            & Question & 0.9840 & 0.9775 & 0.9833 & 0.9793 \\
            & Review & 0.9778 & 0.9654 & 0.9767 & 0.9629 \\
            & Story & 0.9926 & 0.9861 & 0.9910 & 0.9890 \\
            & Short-form text & 0.9784 & 0.9439 & 0.9680 & 0.9434 \\
        \hline
        \multirow{8}{*}{Russian-only} 
            & Article & 0.9883 & 0.9878 & 0.9881 & 0.9856 \\
            & Factual text & 0.9836 & 0.9743 & 0.9810 & 0.9697 \\
            & News & 0.9814 & 0.9805 & 0.9809 & 0.9739 \\
            & Poetry & 0.9812 & 0.9825 & 0.9816 & 0.9746 \\
            & Question & 0.9921 & 0.9760 & 0.9857 & 0.9849 \\
            & Review & 0.9893 & 0.9841 & 0.9878 & 0.9853 \\
            & Story & 0.9878 & 0.9846 & 0.9863 & 0.9841 \\
            & Short-form text & 0.9784 & 0.8691 & 0.9551 & 0.8720 \\
        \hline
        \multirow{9}{*}{Bilingual} 
            & Article & 0.9890 & 0.9878 & 0.9887 & 0.9854 \\
            & Factual text & 0.9853 & 0.9778 & 0.9838 & 0.9772 \\
            & News & 0.9844 & 0.9832 & 0.9839 & 0.9787 \\
            & Paper abstract & 0.9927 & 0.9882 & 0.9905 & 0.9904 \\
            & Poetry & 0.9853 & 0.9839 & 0.9850 & 0.9808 \\
            & Question & 0.9888 & 0.9750 & 0.9837 & 0.9815 \\
            & Review & 0.9839 & 0.9758 & 0.9826 & 0.9747 \\
            & Story & 0.9894 & 0.9845 & 0.9878 & 0.9855 \\
            & Short-form text & 0.9789 & 0.9206 & 0.9653 & 0.9056 \\
        \hline
    \end{tabular}
    \label{tab:classification_results_data_type}
\end{table*}

% --- Table for Classification Results PER-WORDS-BIN ---
\begin{table*}[htbp!]
    \centering
    \caption{Metrics for each dataset and word bin for the Mistral-7B backbone.}
    \begin{tabular}{l l c c c c}
        \hline
        \textbf{Dataset} & \textbf{\# Words} & \textbf{AI F1} & \textbf{Human F1} & \textbf{Mean Accuracy} & \textbf{TPR@FPR=0.01} \\
        \hline
        \multirow{3}{*}{English-only} 
            & \textless 100 & 0.9747 & 0.9674 & 0.9725 & 0.9500 \\
            & 100-400 & 0.9939 & 0.9891 & 0.9936 & 0.9914 \\
            & \textgreater 400 & 0.9860 & 0.9813 & 0.9859 & 0.9778 \\
        \hline
        \multirow{3}{*}{Russian-only} 
            & \textless 100 & 0.9746 & 0.9729 & 0.9740 & 0.9495 \\
            & 100-400 & 0.9946 & 0.9860 & 0.9931 & 0.9941 \\
            & \textgreater 400 & 0.9785 & 0.9932 & 0.9804 & 0.9706 \\
        \hline
        \multirow{3}{*}{Bilingual} 
            & \textless 100 & 0.9747 & 0.9710 & 0.9734 & 0.9478 \\
            & 100-400 & 0.9946 & 0.9881 & 0.9937 & 0.9935 \\
            & \textgreater 400 & 0.9835 & 0.9882 & 0.9839 & 0.9751 \\
        \hline
    \end{tabular}
    \label{tab:classification_results_word_bin}
\end{table*}

% --- Table for Classification Results PER-PROMPT-TYPE ---
\begin{table*}[htbp!]
    \centering
    \caption{Metrics for each dataset and prompt type for the Mistral-7B backbone.}
    \begin{tabular}{l l c c}
        \hline
        \textbf{Dataset} & \textbf{Prompt Type} & \textbf{AI F1} & \textbf{AI Accuracy} \\
        \hline
        \multirow{4}{*}{English-only} 
            & Create & 0.9674 & 0.9368 \\
            & Delete & 0.9940 & 0.9880 \\
            & Expand & 0.9986 & 0.9972 \\
            & Update & 0.9966 & 0.9933 \\
        \hline
        \multirow{4}{*}{Russian-only} 
            & Create & 0.9919 & 0.9839 \\
            & Delete & 0.9820 & 0.9646 \\
            & Expand & 0.9937 & 0.9874 \\
            & Update & 0.9936 & 0.9875 \\
        \hline
        \multirow{4}{*}{Bilingual} 
            & Create & 0.9844 & 0.9692 \\
            & Delete & 0.9878 & 0.9759 \\
            & Expand & 0.9968 & 0.9936 \\
            & Update & 0.9959 & 0.9919 \\
        \hline
    \end{tabular}
    \label{tab:classification_results_prompt_types}
\end{table*}

% --- Table for Classification Results PER-MODEL ---
\begin{table*}[htbp!]
    \centering
    \caption{Metrics for each dataset and generator (only top-5 generators with highest AI F1 metric and top-5 generators with lowest metric are reported) for the Mistral-7B backbone.}
    \begin{tabular}{l l c c}
        \hline
        \textbf{Dataset} & \textbf{Generator} & \textbf{AI F1} & \textbf{AI Accuracy} \\
        \hline
        \multirow{4}{*}{English-only (\textbf{highest metrics})} 
            & llama-3.1-nemotron-70b-instruct-hf & 0.9974 & 0.9949 \\
            & qwen/qwen3-32b & 0.9965 & 0.9930 \\
            & databricks/dbrx-instruct & 0.9949 & 0.9898 \\
            & cohereforai/c4ai-command-r-08-2024 & 0.9943 & 0.9887 \\
            & gpt-4.1-2025-04-14 & 0.9933 & 0.9868 \\
        \hline
        \multirow{4}{*}{English-only (\textbf{lowest metrics})} 
            & qwen/qwen2.5-72b-instruct & 0.9891 & 0.9784 \\
            & deepseek-ai/deepseek-r1-distill-qwen-32b & 0.9873 & 0.9749 \\
            & mistralai/ministral-8b-instruct-2410 & 0.9777 & 0.9565 \\
            & zai-org/glm-4-32b-0414 & 0.9766 & 0.9543 \\
            & mistralai/magistral-small-2507 & 0.9630 & 0.9565 \\
        \hline
        \multirow{4}{*}{Russian-only (\textbf{highest metrics})} 
            & gemma-1.1-7b-it & 1.0000 & 1.0000 \\
            & qwen2-7b-instruct & 0.9986 & 0.9972 \\
            & google/gemma-2-27b-it & 0.9980 & 0.9960 \\
            & cohereforai/c4ai-command-r-08-2024 & 0.9979 & 0.9959 \\
            & phi-3-mini-128k-instruct & 0.9973 & 0.9946 \\
        \hline
        \multirow{4}{*}{Russian-only (\textbf{lowest metrics})} 
            & rugpt & 0.9725 & 0.9465 \\
            & llama-7b & 0.9717 & 0.9449 \\
            & o1-preview-2024-09-12 & 0.9620 & 0.9268 \\
            & jamba-v0.1 & 0.9322 & 0.8730 \\
            & microsoft/phi-3-medium-128k-instruct & 0.9091 & 0.8333 \\
        \hline
        \multirow{4}{*}{Bilingual (\textbf{highest metrics})} 
            & gemma-1.1-7b-it & 1.0000 & 1.0000 \\
            & o1-mini-2024-09-12 & 1.0000 & 1.0000 \\
            & phi-3-mini-128k-instruct & 0.9987 & 0.9973 \\
            & qwen/qwq-32b & 0.9964 & 0.9982 \\
            & nvidia/llama-3.1-nemotron-70b-instruct-hf & 0.9974 & 0.9949 \\
        \hline
        \multirow{4}{*}{Bilingual (\textbf{lowest metrics})} 
            & o1-preview-2024-09-12 & 0.9268 & 0.9268 \\
            & llama-7b & 0.9614 & 0.9256 \\
            & mistralai/magistral-small-2507 & 0.9608 & 0.9246 \\
            & microsoft/phi-3-medium-128k-instruct & 0.9412 & 0.8889 \\
            & jamba-v0.1 & 0.9138 & 0.8413 \\
        \hline
    \end{tabular}
    \label{tab:classification_results_models}
\end{table*}

\section{Creation of manually modified AI texts}
\label{app:manually_modified_texts}

The following protocol was applied consistently to both English and Russian sub-corpora to ensure cross-linguistic stylistic robustness. Initial AI-generated samples were extracted from the test set using a stratified sampling approach based on domain and document length. While we aimed for a balanced distribution, the final composition reflects the inherent constraints of the source data; for instance, certain domains (e.g., social media posts/tweets) naturally contain fewer long-form entries compared to news or academic documents. 

Editors were tasked with manually intervening in the AI-generated text to disrupt predictable statistical patterns. To maximize the structural diversity of the resulting "hybrid" texts, editors followed three core principles:

\begin{itemize}
    \item Length variation: editors maintained a rotating cycle of short, medium, and long-form texts to ensure the model remains length-agnostic.
    \item Variable integration density: the ratio of human-to-AI content was intentionally varied. In some instances, the AI was used only for a single concluding or introductory segment. In others, a "checkered" pattern was applied, where every second sentence was written or edited by a human.
    \item Structural randomization: intervention points were not restricted to sentence boundaries. Editors, for example, were encouraged to interweave human-written phrases (minimum five words) within AI-generated paragraphs.
\end{itemize}

Importantly, human editors did not produce explicit boundary annotations or define AI-generated spans as part of the annotation process. During editing, editors used a technical separator token (e.g., \_SEP\_) solely as a working delimiter to distinguish newly written or modified segments from the surrounding AI-generated text. These separators were not intended as boundary annotations and were not used directly in the experiments. The automatic construction of boundaries relied on a precise character-level comparison between the final edited version and the original AI-generated source text. Any character sequence preserved from the source text and located outside of the editor's delimiters was definitively classified as AI-generated, and its exact offsets in the final text were recorded. Therefore, the final boundary-level segmentation used in downstream experiments was constructed automatically from the resulting edited texts and the known generation history, without manual specification, correction, or post-hoc adjustment. 

By avoiding fixed patterns of intervention (e.g., only editing the beginning or only replacing full sentences), we created a sub-corpus that forces the detection models to focus on granular semantic and stylistic inconsistencies rather than simple structural markers.

\section{Annotation Examples}
\label{app:annotation_examples}

\begin{figure*}[!ht]
    \centering
    \includegraphics[width=0.9\textwidth,keepaspectratio]{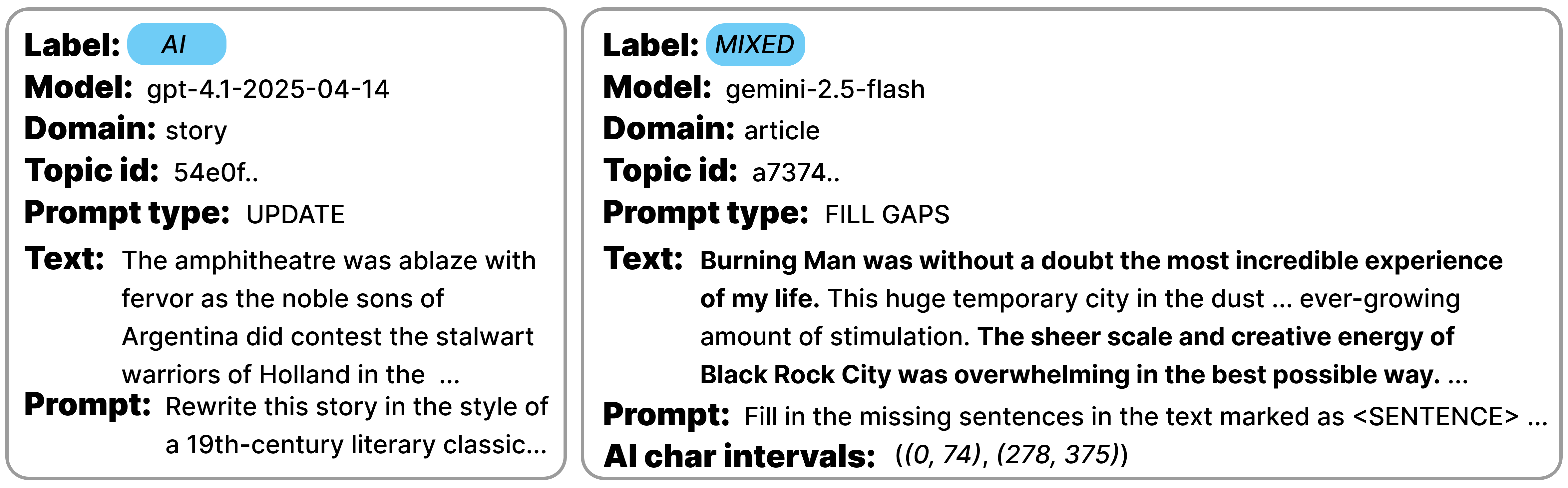}
    \caption{Example annotations from the \DatasetName{} English dataset: a classification sample (top) and a detection sample (bottom).}
    \label{fig:example}
\end{figure*}

Figure~\ref{fig:example} illustrates the two annotation schemas in \DatasetName{}: a binary classification sample (top) and a detection sample with character-level AI-interval annotations (bottom).

\section{Qualitative Examples}
\label{sec:qualitative_examples}

To illustrate the capabilities of our detection baseline, we present qualitative examples from the test set in Table~\ref{tab:examples}. These samples demonstrate the performance of the GigaCheck (DN-DAB-DETR) model trained with a frozen Mistral-7B backbone across different mixed-authorship scenarios. We compare the Ground Truth (GT) annotations with the model's predicted intervals to highlight both successful detections and boundary inconsistencies.

\begin{table*}[!ht]
    \centering
    \caption{Comparison of \textbf{Ground Truth (GT)} labels and model \textbf{Outputs} for AI interval detection. \textbf{Bold text} marks AI-generated segments. While the model successfully detects spans of varying lengths, minor character-level boundary errors (e.g., splitting words) are observed.}
    \begin{tabular}{p{0.95\textwidth}}
        \toprule
        \textbf{Case 1: Manual Editing of AI Texts} \hfill \textit{Generator: OpenAI GPT-3.5} \\
        \midrule
        \textbf{GT:} \textbf{In recent years, deep learning methods have achieved remarkable results in various computer vision tasks such as image classification, object detection, and segmentation. Non-local video denoising is a challenging problem that aims to reduce noise in video sequences while preserving the details and temporal consistency of the frames. In this paper, we present a novel approach for non-local video denoising using Convolutional Neural Networks (CNNs). Our method leverages the spatial and temporal information of video frames to perform non-local filtering. We trained a CNN to learn the mapping between noisy and clean video frames, which can effectively capture the underlying patterns in the video sequences.} Results of the experiments show how out affroach can be more effective than contemporary methods in as subjective, so objective terms, also this method can evaluate quality in fields \textbf{such as surveillance, medical imaging, and entertainment.}
        \\[6pt]
        \textbf{Output:} \textbf{In recent years, deep learning methods have achieved remarkable results in various computer vision tasks such as image classification, object detection, and segmentation. Non-local video denoising is a challenging problem that aims to reduce noise in video sequences while preserving the details and temporal consistency of the frames. In this paper, we present a novel approach for non-local video denoising using Convolutional Neural Networks (CNNs). Our method leverages the spatial and temporal information of video frames to perform non-local filtering. We trained a CNN to learn the mapping between noisy and clean video frames, which can effectively capture the underlying patterns in the video sequences. Results of the} experiments show how out affroach can be more effective than contemporary methods in as subjective, so objective terms, also this method can evaluate quality in fields such as survei\textbf{llance, medical imaging, and entertainment.} \\
        \midrule
        \vspace{2mm} % Space between examples
        \textbf{Case 2: Human Text Gap-Filling} \hfill \textit{Generator: Gemini-2.5 Flash} \\
        \midrule
        \textbf{GT:} It was my turn. \textbf{It was finally time for the ritual to begin.} I sat in a small abandoned church, on the south side of the town we had created. It was tall, honoring our God so very well. \textbf{The stained-glass windows, though dusty and cracked, still glowed with a faint, ethereal light.} I had a very simple white dress on, and I patted it down onto my pale legs. It was not like ones that I had seen on other Daughters in the villages around our town. I was excited, but thankfully not excited enough to break the ritual. \textbf{My hands were clasped in my lap, trying to keep steady.} And there I sat, and waited for the Father. He quickly came, and off we went, down the hallway into the back of the church. He led me slowly, carrying his candle. A candle, instead of a torch or a lantern.\\[6pt]
        \textbf{Output:} It was my turn. It was finally time for the ritual to begin. I sat in a small abandoned church, on the south side of the town we had created. It was tall, honoring our God so very well. Th\textbf{e stained-glass windows, though dusty and cracked, still glowed with a faint, ethereal light}. I had a very simple white dress on, and I patted it down onto my pale legs. It was not like ones that I had seen on other Daughters in the villages around our town. I was excited, but thankfully not excited enough to break the rit\textbf{ual. My hands were clasped in my lap, trying to keep steady.} And there I sat, and waited for the Father. He quickly came, and off we went, down the hallway into the back of the church. He led me slowly, carrying his candle. A candle, instead of a torch or a lantern.\\
        \bottomrule
    \end{tabular}
    \label{tab:examples}
\end{table*}

\section{Zero-Shot Detector Results}
\label{app:zeroshot_results}

Table~\ref{tab:zeroshot_results} reports binary classification performance for three zero-shot detectors evaluated on the same held-out test sets used for supervised baselines (Section~\ref{subsec:baseline_experiments}). Binoculars and Fast-DetectGPT are evaluated with their default English-tuned thresholds. For DetectGPT we report only AUROC, as the method does not produce calibrated class scores suitable for direct accuracy/F1 computation.

\begin{table}[htbp]
\centering
\caption{Zero-shot detector performance on the binary classification test sets (Overall /
English / Russian). DetectGPT reports AUROC only; dashes indicate unavailable metrics.
Fine-tuned BIL Mistral-7B results are reproduced from Table~\ref{tab:classification_results}
for reference.}
\label{tab:zeroshot_results}
\small
\resizebox{\columnwidth}{!}{%
\begin{tabular}{l l c c c c}
\hline
\textbf{Detector} & \textbf{Split} & \textbf{Accuracy} & \textbf{AI F1}
    & \textbf{Human F1} & \textbf{AUROC} \\
\hline
\multirow{3}{*}{Binoculars}
  & Overall & 0.493 & 0.282 & 0.610 & -- \\
  & English & 0.603 & 0.533 & 0.650 & -- \\
  & Russian & 0.412 & 0.022 & 0.580 & -- \\
\hline
\multirow{3}{*}{DetectGPT}
  & Overall & --    & --    & --    & 0.582 \\
  & English & --    & --    & --    & 0.613 \\
  & Russian & --    & --    & --    & 0.588 \\
\hline
\multirow{3}{*}{Fast-DetectGPT}
  & Overall & 0.663 & 0.692 & 0.626 & -- \\
  & English & 0.689 & 0.693 & 0.685 & -- \\
  & Russian & 0.643 & 0.692 & 0.576 & -- \\
\hline
\multirow{3}{*}{\textit{BIL Mistral-7B (fine-tuned)}}
  & \textit{Overall} & \textit{0.985} & \textit{0.986} & \textit{0.980} & \textit{--} \\
  & \textit{English} & \textit{0.985} & \textit{0.986} & \textit{0.979} & \textit{--} \\
  & \textit{Russian} & \textit{0.984} & \textit{0.986} & \textit{0.980} & \textit{--} \\
\hline
\end{tabular}%
}
\end{table}

The results confirm three conclusions. First, zero-shot performance is consistently poor, with even the best detector (Fast-DetectGPT) falling substantially short of the fine-tuned model. Second, Russian is markedly harder for all zero-shot methods: Binoculars in particular struggles severely, suggesting its perplexity-ratio signal is not calibrated for Russian vocabulary. Third, the difficulty is not an artifact of our evaluation protocol — DetectGPT's AUROC remains close to chance level, confirming that meaningful detection on this dataset requires in-domain supervision.
\newpage
\section{Detection Error Analysis}
\label{sec:appendix_errors}

This appendix expands the error analysis summarised in Section~\ref{sec:error_analysis}. The analysis is performed on the \textit{mixed} subset of the \texttt{\DatasetName{}} test split (4{,}716 texts containing 13{,}589 ground-truth AI character intervals), using the two bilingual interval-detection backbones from our main experiments (Section~\ref{subsec:logo_experiments}): \textbf{BIL Mistral-7B-DETR} and \textbf{BIL mmBERT-DETR}. The pipeline runs each detector once, caches per-sample predictions, and then aggregates them along several orthogonal axes (generator metadata and text-level properties) so that individual cuts can be re-derived without re-running inference.

\subsection{Error Taxonomy}
\label{subsec:err_taxonomy}

For one text, let $G = \{g_1, \dots, g_n\}$ be the GT character intervals and $P = \{p_1, \dots, p_m\}$ the predicted intervals. We construct the bipartite overlap graph $E = \{(i,j) : g_i \cap p_j \neq \emptyset\}$ and assign one of six labels to each interval based on its degree:

\begin{itemize}
\item \textbf{False negative.} A GT interval $g_i$ with no overlapping prediction ($\deg_E(i) = 0$).
\item \textbf{False positive.} A predicted interval $p_j$ with no overlapping GT ($\deg_E(j) = 0$).
\item \textbf{Good match.} GT $g_i$ and prediction $p_j$ form a 1-to-1 match ($\deg_E(i) = \deg_E(j) = 1$) with $\mathrm{IoU}(g_i, p_j) \geq 0.8$.
\item \textbf{Poor match.} As above but $\mathrm{IoU}(g_i, p_j) < 0.8$ --- the detector found the right region but the boundary is far off.
\item \textbf{Over-segmentation.} A GT interval matched by multiple predictions ($\deg_E(i) > 1$); the detector fragmented one continuous AI span.
\item \textbf{Under-segmentation.} A predicted interval matched by multiple GTs ($\deg_E(j) > 1$); the detector merged two AI spans separated by a short human insertion.
\end{itemize}

For every good match we additionally record the boundary offsets in characters, $(\mathrm{start}_p - \mathrm{start}_g, \mathrm{end}_p - \mathrm{end}_g)$, used in our boundary-bleeding analysis. From the per-text counts we derive interval-level metrics:
\[
\mathrm{precision} = \frac{N_\mathrm{good} + N_\mathrm{poor}}{N_\mathrm{good} + N_\mathrm{poor} + N_\mathrm{FP}}, \quad
\mathrm{recall} = \frac{N_\mathrm{good} + N_\mathrm{poor}}{N_\mathrm{good} + N_\mathrm{poor} + N_\mathrm{FN}},
\]
$F_1 = 2PR/(P+R)$, and the \textit{clean-match rate} $= N_\mathrm{good} / |G|$ which ignores poor and segmentation-level matches.

\begin{figure}[ht!]
    \centering
    \includegraphics[width=\columnwidth]{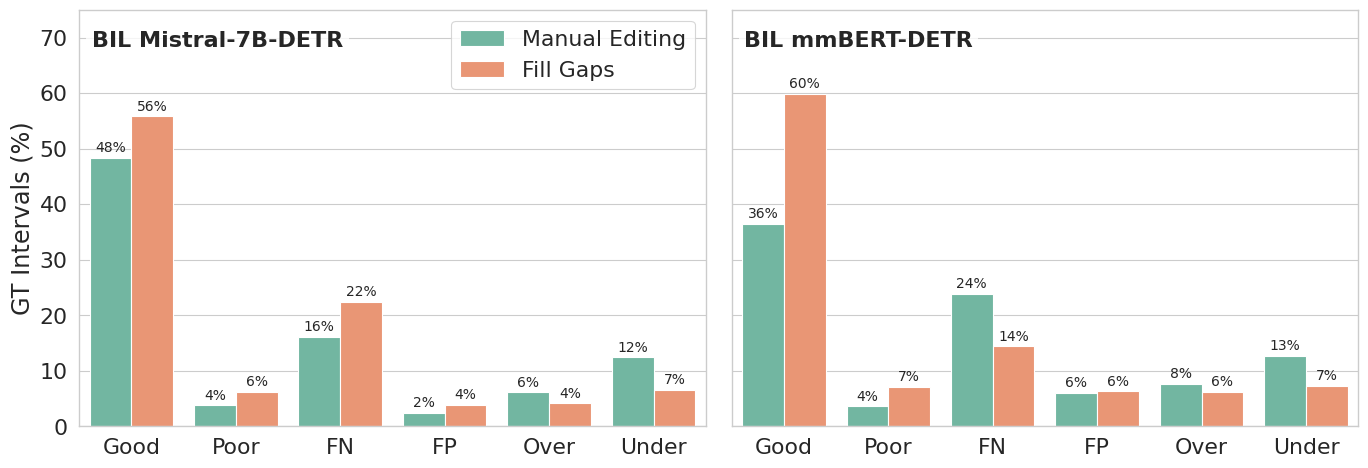}
    \caption{Aggregate distribution of the six error categories on the mixed subset, split by generation regime (manual editing vs. fill gaps) and by detector. Manual editing produces noticeably fewer good matches and more under-segmentation, confirming that targeted human edits disrupt detectable boundaries.}
    \label{fig:err_taxonomy}
\end{figure}

\subsection{Generator Family Effects}
\label{subsec:err_family}

Figure~\ref{fig:err_family_heatmap} (main body) reports the per-family error-type distribution for both detectors. Three observations are worth highlighting.

First, the frontier closed-source families (OpenAI, Gemini) together account for $\geq 90\%$ of the mixed test set; they reach interval-level F1 in the range $[0.82, 0.87]$ for both detectors and constitute the most reliable working point.

Second, legacy and small open-source families (\texttt{ruGPT}, early Llama) are catastrophically harder, with F1 $\leq 0.50$. The size-conditioned analysis (Section~\ref{subsec:err_size}) shows that this is a parameter-count effect rather than a family-specific one.

Third, several families show \textbf{detector-specific} behaviour: \texttt{GigaChat} reaches F1 = 0.89 on BIL Mistral-7B-DETR but only 0.47 on BIL mmBERT-DETR; \texttt{Cohere} drops from 0.64 to 0.41; \texttt{Mistral} drops from 0.77 to 0.52. This indicates that the two architectures pick up substantially different signals on these generators and motivates the cross-detector consistency analysis in Section~\ref{subsec:err_crossdet}.

\begin{figure}[ht!]
    \centering
    \includegraphics[width=\columnwidth*3/4]{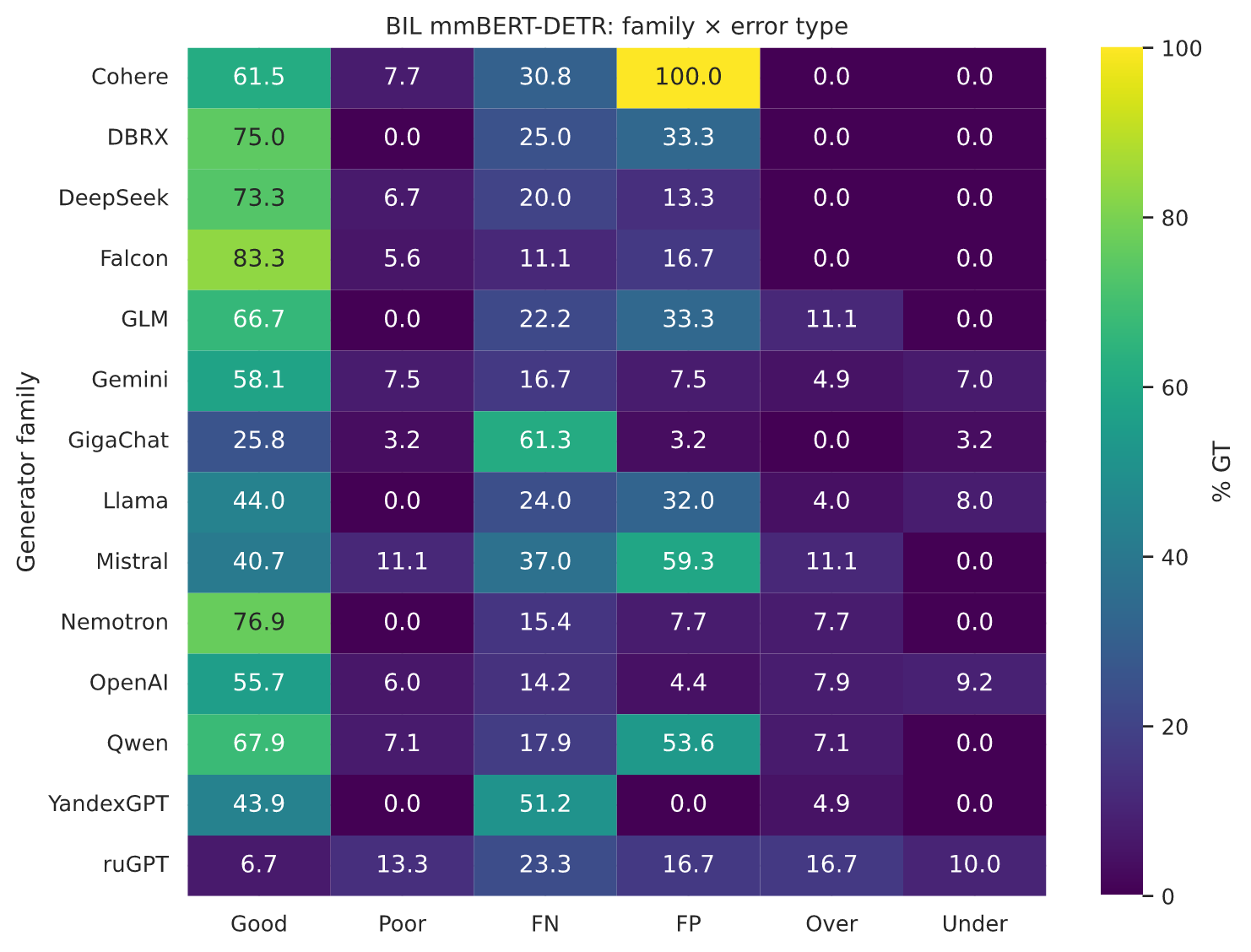}
    \caption{Per-family error-type distribution for the \textbf{BIL mmBERT-DETR} detector (companion to Figure~\ref{fig:err_family_heatmap}). The frontier closed-source families dominate, but the mmBERT backbone shows substantially different behaviour on \texttt{GigaChat}, \texttt{Cohere}, and \texttt{Mistral} compared to the Mistral-7B backbone.}
    \label{fig:err_family_heatmap_mmbert}
\end{figure}

\subsection{Generator Size}
\label{subsec:err_size}

We bucket each generator by parameter count: \textit{small} ($\leq$8B), \textit{medium} (8--30B), \textit{large} ($>$30B open-weights), and \textit{frontier-closed} (proprietary frontier models for which exact parameter counts are not publicly known). Per-bucket interval-level F1 is reported in Table~\ref{tab:err_size}.

\begin{table}[ht]
\centering
\small
\begin{tabular}{lrcc}
\toprule
size bin & $n_\mathrm{texts}$ & F1 (Mistral) & F1 (mmBERT) \\
\midrule
small ($\leq$8B)        & 32   & 0.475 & 0.453 \\
medium (8--30B)         & 32   & 0.903 & 0.746 \\
large ($>$30B open)     & 113  & 0.859 & 0.695 \\
frontier-closed         & 4539 & 0.828 & 0.855 \\
\bottomrule
\end{tabular}
\caption{Interval-level F1 by generator size bin. The small bucket is dominated by \texttt{llama-7b}, \texttt{Ministral-8B}, and \texttt{ruGPT}, all of which were rare in the detector training data.}
\label{tab:err_size}
\end{table}

\begin{figure}[ht!]
    \centering
    \includegraphics[width=\columnwidth]{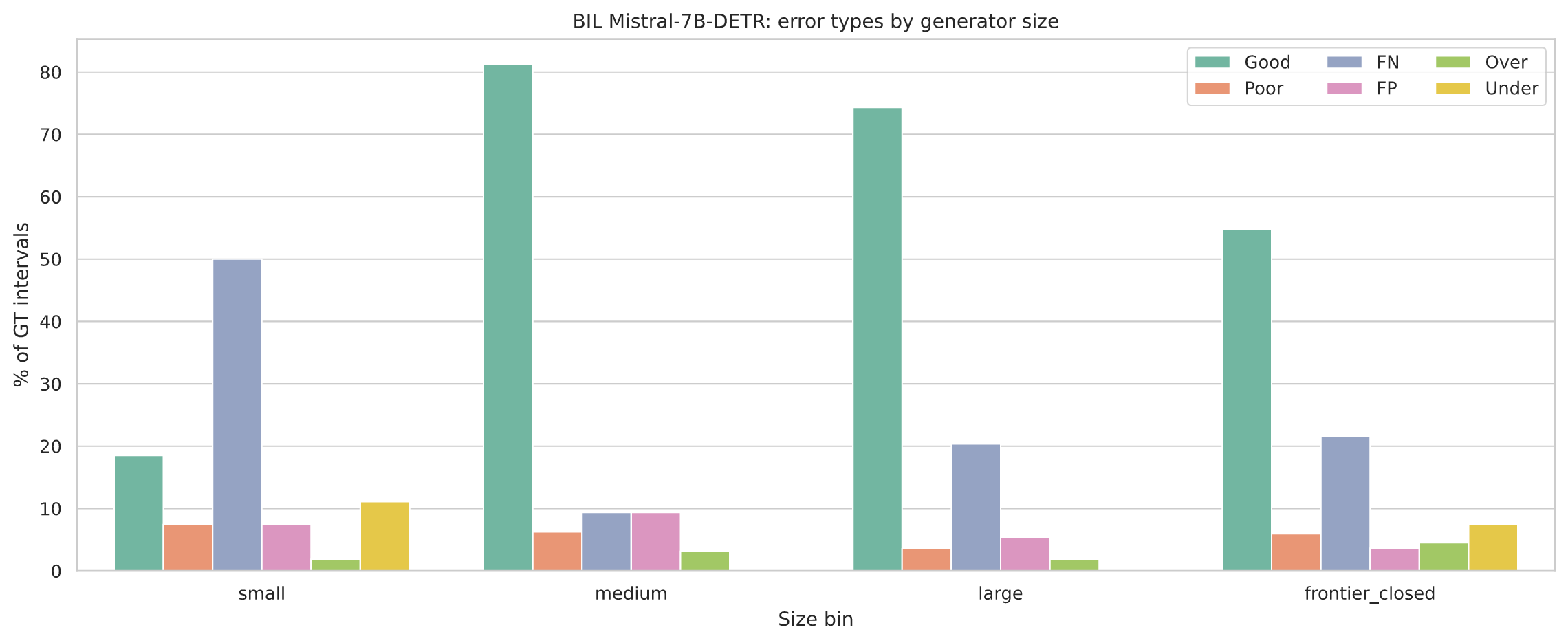}
    \\[4pt]
    \includegraphics[width=\columnwidth]{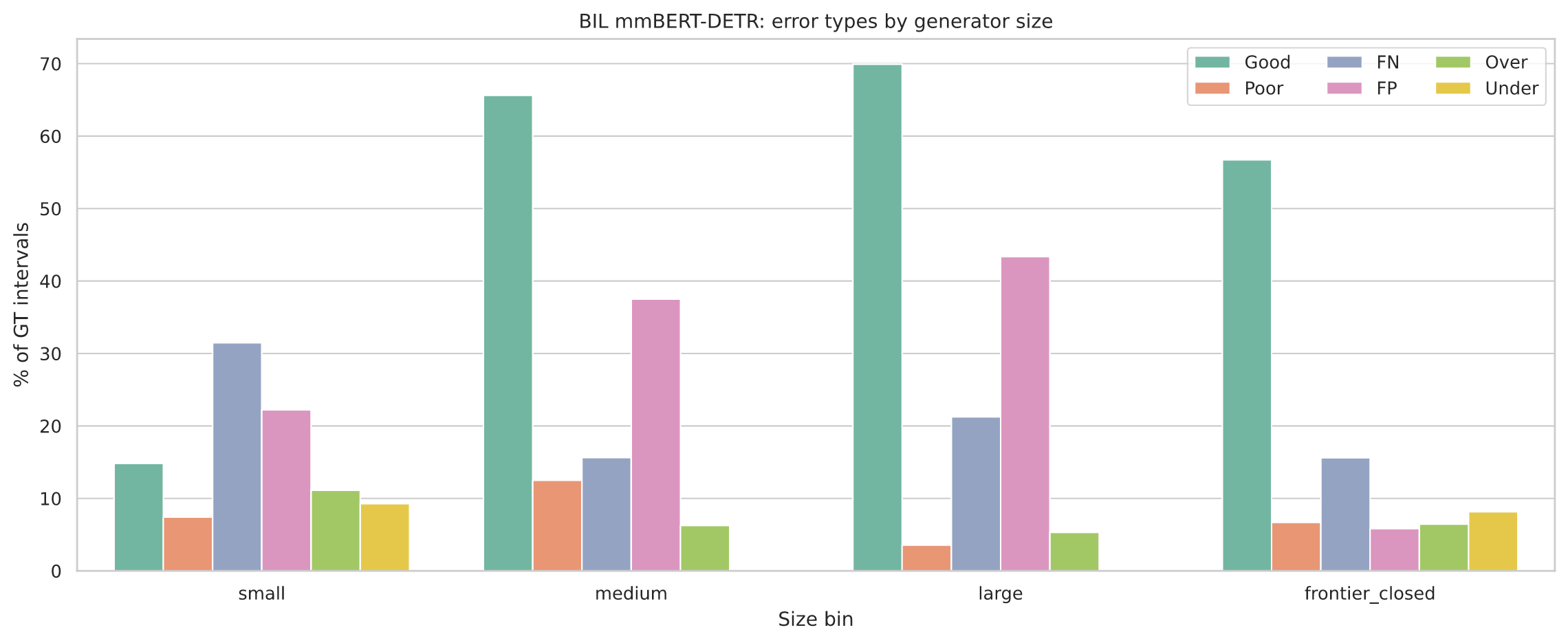}
    \caption{Per-error-category share by generator size bin, BIL Mistral-7B-DETR (top) and BIL mmBERT-DETR (bottom).}
    \label{fig:err_size}
\end{figure}

The small bucket is the only one where both detectors agree on a sharp drop (F1 below 0.50). We caution that the small/medium bins are observational (32 mixed texts each), so absolute numbers should be read as indicative rather than conclusive; we report these bins for completeness, but the frontier-closed bin ($n=4{,}539$) provides the only statistically powered comparison. Figure~\ref{fig:err_size} shows the per-error-category breakdown for both detectors.

\subsection{Generator Recency}
\label{subsec:err_recency}

\begin{figure*}[ht!]
    \centering
    \begin{subfigure}{0.48\textwidth}
        \centering
        \includegraphics[width=\linewidth]{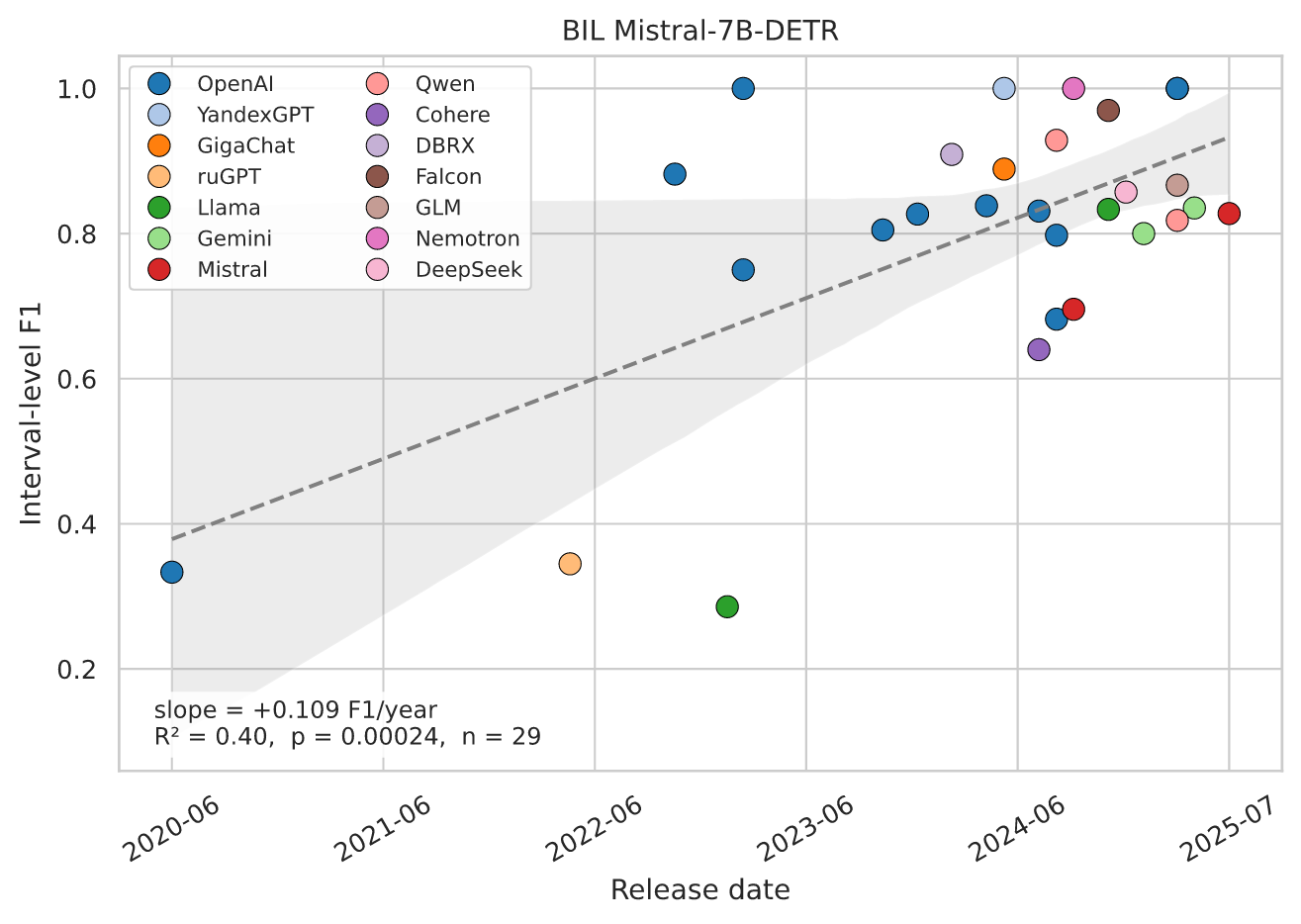}
        \caption{BIL Mistral-7B-DETR}
        \label{fig:err_release_mistral}
    \end{subfigure}
    \hfill
    \begin{subfigure}{0.48\textwidth}
        \centering
        \includegraphics[width=\linewidth]{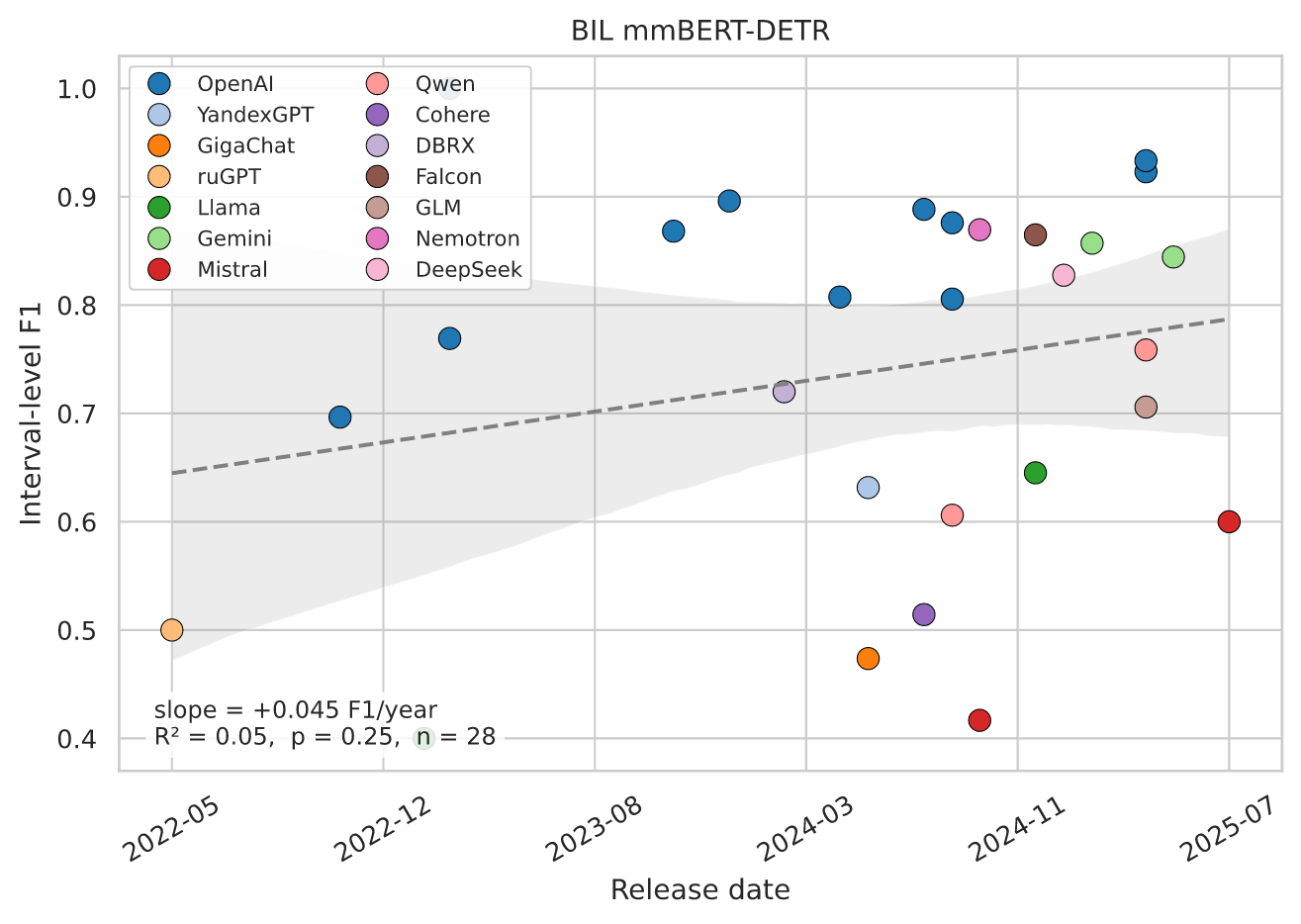}
        \caption{BIL mmBERT-DETR}
        \label{fig:err_release_mmbert}
    \end{subfigure}
    \caption{Per-generator interval-level F1 vs.\ release date with OLS fits and 95\% CI bands. (a) Mistral-7B backbone: positive slope ($+0.11$ F1/year, $p = 2.4\!\times\!10^{-4}$) driven by legacy small open-source generators (\texttt{llama-7b}, \texttt{ruGPT}) being out-of-distribution for the detector. (b) mmBERT backbone: same direction but the slope is not significant ($p = 0.25$). Both panels are explained by the recency--training-distribution confound discussed above.}
    \label{fig:err_release}
\end{figure*}

Figure~\ref{fig:err_release} shows per-generator F1 against release date together with OLS fits. For BIL Mistral-7B-DETR the slope is $+0.109$ F1/year ($R^2 = 0.40$, $p = 2.4\!\times\!10^{-4}$, $n = 29$); for BIL mmBERT-DETR it is $+0.045$ F1/year ($R^2 = 0.05$, $p = 0.25$, $n = 28$, not significant). The positive direction is the opposite of what one would expect if newer generators were inherently harder to detect.

The most plausible explanation is a \textbf{confound between recency and detector training distribution}. The older generators present in our test set --- \texttt{llama-7b} (2023-02), \texttt{ruGPT} (2022-05) --- are precisely the legacy small open-source models that were under-represented in the detector training corpus, while modern closed-source generators (\texttt{gpt-4o}, \texttt{gemini-2.5-flash}) match the training distribution well. Once the size-based out-of-distribution effect (Section~\ref{subsec:err_size}) is taken into account, recency itself does not appear to drive detection difficulty in the date range covered by our dataset. We therefore advise against interpreting Figure~\ref{fig:err_release} as evidence of a recency benefit.

A cleaner test of pure recency would require holding generator size and openness constant across release dates, which our current test set does not allow.

\subsection{Generator Quality (Arena Elo Proxy)}
\label{subsec:err_quality}

As a public proxy for generator output quality we use Chatbot Arena Elo scores (snapshot late 2025). Elo is a noisy community-voting signal and our use of it is purely ordinal. For the 23 generators with available Elo we observe a positive slope: $+0.08$ F1 per $+100$ Elo for BIL Mistral-7B-DETR ($R^2 = 0.17$, $p = 0.052$) and $+0.10$ for BIL mmBERT-DETR ($R^2 = 0.32$, $p = 4.9\!\times\!10^{-3}$).

Higher-quality generators are detected slightly \textit{better}, consistent with the hypothesis that more polished AI text exhibits a more distinctive stylistic signature. However, Elo and detector training distribution are themselves correlated --- the highest-Elo generators (GPT-4o, Gemini, o1) are also the best represented in training data --- so this is not a clean causal claim. We include this analysis to examine whether generator quality, as measured by a public proxy, correlates with detection difficulty.

\begin{figure}[ht!]
    \centering
    \includegraphics[width=\columnwidth]{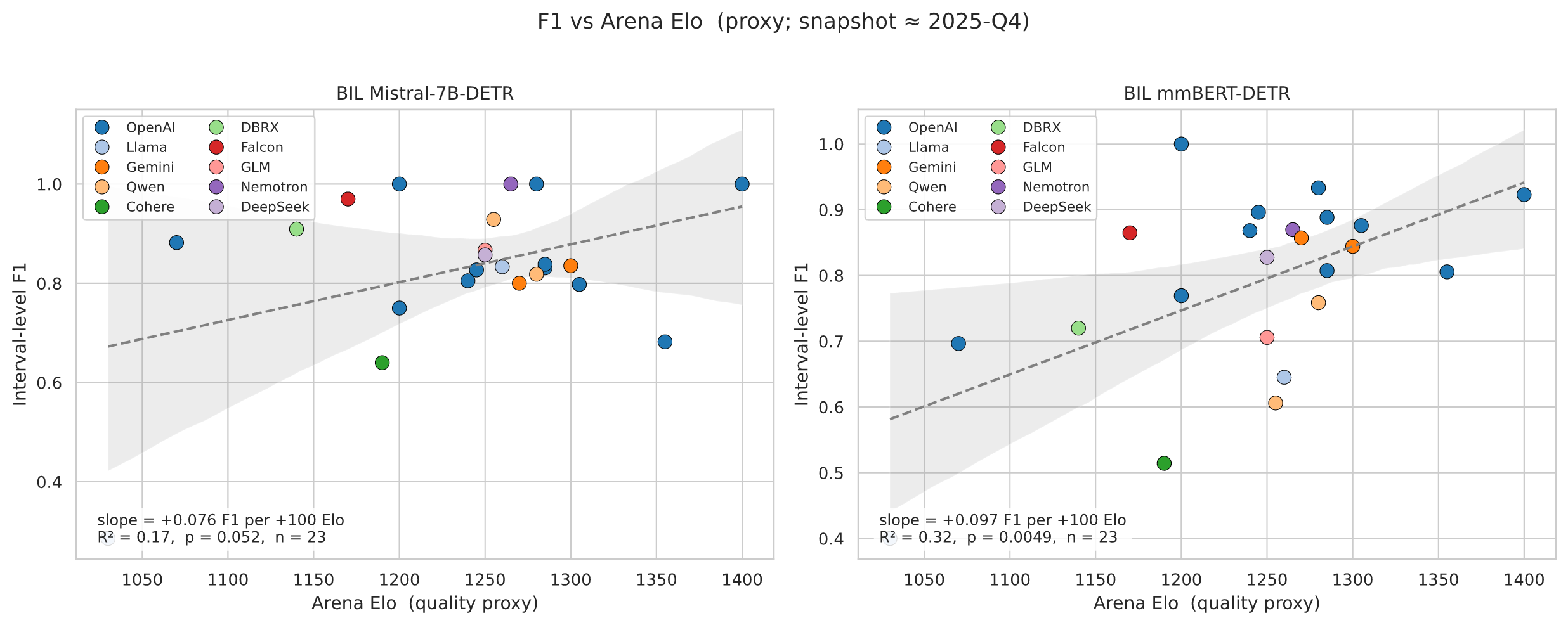}
    \caption{Per-generator F1 vs. Chatbot Arena Elo (proxy for generation quality), with OLS fit and 95\% CI band.}
    \label{fig:err_arena}
\end{figure}

\subsection{Text-level Factors}
\label{subsec:err_text}

\begin{figure}[ht!]
    \centering
    \includegraphics[width=\columnwidth]{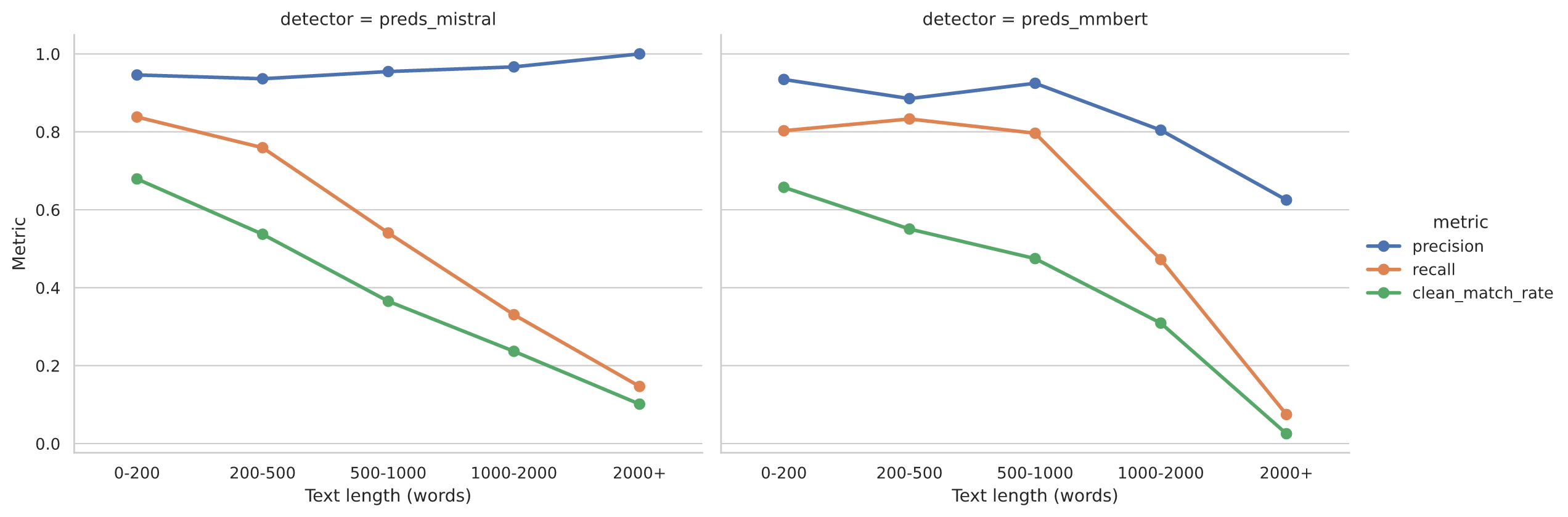}
    \caption{Precision, recall, and clean-match rate vs. text length (words), for both detectors.}
    \label{fig:err_length}
\end{figure}

\paragraph{Text length.}
Figure~\ref{fig:err_length} reports precision, recall, and clean-match rate as a function of text length. Precision stays high ($\geq 0.88$) across all length bins for both detectors, while recall collapses on long texts: from 0.84 ($\leq$200 words) to 0.15 ($\geq$2000 words) for BIL Mistral-7B-DETR. The detector silently misses AI spans in long contexts rather than producing spurious predictions. Per-sample Spearman correlation between text length (words) and clean-match rate is $\rho = -0.31$ ($p < 10^{-100}$, $n = 4716$) for BIL Mistral-7B-DETR and $\rho = -0.27$ for BIL mmBERT-DETR.

\begin{figure}[ht!]
    \centering
    \includegraphics[width=\columnwidth]{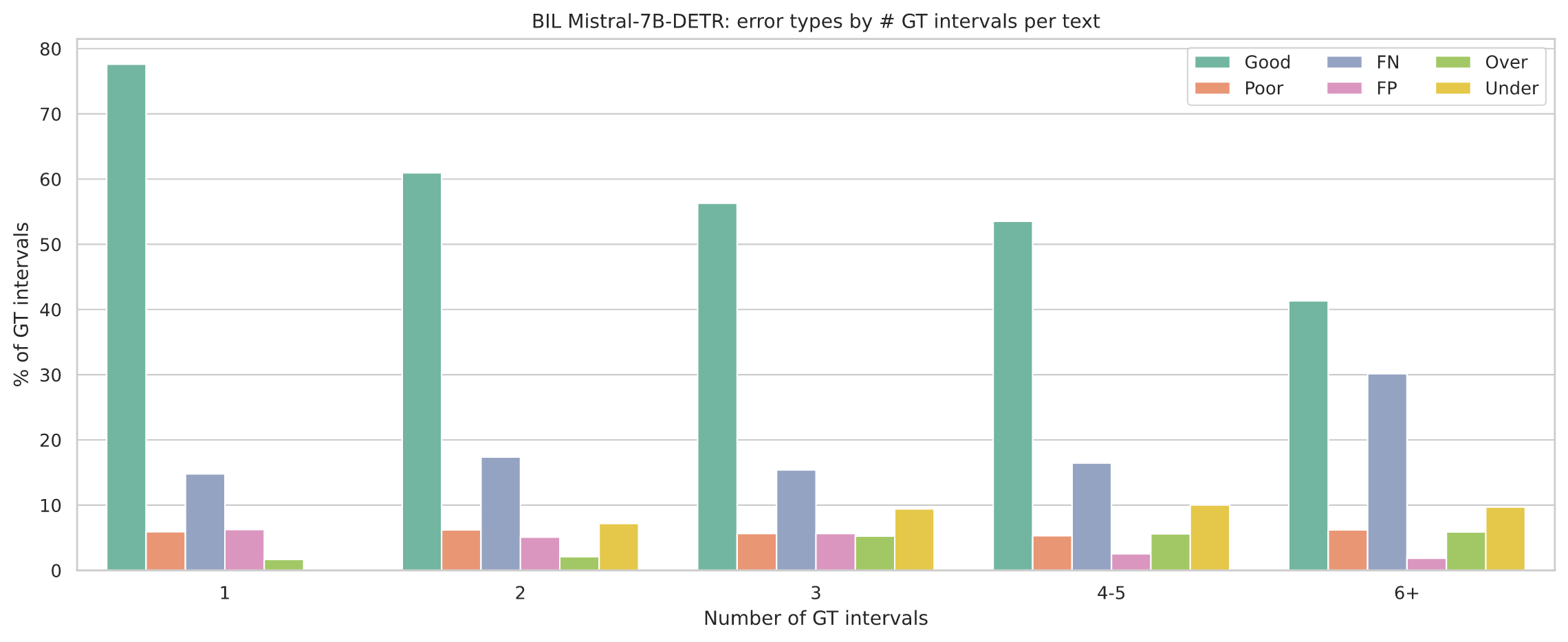}
    \\[4pt]
    \includegraphics[width=\columnwidth]{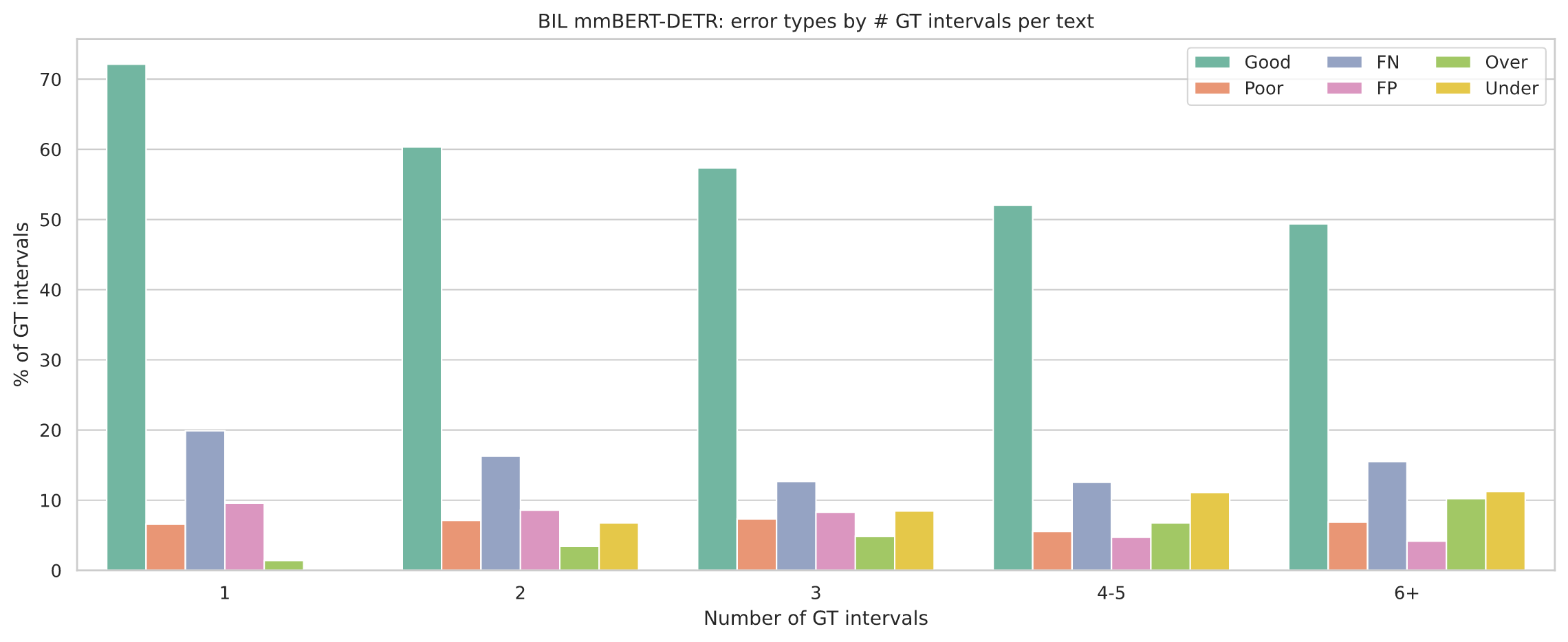}
    \caption{Per-error-category share by number of GT intervals per text, BIL Mistral-7B-DETR (top) and BIL mmBERT-DETR (bottom).}
    \label{fig:err_nintervals}
\end{figure}

\paragraph{Number of GT intervals per text.}
F1 stays in $[0.85, 0.89]$ for texts with 1--5 GT intervals and drops to $\sim 0.75$ for texts with 6+ intervals (Figure~\ref{fig:err_nintervals}). The dominant degradation mode at high interval count is under-segmentation: the rate of GT intervals being merged into one prediction grows from 0\% (single-interval texts) to roughly 10\% (6+ intervals). Per-sample Spearman $\rho$ between interval count and clean-match rate is $-0.38$ for BIL Mistral-7B-DETR ($p < 10^{-150}$).

\begin{figure}[ht!]
    \centering
    \includegraphics[width=\columnwidth]{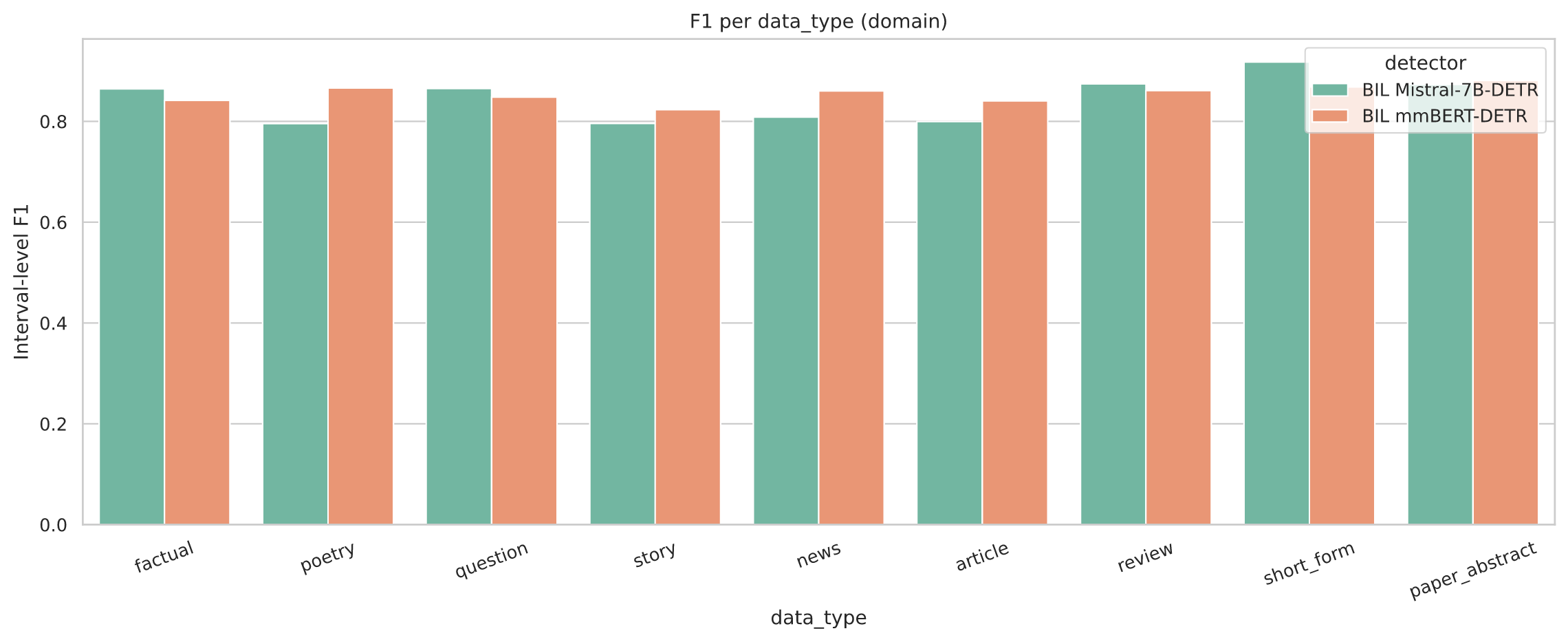}
    \caption{Interval-level F1 per \texttt{data\_type} (domain proxy), both detectors.}
    \label{fig:err_domain}
\end{figure}

\paragraph{Domain.}
F1 varies only mildly across the nine \texttt{data\_type} categories (Figure~\ref{fig:err_domain}). Both detectors find templated short forms (\texttt{paper\_abstract}, \texttt{short\_form}, \texttt{review}) easier and creative long forms (\texttt{story}, \texttt{article}) harder. Poetry shows a notable detector-architecture flip: it is the worst domain for BIL Mistral-7B-DETR (F1 = 0.80) but among the best for BIL mmBERT-DETR (F1 = 0.87), suggesting that the two architectures rely on different stylistic cues for highly structured text.

\begin{figure}[ht!]
    \centering
    \includegraphics[width=\columnwidth]{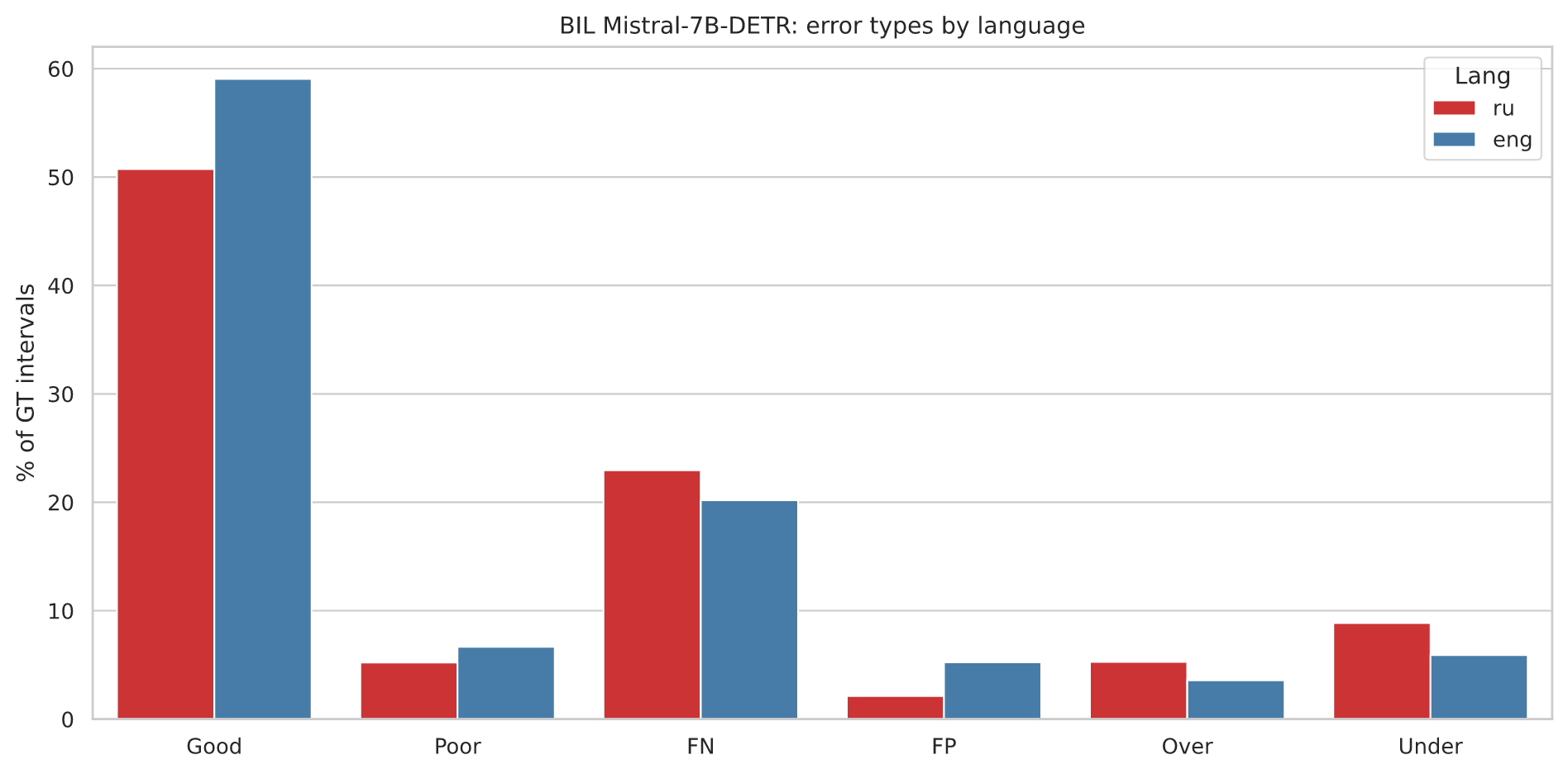}
    \\[4pt]
    \includegraphics[width=\columnwidth]{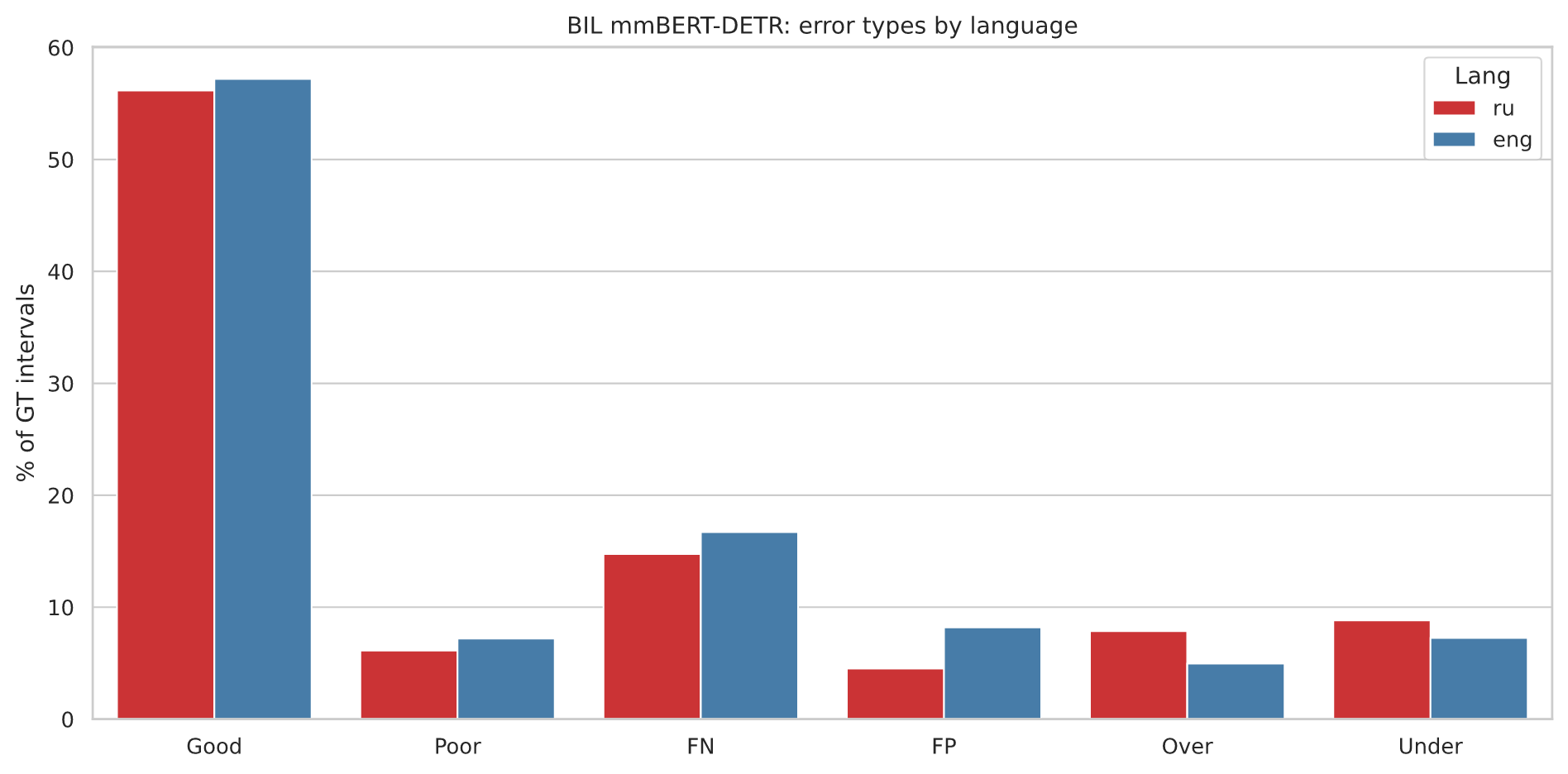}
    \caption{Per-language error-category breakdown for BIL Mistral-7B-DETR (top) and BIL mmBERT-DETR (bottom).}
    \label{fig:err_lang}
\end{figure}

\paragraph{Language.}
The bilingual split shows no systematic disadvantage for either language (Figure~\ref{fig:err_lang}); BIL Mistral-7B-DETR is slightly better on English ($\Delta F_1 = +0.02$) while BIL mmBERT-DETR is slightly better on Russian ($\Delta F_1 = +0.03$). All differences fall within $\pm 3$ F1 points and do not change the qualitative conclusions of the previous subsections.

\subsection{Cross-detector Agreement}
\label{subsec:err_crossdet}

To separate generator-intrinsic difficulty from architecture-specific failure modes, we compute Spearman and Pearson correlations between the two detectors' per-group F1 across multiple groupings (Table~\ref{tab:err_crossdet}). The per-model Spearman of $0.40$ ($p = 0.04$) is statistically significant but modest in magnitude --- the two architectures do not fully agree on which generators are hard.

\begin{table}[ht]
\centering
\small
\begin{tabular}{lrccc}
\toprule
grouping & $n$ & Spearman $\rho$ & $p$ & Pearson $r$ \\
\midrule
by model           & 28 & $+0.40$ & 0.04 & 0.59 \\
by family          & 14 & $+0.45$ & 0.11 & 0.55 \\
by size bin$^\dagger$        & 4  & $+0.40$ & 0.60 & 0.87 \\
by release y--m    & 18 & $-0.20$ & 0.42 & 0.61 \\
by Arena Elo bin$^\dagger$   & 5  & $\phantom{+}0.00$ & 1.00 & 0.42 \\
\bottomrule
\end{tabular}
\caption{Cross-detector agreement (BIL Mistral-7B-DETR vs.\ BIL mmBERT-DETR) on per-group F1. $^\dagger$Correlations for groupings with $n < 10$ are reported for completeness but lack statistical power.}
\label{tab:err_crossdet}
\end{table}

We interpret this as evidence that detection difficulty has both \textbf{data-intrinsic (generator-driven)} and \textbf{architecture-specific} components. Practically: detector behaviour on a previously unseen generator family cannot be reliably extrapolated from a single architecture's per-family profile. Reporting analyses for at least two detector backbones, as we do throughout this appendix, is therefore important for any claim of the form ``generator $X$ is hard / easy''.

\subsection{Transferability to External Benchmarks}
\label{subsec:transferability}

To assess transferability beyond the LLMTrace distribution, we evaluate both detectors zero-shot (no target-domain training) on the RoFT~\citep{dugan2023real} and RoFT-ChatGPT~\citep{kushnareva2023ai} boundary benchmarks. Following~\citet{tolstykh2024gigacheck}, we keep only the top-1 predicted interval per text and project its start to a sentence index using the same character-to-sentence mapping; we report SoftAcc1 and MSE.

Table~\ref{tab:transferability_results} reports our results alongside the baselines from~\citet{kushnareva2023ai} and the \texttt{GigaCheck-DETR\dag} variant of~\citet{tolstykh2024gigacheck}, which shares our pre-trained Mistral-7B-v0.3 backbone but trains its DETR head directly on each target benchmark. Despite using no target-domain training, our Mistral detector matches the human baseline on RoFT (SoftAcc1 0.39 vs 0.40) and outperforms both topology-based baselines (PHD+TS, TLE+TS) on RoFT-ChatGPT --- all of which were trained on the target. It expectedly trails the methods that fully fit the target benchmark (RoBERTa variants, \texttt{GigaCheck-DETR\dag}), and the larger gap on RoFT is consistent with the generator-coverage analysis above: legacy GPT-2/CTRL generators dominating RoFT are under-represented in LLMTrace, whereas the modern generators in RoFT-ChatGPT match our training distribution. The smaller mmBERT backbone transfers noticeably worse despite matching Mistral within ${\sim}1$\% mAP in-distribution (Table~\ref{tab:detection_results}), indicating that backbone capacity becomes more critical out-of-distribution. Overall, LLMTrace-trained detectors transfer non-trivially to external benchmarks without any target-domain fine-tuning, and the residual gap to fully supervised models tracks rather than contradicts the in-distribution diagnosis --- pointing to broader generator coverage, especially of legacy and small open-source models, as the main lever for closing it.

\begin{table}[htbp!]
\centering
\caption{Zero-shot transfer to RoFT and RoFT-ChatGPT. All baselines except ours are trained or calibrated on the target benchmark; their numbers are taken from~\citet{kushnareva2023ai}, except \texttt{GigaCheck-DETR\dag} which is taken from~\citet{tolstykh2024gigacheck} and shares our pre-trained Mistral-7B-v0.3 backbone (its DETR head is trained on each target benchmark). SoftAcc1 and MSE are computed on the top-1 predicted interval, projected to sentence indices following~\citet{tolstykh2024gigacheck}.}
\label{tab:transferability_results}
\small
\begin{tabular}{lcccc}
\toprule
\multirow{2}{*}{\textbf{Method}} & \multicolumn{2}{c}{\textbf{RoFT}} & \multicolumn{2}{c}{\textbf{RoFT-ChatGPT}} \\
 & SoftAcc1 & MSE & SoftAcc1 & MSE \\ \midrule
RoBERTa + SEP & 0.80 & 2.63 & 0.79 & 3.06 \\
Phi-1.5 Perpl. + GB & 0.45 & 6.11 & 0.71 & 3.07 \\
Phi-1.5 Perpl. + LR & 0.50 & 11.9 & 0.73 & 4.77 \\
PHD + TS ML        & 0.46 & 14.40 & 0.36 & 14.45 \\
TLE + TS Binary    & 0.30 & 22.23 & 0.35 & 18.52 \\
Human baseline     & 0.40 & 13.88 & --   & --   \\
GigaCheck-DETR\dag & 0.81 & 2.77  & 0.80 & 1.93 \\
\midrule
BIL Mistral-DETR (zero-shot) & 0.39 & 15.58 & 0.58 & 7.15 \\
BIL mmBERT-DETR (zero-shot)  & 0.32 & 21.67 & 0.48 & 13.04 \\
\bottomrule
\end{tabular}
\end{table}

Beyond boundary detection, we additionally evaluate the binary classifiers from Table~\ref{tab:classification_results} on the MAGE benchmark~\citep{li2023mage}. Following~\citet{tolstykh2024gigacheck}, we report AvgRec (the per-class accuracy average) and AUROC on the \emph{Arbitrary-domains \& Arbitrary-models} test split, which is the standard in-distribution test set used by MAGE-trained baselines. Without any MAGE training, our BIL Mistral-7B classifier retains $\approx 71\%$ of the in-distribution trained classifier's AvgRec (0.68 vs.\ 0.96~\citep{tolstykh2024gigacheck}) and $\approx 80\%$ of its AUROC (0.79 vs.\ 0.99). The BIL mmBERT classifier drops further (AvgRec 0.59, AUROC 0.63), reproducing the smaller-backbone deficit observed above on RoFT (Table~\ref{tab:transferability_results}); the two transferability experiments thus agree that backbone capacity, not architectural family, is the dominant lever once the target distribution drifts away from training. Full numbers and the trained baselines are listed in Table~\ref{tab:transferability_mage}.

\begin{table}[htbp!]
\centering
\caption{Zero-shot transfer of the binary classifiers from Table~\ref{tab:classification_results} to MAGE~\citep{li2023mage} on the \emph{Arbitrary-domains \& Arbitrary-models} test split (n${=}56{,}819$). Trained baseline numbers are taken from~\citet{tolstykh2024gigacheck}; the Longformer row in turn cites the value reported there from~\citet{li2023mage}.}
\label{tab:transferability_mage}
\small
\begin{tabular}{lcc}
\toprule
\textbf{Method} & \textbf{AvgRec} & \textbf{AUROC} \\
\midrule
Longformer~\citep{beltagy2020longformer} (trained on MAGE)            & 0.91 & 0.99 \\
GigaCheck Mistral-7B~\citep{tolstykh2024gigacheck} (trained on MAGE) & 0.96 & 0.99 \\
\midrule
BIL Mistral-7B classifier (zero-shot) & 0.68 & 0.79 \\
BIL mmBERT classifier (zero-shot)     & 0.59 & 0.63 \\
\bottomrule
\end{tabular}
\end{table}

\end{document}